\documentclass[11pt]{article}

\usepackage[margin=1in]{geometry}
\usepackage{iftex}
\ifPDFTeX
  \usepackage[T1]{fontenc}
  \usepackage{lmodern}
\else
  \usepackage{fontspec}
\fi
\usepackage{microtype}
\usepackage{amsmath,amssymb}
\usepackage{booktabs}
\usepackage{multirow}
\usepackage{array}
\usepackage{calc}
\usepackage{tabularx}
\usepackage{graphicx}
\usepackage{xcolor}
\usepackage{caption}
\usepackage{subcaption}
\usepackage{enumitem}
\usepackage{seqsplit}
\usepackage{tikz}
\usetikzlibrary{positioning,arrows.meta,calc,fit,backgrounds}
\usepackage[numbers,sort&compress]{natbib}
\usepackage[colorlinks=true,linkcolor=blue!50!black,citecolor=blue!50!black,urlcolor=blue!50!black]{hyperref}
\usepackage[capitalise,noabbrev]{cleveref}

\setlist{itemsep=1pt,topsep=3pt}
\providecommand{\tightlist}{\setlength{\itemsep}{1pt}\setlength{\parskip}{0pt}}
\newcommand{\kit}{\texttt{drydock-eval}}
\newcommand{\arm}[1]{\texttt{#1}}
\DeclareRobustCommand{\papersec}[1]{\ifcase#1\or\cref{sec:intro}\or\cref{sec:setting}\or\cref{sec:contain}\or
  \cref{sec:conform}\or\cref{sec:compare}\or\cref{sec:case}\or\cref{sec:related}\or\cref{sec:limits}\or\cref{sec:avail}\fi}
\DeclareRobustCommand{\papertab}[1]{\ifcase#1\or\cref{tab:versions}\or\cref{tab:checks}\or\cref{tab:simulated}\or
  \cref{tab:casearms}\fi}

\newcommand{\RunsTotal}{5{,}024}
\newcommand{\BlocksTotal}{31}
\newcommand{\ArmsTotal}{37}
\newcommand{\RunsThirtyOne}{710}
\newcommand{\RunsEFour}{4{,}314}
\newcommand{\RunsWorkstation}{4{,}082}
\newcommand{\ControlRows}{2{,}494}
\newcommand{\ServeRows}{72}
\newcommand{\CaseRows}{4{,}636}
\newcommand{\AllAgentRows}{9{,}660}
\newcommand{\ServeBigOne}{38.1}
\newcommand{\ServeBigLong}{35.5}
\newcommand{\ServeBigFourPer}{27.9}
\newcommand{\ServeBigFourAgg}{107}
\newcommand{\ServeMidOne}{76.1}
\newcommand{\ServeSmallOne}{105.8}
\newcommand{\ServeSmallSixteenAgg}{858}
\newcommand{\ServeBigSixteenAgg}{279}
\newcommand{\ServeBigPrefill}{1{,}539}
\newcommand{\ServeRatioSmallBig}{2.8}
\newcommand{\ServeBigStartup}{337}
\newcommand{\ServeBigKV}{46{,}352}
\newcommand{\ServeMidKV}{160{,}416}
\newcommand{\ServeSmallKV}{298{,}464}
\newcommand{\AdapterLossMin}{9}
\newcommand{\AdapterLossMax}{13}
\newcommand{\AdapterKV}{37{,}328}
\newcommand{\AdapterKVLoss}{19}
\newcommand{\AdapterBaseMin}{-1.1}
\newcommand{\AdapterBaseMax}{-0.9}

\newcommand{\PubUnionSix}{50}
\newcommand{\PubAlwaysSix}{25}
\newcommand{\PubNeverSix}{20}
\newcommand{\PubBetweenSix}{25}
\newcommand{\PubMinSeventy}{37}
\newcommand{\PubMaxSeventy}{44}
\newcommand{\PairAWthree}{16}
\newcommand{\PairASh}{20}
\newcommand{\PairAOnlySh}{4}
\newcommand{\PairAOnlyW}{0}
\newcommand{\PairAP}{0.125}
\newcommand{\PairBWthree}{22}
\newcommand{\PairBSh}{21}
\newcommand{\PairBOnlySh}{1}
\newcommand{\PairBOnlyW}{2}
\newcommand{\PairBP}{1.00}
\newcommand{\PairCBf}{20}
\newcommand{\PairCW}{19}
\newcommand{\PairCOnlyBf}{4}
\newcommand{\PairCOnlyW}{3}
\newcommand{\PairCP}{1.00}
\newcommand{\PairDBf}{17}
\newcommand{\PairDW}{20}
\newcommand{\PairDOnlyBf}{0}
\newcommand{\PairDOnlyW}{3}
\newcommand{\PairDP}{0.25}
\newcommand{\CoderOnlyFirst}{5}
\newcommand{\CoderOnlySecond}{1}
\newcommand{\CoderP}{0.22}
\newcommand{\MaskBfStd}{17}
\newcommand{\MaskBfMsk}{19}
\newcommand{\MaskShStd}{17}
\newcommand{\MaskShMsk}{17}
\newcommand{\LfourNFive}{10}
\newcommand{\LfourRealFive}{6}
\newcommand{\LfourEmuFive}{3}

\newcommand{\LfourWallDiffFive}{4}
\newcommand{\LfourNSix}{24}
\newcommand{\LfourRealSix}{14}
\newcommand{\LfourEmuSix}{17}

\newcommand{\LfourWallDiffSix}{8}
\newcommand{\HostedRatioMin}{0.22}
\newcommand{\HostedRatioMax}{0.34}
\newcommand{\TimingAgentMin}{179}
\newcommand{\TimingAgentMax}{233}
\newcommand{\CheckBase}{11}
\newcommand{\CheckCand}{13}
\newcommand{\CheckRuns}{32}
\newcommand{\CheckUp}{2}
\newcommand{\CheckDown}{0}
\newcommand{\CheckTime}{6}
\newcommand{\SmallPubUnion}{6}
\newcommand{\SmallPubWall}{117}
\newcommand{\SmallPubTools}{10}
\newcommand{\SmallPubAllThree}{4}
\newcommand{\SmallPubBigBF}{16}
\newcommand{\SmallPubBigSH}{19}
\newcommand{\KOneBaseBetter}{4}
\newcommand{\KOneOtherBetter}{1}
\newcommand{\KOneP}{0.38}
\newcommand{\KTwoRepos}{8}
\newcommand{\ParityMin}{2}
\newcommand{\ParityMax}{5}
\newcommand{\ParityWsTotal}{25}
\newcommand{\ParityWsRuns}{168}
\newcommand{\ParityWsPct}{14.9}
\newcommand{\ParityHostedTotal}{8}
\newcommand{\ParityHostedPct}{16.7}
\newcommand{\LeadOne}{+6.2}
\newcommand{\GainOne}{-0.35}
\newcommand{\GatePOne}{0.6034}

\newcommand{\LeadTwo}{+1.0}
\newcommand{\GainTwo}{-0.39}
\newcommand{\GatePTwo}{0.6034}

\newcommand{\LeadThree}{+5.2}
\newcommand{\GainThree}{+4.10}
\newcommand{\GatePThree}{0.0405}
\newcommand{\ConfBaseThree}{97}
\newcommand{\ConfCandThree}{118}

\newcommand{\RepeatSRange}{2.1}
\newcommand{\RepeatHFirst}{103}
\newcommand{\RepeatHSecond}{97}
\newcommand{\RepeatHDiff}{1.17}
\newcommand{\HNever}{83}
\newcommand{\HAlways}{10}
\newcommand{\HBetween}{35}
\newcommand{\HNeverPct}{65}
\newcommand{\NoiseSdRuns}{5.2}
\newcommand{\NoiseSdPts}{1.01}
\newcommand{\NoiseSdDiffPts}{1.44}

\newcommand{\HBasePct}{19.5}
\newcommand{\NoiseSdBinomPts}{1.75}
\newcommand{\NoiseShare}{34}
\newcommand{\CandThreeTasksUp}{24}
\newcommand{\CandThreeTasksDown}{11}
\newcommand{\CandThreeTasksSame}{93}
\newcommand{\CandThreeDiffRuns}{21}
\newcommand{\CandThreeNoiseRatio}{2.9}
\newcommand{\CandThreeDiffEarlier}{15}
\newcommand{\CandThreeNoiseRatioEarlier}{2.0}
\newcommand{\BaseSessionsDiffRuns}{6}
\newcommand{\BaseSessionsNoiseRatio}{0.8}
\newcommand{\ScreenSdArmRuns}{2.0}
\newcommand{\ScreenSdArmPts}{2.1}
\newcommand{\ScreenSdDiffPts}{2.9}
\newcommand{\ScreenNullLeadFive}{2.4}
\newcommand{\ScreenNullLeadTwo}{1.2}
\newcommand{\ScreenLeadRunsOne}{6}

\newcommand{\ScreenNullPOne}{0.09}
\newcommand{\ScreenLeadRunsTwo}{1}

\newcommand{\ScreenNullPTwo}{0.59}
\newcommand{\ScreenLeadRunsThree}{5}

\newcommand{\ScreenNullPThree}{0.10}
\newcommand{\HNeverAny}{76}
\newcommand{\HNeverRescued}{7}
\newcommand{\HNeverRescuedRuns}{10}
\newcommand{\HNeverCandRuns}{664}
\newcommand{\SNeverAny}{28}
\newcommand{\SNeverBase}{34}
\newcommand{\WsTasksAny}{263}
\newcommand{\WsTasksSolvedAny}{107}
\newcommand{\WsTasksNeverPct}{59}
\newcommand{\WsRunsPerTaskMin}{6}
\newcommand{\WsRunsPerTaskMax}{31}
\newcommand{\WsUnsolved}{3{,}212}
\newcommand{\WsUnsolvedNormalPct}{97}
\newcommand{\WsUnsolvedEmptyPct}{7}
\newcommand{\WsUnsolvedWrongPct}{93}
\newcommand{\WsUnsolvedTimeoutPct}{0.4}
\newcommand{\CommonTasks}{26}
\newcommand{\CommonBigRuns}{104}
\newcommand{\CommonBigSolved}{40}
\newcommand{\CommonBigPct}{38.5}
\newcommand{\CommonBigMean}{226}
\newcommand{\CommonBigPerSolved}{589}
\newcommand{\CommonSmallRuns}{156}
\newcommand{\CommonSmallSolved}{20}
\newcommand{\CommonSmallPct}{12.8}
\newcommand{\CommonSmallMean}{64}
\newcommand{\CommonSmallPerSolved}{496}
\newcommand{\CommonBigTasks}{11}
\newcommand{\CommonSmallTasks}{5}

\newcommand{\HostedSdTasks}{1.56}
\newcommand{\HostedSdDiffPts}{3.8}
\newcommand{\HostedDrawsNeeded}{10}
\newcommand{\LocalBlockRatio}{7}
\newcommand{\BaseWsRuns}{1{,}600}
\newcommand{\BaseWsSolved}{323}
\newcommand{\BaseWsPct}{20.2}
\newcommand{\BaseWsTasks}{248}
\newcommand{\BaseWsRepos}{22}
\newcommand{\RepoTopPct}{53}
\newcommand{\RepoZero}{3}
\newcommand{\RepoTopThreeShare}{33}
\newcommand{\TimeoutBigDevPct}{12.6}
\newcommand{\TimeoutBigBerthPct}{8.7}
\newcommand{\TimeoutSmallWsPct}{0.3}
\newcommand{\BudgetSmallWsPct}{1.2}

\newcommand{\CostBigMean}{278}
\newcommand{\CostSmallMean}{75}
\newcommand{\CostSmallSolvedT}{45}
\newcommand{\CostSmallUnsolvedT}{84}
\newcommand{\CostBigSolvedT}{231}
\newcommand{\CostBigUnsolvedT}{341}
\newcommand{\WsAgentHours}{85}
\newcommand{\WsTokensOutM}{33}
\newcommand{\WsTokensInM}{637}
\newcommand{\PoolTasks}{580}
\newcommand{\PoolEmptyPass}{1}
\newcommand{\PoolValid}{579}
\newcommand{\HostedCtrlRef}{107}
\newcommand{\HostedCtrlEmpty}{209}
\newcommand{\PoolRepos}{24}
\newcommand{\CtrlPairsMulti}{1{,}016}
\newcommand{\CtrlPairsDisagree}{0}
\newcommand{\PoolUsed}{479}
\newcommand{\PoolUntouched}{100}
\newcommand{\PoolUntouchedRepos}{6}
\newcommand{\PoolUntouchedTopRepo}{68}
\newcommand{\CaseOnlyTasks}{216}
\newcommand{\PoolUntouchedSmallRepos}{3}

\newcommand{\RecheckQuantities}{104}
\newcommand{\RecheckTimeOne}{1.026}
\newcommand{\RecheckTimeTwo}{1.182}
\newcommand{\RecheckTimeThree}{1.103}

\newcommand{\SLoadRows}{4{,}620}
\newcommand{\SLoadFlag}{5}
\newcommand{\SInfraNoRow}{8}

\newcommand{\STasksNever}{135}
\newcommand{\STasksAlways}{13}
\newcommand{\STasksMixed}{83}
\newcommand{\SHTasksDown}{34}
\newcommand{\SHTasksUp}{20}

\newcommand{\SCRejectedShare}{12.3}
\newcommand{\SHUnsolvedAtLimitShare}{11.5}
\newcommand{\SFUnsolvedAtLimitShare}{1.8}
\newcommand{\SFCompactedShare}{48}

\newcommand{\SHLooLo}{3.01}
\newcommand{\SHLooHi}{1.44}

\newcommand{\SMinRate}{21.8}
\newcommand{\SMaxRate}{24.6}
\newcommand{\SFirstStart}{2026-11-10 06:39}
\newcommand{\SLastEnd}{2026-11-11 03:56}
\newcommand{\SExposureMin}{85}
\newcommand{\SExposureMax}{90}

\newcommand{\SAgentRows}{4{,}620}
\newcommand{\SInfraRowsWritten}{16}
\newcommand{\SInfraRowsAll}{24}
\newcommand{\SJobsStarted}{4{,}644}
\newcommand{\STasks}{231}
\newcommand{\SRepos}{15}

\newcommand{\SChecksPassed}{4{,}620}
\newcommand{\SChecksFailed}{0}
\newcommand{\SLoadAbove}{276}
\newcommand{\SLoadShare}{6.0}
\newcommand{\SLoadThreshold}{240}

\newcommand{\SFTimeMean}{78}
\newcommand{\SFTimeMedian}{62}
\newcommand{\SHTimeMean}{137}
\newcommand{\SHTimeMedian}{106}

\newcommand{\SNoDifferMin}{172}
\newcommand{\SNoDifferMax}{183}
\newcommand{\SMinRepoP}{0.18}

\newcommand{\SHUnsolvedAtLimit}{106}
\newcommand{\SFUnsolvedAtLimit}{17}
\newcommand{\SCRejected}{114}

\newcommand{\SHTokOut}{14{,}397}
\newcommand{\SFTokOut}{7{,}690}
\newcommand{\SHTimeCapMissing}{-1.25}
\newcommand{\SHTimeCapMissingP}{0.34}
\newcommand{\SHd}{2.06}
\newcommand{\SHp}{0.11}
\newcommand{\SHHolm}{0.44}
\newcommand{\SHSimLo}{5.1}
\newcommand{\STSimHi}{3.5}

\newcommand{\SCChanged}{21}

\newcommand{\SRequests}{75{,}606}

\newcommand{\SReproMoveNinetyFive}{0.13}
\newcommand{\SReproMoveSim}{0.02}

\newcommand{\SReproHRegHi}{0.38}
\newcommand{\SReproHCmdHi}{0.25}
\newcommand{\SReproStepLo}{0.0501}
\newcommand{\SReproStepHi}{0.0509}
\newcommand{\SReproDraws}{99{,}999}
\newcommand{\SReproSeed}{20261109}

\newcommand{\SGateMinGain}{3.45}
\newcommand{\SGateAlpha}{0.025}
\newcommand{\SGateCount}{three times}
\newcommand{\SGateAlphaSplit}{0.0083}
\newcommand{\SGateOneTasks}{72}

\newcommand{\SGateOneRuns}{288}

\newcommand{\SGateOneBase}{63}
\newcommand{\SGateOneCand}{62}
\newcommand{\SGateOneGain}{${-}0.35$}
\newcommand{\SGateOneP}{0.60}

\newcommand{\SGateTwoTasks}{128}

\newcommand{\SGateTwoRuns}{512}
\newcommand{\SGateTwoSessions}{1{,}024}
\newcommand{\SGateTwoBase}{103}
\newcommand{\SGateTwoCand}{101}
\newcommand{\SGateTwoGain}{${-}0.39$}
\newcommand{\SGateTwoP}{0.60}

\newcommand{\SGateThreeTasks}{128}
\newcommand{\SGateThreeRepos}{19}
\newcommand{\SGateThreeRuns}{512}

\newcommand{\SGateThreeBase}{97}
\newcommand{\SGateThreeCand}{118}
\newcommand{\SGateThreeGain}{${+}4.10$}
\newcommand{\SGateThreeP}{0.04}

\newcommand{\SGateThreeReposUp}{9}
\newcommand{\SGateThreeReposDown}{6}
\newcommand{\SGateThreeLoo}{${+}2.94$}
\newcommand{\SGateThreeMore}{21}
\newcommand{\SGateThreeNet}{${+}5.25$}
\newcommand{\SGateThreeTopThree}{${+}4.25$}
\newcommand{\SGateThreeTopThreeTasks}{27}

\title{Contain, Conform, Compare (Extended Version):\\
A Red-Teamed Evaluation Kit for Small Coding Agents\\ and the Full Record of Its Experiments}

\author{Nguyen Minh Thang\\
\small VNU University of Science, 334 Nguyen Trai, Thanh Xuan, Hanoi, Vietnam}

\date{November 2026}

\begin{document}
\maketitle

\begin{abstract}
A hosted score in the Gemma 4 Developer Agent competition \citep{googledeepmind2026competition} is one noisy number, so
development needs a local evaluator that is fast, safe for agent-written code, and matching the hosted scorer's prompts.
\kit{} evaluates a small coding agent (Gemma 4 E4B \citep{gemmateam2026gemma4}) at about 250 runs per hour on one RTX
5090. \emph{Contain} runs each job in a fresh container without network and passed 200 scripted escape tests;
\emph{conform} provides eight checks that compare a local stack's requests with the scorer's configuration, down to
rendered tokens; \emph{compare} provides paired statistics that respect repository clusters. AI agents built and attacked
the kit; a human owner approved machine use and release. This extended version repeats the short paper and adds the full
record of the experiments made with and around the kit: \AllAgentRows{} rows of agent runs, \ControlRows{} grading
controls and \ServeRows{} serving measurements, all released. On four L4 GPUs the 31B model decodes \ServeBigOne{}
tokens per second in one stream and E4B \ServeSmallOne. Six configurations of a 31B agent solved between
\PubMinSeventy{} and \PubMaxSeventy{} of 70 development tasks, and no pairwise difference was detected. A pre-registered
five-arm experiment (231 tasks, 4,620 jobs) measured four prompt divergences: no E4B solve-rate change was detected
(largest estimate $-$2.1 points, 95\% interval $-$4.4 to +0.4). Three changes of the E4B agent led their screens by
\LeadTwo{} to \LeadOne{} points; each was then tested once on new tasks under a frozen pass rule, and none was
confirmed (gains \GainOne, \GainTwo{} and \GainThree{} points). The unchanged baseline, run again on the same 128 tasks,
moved by \RepeatHDiff{} points, and \HNeverPct\% of these tasks were solved in none of eight runs. Prompts and agent
bundles are withheld; every run-level result is public.
\end{abstract}

\section{Introduction}\label{sec:intro}

The competition scores an agent bundle on hidden GitHub-issue tasks, with Gemma 4 served by vLLM \citep{kwon2023vllm}.
Third-party analyses put the noise of one hosted score near 1.5 of 58 tasks \citep{afshar2026census,mobassir2026toosmall}.
With one submission per day, a hosted score cannot separate two agents that differ by a few tasks. Local evaluation must
therefore carry development; on a Kaggle kernel with four L4 GPUs our stack finished 35.2 agent runs per hour, serially.
A local evaluator has to answer three questions before its numbers mean anything.

\paragraph{Safety.} An agent run executes model-written shell commands and installs model-edited packages; our first
local setup ran them with no sandbox.

\paragraph{Fidelity.} We call a \emph{rendering trap} a configuration detail that silently changes the token sequence
the model receives while runs still finish and score. The \emph{scorer path} is the scorer's configuration reproduced
locally. In our first setup the served model name contained ``gemma4'', so the agent framework (google-adk 1.36.1) sent
tool results with role \texttt{tool\_responses} instead of \texttt{tool} \citep{adk2026toolresponses}. Prompts rebuilt
from its traces matched the server's token counts on 106 of 114 calls only when rendered with \texttt{tool\_responses}
(0 of 114 with \texttt{tool}). Solve counts hid it: that setup solved 3 and 5 of 24 tasks against 4 and 4 on Kaggle
(Fisher p 0.74 and 0.75).

These traps are public: the role rule \citep{adk2026toolresponses}, thought handling in history
\citep{google2026gemma4prompt,dimondev2026hf44,delta90002026vllm38488}, a newline after thinking text
\citep{gerganov2026hf135}, and the scorer's compaction settings \citep{holbrook2026starter}. We discovered none of them;
we contribute checks that detect them (\cref{sec:conform}) and a measurement of three of them plus a template swap
(\cref{sec:case}).

\paragraph{Inference.} Tasks from one repository succeed and fail together, a small model solves about one task in
five, and a development team looks at many variants. A difference between two arms then needs a paired design, an
analysis that respects repositories, and a rule that is fixed before the outcomes are read.

\paragraph{What this version adds.} This is the extended version of a paper of 3,000 words written for the
competition's paper track \citep{markowitz2026papertrack}. \Cref{sec:contain,sec:conform,sec:compare,sec:case-design}
repeat the text of that paper, and \cref{sec:case,sec:programmes} include the text of its supplement. Everything else is
new here. The short paper describes the kit and one experiment. It leaves out most of what was measured while the kit
was built and used: the throughput of the serving stack, the experiments with the 31B model that came before the work
on the small model, the first experiments with E4B, the runs that compared the workstation with the hosted notebook,
and the three candidate changes of the E4B agent that were taken through a screen and a confirmation. This version
reports all of them, with every run as a released row.

\paragraph{Contributions.}
\begin{itemize}
\item \textbf{A kit.} \kit{} has three modules. \emph{Contain} is a container runner for agent jobs together with the
record of the attacks that seven versions of it failed (\cref{sec:contain}). \emph{Conform} is a set of eight checks
that compare the requests of a local stack with the scorer path (\cref{sec:conform}). \emph{Compare} is a set of paired
tests with repositories as clusters and a pass rule whose behaviour was simulated before use (\cref{sec:compare}).
\item \textbf{A measurement of four prompt divergences} on E4B in a pre-registered experiment with five arms, 231 tasks
and 4,620 jobs (\cref{sec:case}).
\item \textbf{Serving measurements} of three Gemma 4 models on the hardware type of the scorer, including the cost of
serving a LoRA adapter (\cref{sec:serving}).
\item \textbf{The record of \RunsTotal{} further agent runs} in \BlocksTotal{} blocks and \ArmsTotal{} arms:
\RunsThirtyOne{} with the 31B model (\cref{sec:big}) and \RunsEFour{} with E4B (\cref{sec:small,sec:programmes}),
together with \ControlRows{} control gradings of the task pool.
\item \textbf{Three complete screen-and-confirm programmes} for a change of the E4B agent, each with its selection
step, its frozen rule and its outcome (\cref{sec:programmes}), and an analysis of what they say about noise and
selection (\cref{sec:analysis}).
\item \textbf{Data and scripts.} Every table and figure of \cref{sec:serving,sec:big,sec:small,sec:programmes,sec:analysis}
is computed by a released script from released files (\cref{sec:avail}).
\end{itemize}

\paragraph{Main findings.}
Most of the results are null results, and we report them as such.
\begin{itemize}
\item No attack on the runner escaped the container's namespaces or privileges, but nine demonstrated high-severity
vectors came through what the host shared with it, in designs that passed their own escape tests.
\item Four prompt divergences from the scorer path changed the E4B solve rate by less than the experiment could
detect; its simultaneous intervals leave room for a drop of up to \SHSimLo{} points and a gain of up to
\STSimHi{} points.
\item Six configurations of a 31B agent, from a public single agent to a wrapped agent with a shell-only coder, solved
\PubMinSeventy{} to \PubMaxSeventy{} of the same 70 tasks; no pairwise difference was detected, and
\PubBetweenSix{} of the 70 tasks were solved by some configurations and not by others.
\item Each of three E4B candidates led its screen. On new tasks the first two had estimated gains below zero. The third
gained \GainThree{} points, with a one-sided $p$ of \GatePThree{} against a required 0.025: not confirmed.
\item When the unchanged baseline ran again on the same 128 tasks, its total moved by \RepeatHDiff{} points.
\HNever{} of these tasks were solved in none of eight runs and \HAlways{} in all eight, so the remaining
\HBetween{} tasks carry all of the repeat-to-repeat noise.
\end{itemize}

\Cref{sec:setting} gives the setting and the terms. \Cref{sec:serving} reports the serving measurements.
\Cref{sec:contain,sec:conform,sec:compare} describe the three modules. \Cref{sec:big,sec:small} report the experiments
with the two models, \cref{sec:case} the case study, and \cref{sec:programmes} the three candidate programmes.
\Cref{sec:analysis} draws the experiments together; related work, limitations and availability follow.

\section{Setting, terms and rules}\label{sec:setting}

% Figure: where the three modules sit. Every label restates Sections 3-5; no number here is new.
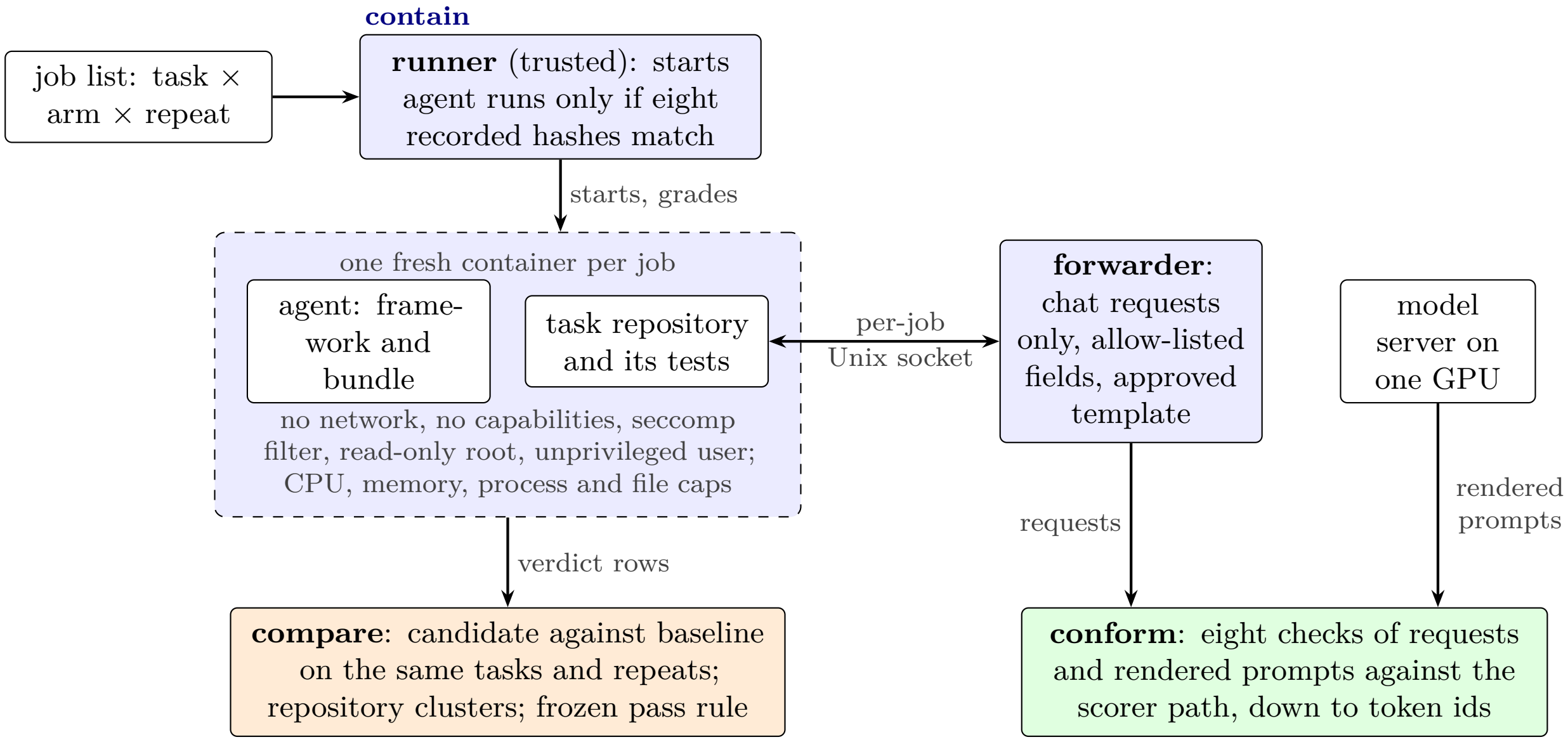
\begin{figure}[!tb]
\centering
\footnotesize
\begin{tikzpicture}[
  box/.style={draw, rounded corners=2pt, align=center, inner sep=4pt, text width=#1, minimum height=0.95cm},
  con/.style={box=#1, fill=blue!8},
  cfm/.style={box=#1, fill=green!12},
  cmp/.style={box=#1, fill=orange!16},
  plain/.style={box=#1, fill=white},
  arr/.style={-{Stealth[length=1.8mm]}, thick},
  darr/.style={{Stealth[length=1.8mm]}-{Stealth[length=1.8mm]}, thick},
  lab/.style={font=\scriptsize, align=center, inner sep=1.5pt, text=black!75}]
% row 1: what starts a job
\node[plain=2.5cm] (jobs)   at (1.2,0)    {job list: task $\times$ arm $\times$ repeat};
\node[con=3.9cm]   (runner) at (5.6,0)    {\textbf{runner} (trusted): starts agent runs only if eight recorded hashes match};
% row 2: the job
\node[lab, anchor=south] (ctrlab) at (5.05,-1.93) {one fresh container per job};
\node[plain=2.25cm] (agent) at (3.6,-2.55) {agent: framework and bundle};
\node[plain=2.25cm] (repo)  at (6.5,-2.55) {task repository and its tests};
\node[lab, anchor=north, text width=5.5cm] (restr) at (5.05,-3.2)
  {no network, no capabilities, seccomp filter, read-only root, unprivileged user; CPU, memory, process and file caps};
\begin{scope}[on background layer]
\node[draw, dashed, rounded corners=3pt, fill=blue!8, inner xsep=7pt, inner ysep=4pt, fit=(ctrlab)(agent)(repo)(restr)] (ctr) {};
\end{scope}
\node[con=2.45cm] (fwd) at (11.55,-2.55) {\textbf{forwarder}: chat requests only, allow-listed fields, approved template};
\node[plain=1.75cm] (srv) at (14.75,-2.55) {model server on one GPU};
% row 3: what reads the outputs
\node[cmp=5.5cm] (cmpn) at (5.05,-6.0)  {\textbf{compare}: candidate against baseline on the same tasks and repeats; repository clusters; frozen pass rule};
\node[cfm=5.2cm] (cfmn) at (13.15,-6.0) {\textbf{conform}: eight checks of requests and rendered prompts against the scorer path, down to token ids};
% arrows
\draw[arr] (jobs) -- (runner);
\draw[arr] (runner.south) -- node[lab, right, xshift=1pt] {starts, grades} (runner.south |- ctr.north);
\draw[darr] (repo.east) -- (fwd.west);
\path (ctr.east |- repo.east) -- node[lab, above] {per-job} node[lab, below] {Unix socket} (fwd.west);
\draw[arr] (ctr.south) -- node[lab, right, xshift=1pt] {verdict rows} (ctr.south |- cmpn.north);
\draw[arr] (fwd.south) -- node[lab, left, xshift=-1pt] {requests} (fwd.south |- cfmn.north);
\draw[arr] (srv.south) -- node[lab, right, xshift=1pt, text width=1.3cm] {rendered prompts} (srv.south |- cfmn.north);
% module tag
\node[lab, anchor=south west, text=blue!50!black] at ([yshift=1pt]runner.north west) {\textbf{contain}};
\end{tikzpicture}
\caption{The kit: where each module sits.}
\label{fig:overview}
\end{figure}

\subsection{The competition and its scorer}

A submission to the competition is an \emph{agent bundle}: a directory of configuration files for the google-adk agent
framework, which names the agents, their prompts, their tools, their sampling settings and their limits. The base
model is not part of the bundle: the scorer loads the bundle into the competition's harness, serves one Gemma 4 model
with vLLM on four L4 GPUs, and runs the agent on hidden tasks one after another. A task is an issue of a public Python
repository at a fixed commit. The agent works in a sandbox with the repository checked out, and ends by submitting a
patch. The task counts as solved if hidden tests pass after the patch is applied. The whole scoring run has a time limit
of 12 hours, so a slow agent loses tasks at the end. A team may submit once per day, and the public leaderboard shows the
solved share on 58 of the hidden tasks.

Two properties of this setting drive the rest of the paper. First, one hosted score is a draw with a standard
deviation near 1.5 tasks of 58 \citep{afshar2026census,mobassir2026toosmall}, and there is one draw per day. Second,
the scorer's configuration is public in a starter notebook \citep{holbrook2026starter}, but a local stack can differ
from it in ways that do not stop a run (\cref{sec:conform}).

\subsection{Models and software}

\paragraph{Stack.} Gemma 4 E4B, 4-bit quantization-aware checkpoint (Kaggle model version 2); vLLM 0.19.1; google-adk
1.36.1; the competition's harness packages. The experiments of \cref{sec:big} use the 31B model, also as a 4-bit
quantization-aware checkpoint. The serving measurements of \cref{sec:serving} add the 12B model. We write 31B, 12B and
E4B for the three models. E4B is one of the two models that are eligible for the small-model prize of the competition,
and it is the model of the kit's main use.

\subsection{Machines}

Agent runs were made on three kinds of machine, and every released row names its kind.

\paragraph{Emulated latency.} The first experiments with the 31B model ran on a rented cloud machine with one large
GPU. Its server answers faster than the scorer's four L4 GPUs, so each model call was padded to the duration that a
latency model predicts for the scorer's hardware: $0.15 + 0.032\,c + 0.0007\,n$ seconds for a call that generates $c$
tokens after $n$ new prompt tokens. We fitted this model to public call logs from real four-L4 hardware
\citep{moshegolderson2026freshswe}. Several jobs ran at the same time, and the load differed between
blocks; this is why arms are compared within a block wherever possible. \Cref{sec:big-real} compares the emulation with
real hardware.

\paragraph{Four L4 GPUs.} Kaggle notebooks with four L4 GPUs are the hardware type of the scorer. Jobs ran there with the
server flags of the starter notebook and without padding. Notebook hours are limited by a weekly quota, so
these blocks are small.

\paragraph{Workstation.} Experiments with E4B from \cref{sec:parity} on ran on a shared single-GPU workstation
(RTX 5090, 32 GB), with eight jobs at a time and each job in a container of the runner of \cref{sec:contain}. This is
the machine for which the kit was built; it finishes about 250 runs per hour.

\subsection{Tasks}

\paragraph{Development tasks.} The competition distributes development tasks with its harness. We used 70 of them,
from four repositories: a first set of 30 and 40 further tasks.
Competition data may not be redistributed, so the released rows name these tasks by opaque codes
(\texttt{dev-001} and so on) and give only their repository.

\paragraph{Berth.} Tasks for the work on E4B come from Berth \citep{patel2026berth}: 579 of its 653 tasks load in the
harness and pass our validation (the reference patch passes, an empty patch fails). The released rows keep the Berth
task identifiers. Four task sets were drawn from these 579 tasks: a screen set S of 48 tasks, a small panel C of 72 tasks in 18 repositories, a large panel of 200 tasks
in 20 repositories that contains C, and the case-study panel of 231 tasks in 15 repositories. We write H for the 128
tasks of the large panel that are outside C. S, the large panel and the case-study panel share no task. The two panels were frozen before any outcome on them
existed (\cref{sec:compare}). S contains the 24 tasks of the parity runs of \cref{sec:parity}, on which results of the
baseline were known before the screens ran; \cref{sec:programmes} comes back to this.

\subsection{Terms}

A \emph{run} is one agent job on one task: the harness prepares the sandbox, the agent works until it submits or is
stopped, and the harness grades the patch. A run is \emph{solved} or not; a run that ends without a patch or by a limit
counts as not solved. An \emph{arm} is one agent configuration, that is, one bundle with its settings. A \emph{block}
is one session on one machine in which one or more arms ran over one task list in a single job queue; a block with
several arms interleaves them, so they share the load and the time of day. A \emph{repeat} is one of several runs of
the same arm on the same task within a block. \emph{Points} are percentage points of the solve rate. A
\emph{task-equivalent} is one task solved in every repeat, or the same number of solved runs spread over tasks.
Arms carry neutral labels (\cref{tab:arms}); block labels name the experiment (\cref{tab:blocks}).

Counts of solved runs come with 95\% Wilson intervals where a single arm is described. Two arms on the same tasks
with one run per task are compared by the exact two-sided sign test on the tasks that only one of them solved. With
repeats, and for every decision about a candidate, the tests of \cref{sec:compare} are used. A null result is worded
``no difference was detected''; none of the experiments below was sized to show that two arms are equal.

\subsection{Threat model}

The adversary is the code the agent writes. It may try to read host files or other jobs, reach the network, exhaust
host resources, forge its verdict, or outlive its job. Kernel exploits are out of scope.

\subsection{Who did the work, and what is withheld}

\paragraph{Who did the work.} AI agents wrote the kit's code, attacks and analyses (builders). Separate AI agents
(verifiers) re-implemented or re-ran the work adversarially. A lead AI session froze the protocols; a human owner
approved isolation, machine time and release. ``Verified'' below means this process, not human review. The same holds
for the experiments that this version adds: AI agents built the arms, ran the blocks and wrote the analysis scripts
and this text, and the owner approved each hosted submission and each release.

\paragraph{What is withheld.} The competition is still running. The E4B agent bundle that serves as the baseline of
\cref{sec:small,sec:case,sec:programmes}, the changed bundles derived from it, their prompts, the fine-tuned adapter
of \cref{sec:small-first} and its training data are withheld for now; the release of the bundle and of the request
dumps of the case study is planned for 2026-12-03. The arms are therefore described by the kind of change and not by
their text (\cref{tab:arms}). What is released is the outcome and the measurements of every run. Two 31B arms are
public bundles run unchanged: \arm{BF5} is the public Budget-Fit V5 configuration, the most-run public
configuration in the hosted-score census of \citet{afshar2026census}, and \arm{ID24} is a bundle of the public collection of \citet{idris2026runs}.

\section{Serving measurements}\label{sec:serving}

How fast the scorer's server generates tokens decides how many model calls an agent can afford per task, and whether
more than one agent can work on a task at the same time. We measured it before designing any agent.

\paragraph{Method.} A Kaggle notebook with four L4 GPUs started the vLLM server with the flags of the starter notebook
\citep{holbrook2026starter}: the model split over the four GPUs, a maximum context of 32,768 tokens, and the Gemma 4
parsers. A client then sent streaming completion requests with prompts of random tokens. For each prompt length and
each number of concurrent streams, a first request generated one token, which measures the processing of the prompt
(prefill) and fills the prefix cache; a second request with the same prompt generated 160 tokens, which measures
decoding on a cached prefix. The 31B, 12B and E4B models were measured one after another, each with its own server.
A second session repeated the 31B measurement with a LoRA adapter loaded. The \ServeRows{} measured cells and the
server records are in \texttt{serving\_bench.csv} and \texttt{serving\_servers.csv}.

\begin{table}[t]
\centering
\small
\caption{Serving throughput on four L4 GPUs (vLLM, one server per model). Context: prompt tokens already in the request. Streams: concurrent requests. Each request generates 160 tokens.}
\label{tab:serving}
\begin{tabular}{lrrrrr}
\toprule
Model & Context & Streams & Prefill tok/s & Decode tok/s per stream & Decode tok/s, all streams \\
\midrule
31B & 2{,}048 & 1 & 1{,}539 & 38.1 & 38 \\
31B & 2{,}048 & 2 & 1{,}545 & 34.4 & 66 \\
31B & 2{,}048 & 4 & 1{,}546 & 31.2 & 122 \\
31B & 2{,}048 & 8 & 1{,}550 & 17.9 & 78 \\
31B & 2{,}048 & 16 & 1{,}548 & 18.2 & 279 \\
31B & 8{,}192 & 1 & 1{,}479 & 36.3 & 36 \\
31B & 8{,}192 & 2 & 1{,}480 & 33.7 & 66 \\
31B & 8{,}192 & 4 & 1{,}485 & 29.3 & 114 \\
31B & 8{,}192 & 8 & 1{,}486 & 9.2 & 25 \\
31B & 14{,}336 & 1 & 1{,}426 & 35.5 & 35 \\
31B & 14{,}336 & 2 & 1{,}427 & 32.8 & 64 \\
31B & 14{,}336 & 4 & 1{,}429 & 27.9 & 107 \\
\addlinespace
12B & 2{,}048 & 1 & 3{,}054 & 76.1 & 75 \\
12B & 2{,}048 & 4 & 3{,}076 & 60.6 & 235 \\
12B & 2{,}048 & 8 & 3{,}089 & 48.9 & 379 \\
12B & 2{,}048 & 16 & 3{,}082 & 34.6 & 534 \\
12B & 8{,}192 & 1 & 2{,}973 & 69.8 & 69 \\
12B & 8{,}192 & 2 & 2{,}994 & 63.2 & 123 \\
12B & 8{,}192 & 4 & 3{,}002 & 55.0 & 213 \\
\addlinespace
E4B & 2{,}048 & 1 & 5{,}267 & 105.8 & 104 \\
E4B & 2{,}048 & 4 & 5{,}385 & 85.8 & 329 \\
E4B & 2{,}048 & 8 & 5{,}372 & 75.9 & 578 \\
E4B & 2{,}048 & 16 & 5{,}404 & 57.4 & 858 \\
E4B & 8{,}192 & 1 & 5{,}212 & 95.9 & 94 \\
E4B & 8{,}192 & 2 & 5{,}235 & 85.0 & 164 \\
E4B & 8{,}192 & 4 & 5{,}254 & 73.5 & 282 \\
\bottomrule
\end{tabular}
\end{table}

\begin{figure}[t]
\centering
\includegraphics[width=\linewidth]{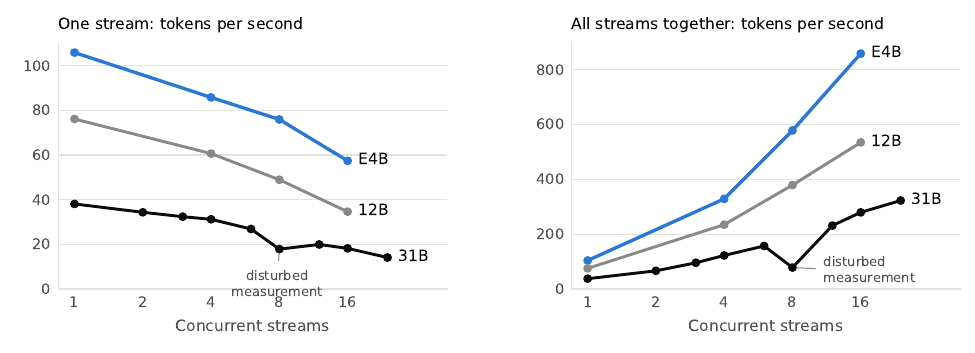}
\caption{Decode throughput on four L4 GPUs with a context of 2,048 tokens. Left: tokens per second of one stream as
the number of concurrent streams grows. Right: tokens per second summed over the streams.}
\label{fig:serving}
\end{figure}

\paragraph{One stream.} The 31B model decodes \ServeBigOne{} tokens per second with a short context and
\ServeBigLong{} with 14,336 tokens of context (\cref{tab:serving}). The 12B model reaches \ServeMidOne{} and E4B
\ServeSmallOne{} tokens per second, \ServeRatioSmallBig{} times the 31B model. Prefill is fast in comparison: about
\ServeBigPrefill{} tokens per second for 31B. An agent that keeps its prompt prefix stable pays this once, because the
server reuses cached prefixes. Generated tokens are what costs time: a 31B call that writes 1,000 tokens takes about
half a minute. The server needs \ServeBigStartup{}~s to start with the 31B model, which counts against the time limit
of a scoring run.

\paragraph{Several streams.} Concurrent streams share the GPUs (\cref{fig:serving}). With four streams and a long
context, each 31B stream still decodes \ServeBigFourPer{} tokens per second, \ServeBigFourAgg{} in total. Apart
from one cell discussed below, the total grows up to 16 streams for all three models, to \ServeBigSixteenAgg{} tokens per second for 31B and
\ServeSmallSixteenAgg{} for E4B at 2,048 tokens of context. The scorer runs tasks one after another, so this capacity
is only useful to an agent that runs several model calls in parallel within one task.

\paragraph{Two cells that do not fit.} With eight streams of 8,192 tokens, the 31B server delivered 9.2 tokens per
second per stream, a third of the four-stream value. The server reported a key-value cache of \ServeBigKV{} tokens
for 31B, against \ServeMidKV{} for 12B and \ServeSmallKV{} for E4B, and eight contexts of 8,192 tokens hold 65,536
tokens. We read the drop as cached prefixes being evicted and processed again; we did not test this reading. With
eight streams of 2,048 tokens the 31B server gave 17.9 tokens per second per stream and 78 in total, less in total
than with four streams (122) and with 16 (\ServeBigSixteenAgg). We have no explanation for this cell and report it as measured;
\cref{fig:serving} marks it.

\begin{table}[t]
\centering
\small
\caption{Cost of serving a LoRA adapter with the 31B model (same server settings; requests addressed to the adapter). Decode speed in tokens per second per stream; change against the control in brackets.}
\label{tab:adapter}
\begin{tabular}{lrrrr}
\toprule
Server & KV cache, tokens & 2k context, 1 stream & 8k, 1 stream & 8k, 4 streams \\
\midrule
no adapter (control) & 46{,}352 & 38.3 & 36.4 & 29.4 \\
rank 16, attention projections & 37{,}328 & 33.9 (\ensuremath{-}12\%) & 32.4 (\ensuremath{-}11\%) & 26.9 (\ensuremath{-}9\%) \\
rank 16, all linear layers & 37{,}328 & 33.8 (\ensuremath{-}12\%) & 32.5 (\ensuremath{-}11\%) & 26.9 (\ensuremath{-}9\%) \\
rank 8, two module types & 37{,}568 & 33.3 (\ensuremath{-}13\%) & 32.1 (\ensuremath{-}12\%) & 26.7 (\ensuremath{-}9\%) \\
\bottomrule
\end{tabular}
\end{table}

\paragraph{Cost of an adapter.} The competition allows LoRA adapters on the one served model. With an adapter of rank
16 or rank 8 loaded, requests addressed to the adapter decoded \AdapterLossMin{} to \AdapterLossMax\% slower than
requests to the same server without an adapter (\cref{tab:adapter}), and the reported key-value cache shrank by
\AdapterKVLoss\% to \AdapterKV{} tokens. Requests addressed to the base model on a server with a loaded adapter
changed by \AdapterBaseMin{} to \AdapterBaseMax\% (one stream, short context). Which layers the adapter covers made no visible difference. The
measurements were made once per cell.

\paragraph{What followed for the design.} At 35 tokens per second, an agent with a limit of eight minutes per task,
the limit of the public configuration \arm{BF5}, can generate at most about 17,000 tokens, before the time of its tool
calls and of prefill is taken off. A second agent that works in parallel
slows the first by about a tenth at two streams. E4B generates \ServeRatioSmallBig{} times as fast, so the same task
limit is rarely what stops an E4B agent (\cref{sec:cost}).

\section{Contain: a runner and the attacks it failed}\label{sec:contain}

% written by scripts/reuse_short.py from the sources of the short paper; do not edit
In the final design each job gets one fresh container: no network, no capabilities, a seccomp filter, a read-only root, an unprivileged user, 2 CPUs, 10 GiB of memory, 4,096 processes, and temporary filesystems capped in bytes and inodes. The only request channel is a per-job Unix socket to a forwarder, which passes only chat-completion requests with allow-listed fields, text-only content and an approved template hash, under rate and size caps. Agent runs start only if eight recorded hashes (image, code, flags, model files) still match.

A \emph{canary} is a scripted attack with a machine-checked outcome.

\begin{table}[!tb]
\centering
\caption{Runner versions.}\label{tab:versions}
\footnotesize
\begin{tabular}{@{}
  >{\raggedright\arraybackslash}p{(\linewidth - 4\tabcolsep) * \real{0.2000}}
  >{\raggedright\arraybackslash}p{(\linewidth - 4\tabcolsep) * \real{0.1900}}
  >{\raggedright\arraybackslash}p{(\linewidth - 4\tabcolsep) * \real{0.6100}}@{}}
\toprule
\begin{minipage}[b]{\linewidth}\raggedright
Runner version
\end{minipage} & \begin{minipage}[b]{\linewidth}\raggedright
Canaries passed
\end{minipage} & \begin{minipage}[b]{\linewidth}\raggedright
Security findings of independent verifiers
\end{minipage} \\
\midrule
1. Container per command & 52/52 & 4 high: host-side install ran the agent's \texttt{setup.py} (builder-found); symlink leaked a host file; 10 GiB of host disk reserved in 0.005 s; host copied a sparse file in full \\
2. Hardened, per command & 60/60 (57/60 before canary fixes) & 4 high, 1 untested: host overwrote a file through a planted link; cleanup followed links; 6 GiB reserved in 0.012 s; 200,000 files in 2.4 s \\
3. Container per job & 98/98 & 1 high: the model server could fetch URLs named in requests; 2 medium \\
4--6. Request filter added, tightened & 134/134, 159/159, 191/191 & medium at most \\
7. Final & 200/200 & no high or medium \\
\bottomrule
\end{tabular}
\end{table}

\begin{figure}[!tb]
\centering
\includegraphics[width=\linewidth]{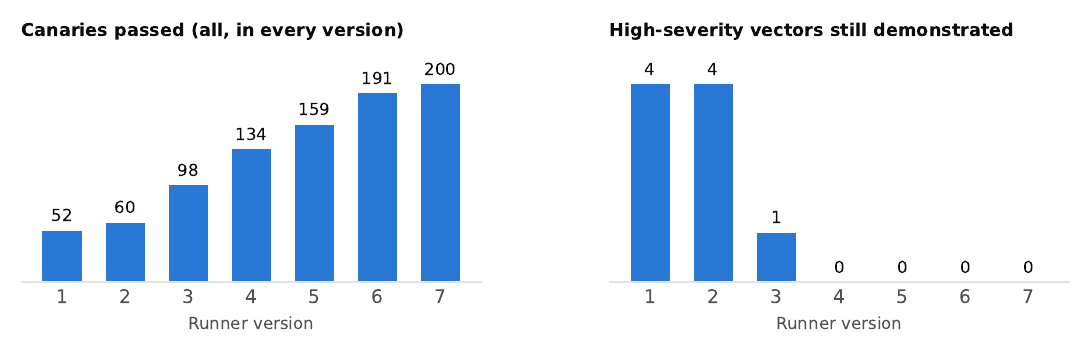}
\caption{Canaries passed and high-severity vectors found, per runner version.}
\label{fig:versions}
\end{figure}

\paragraph{Passing canaries proved little.} No attack escaped the container's namespaces or privileges in any round. Nine demonstrated high-severity vectors came through what the host shared with it: host-side steps that touched agent-written files, resource limits (a known class \citep{gao2019houdini}) and the model server.

\paragraph{Final version.} Controls on 481 tasks (reference patch 481/481, empty patch 0/481) showed no verdict, exit-code or test-summary difference from the previous version. After the runner was killed, the container was gone within 149 s with no further model request. The final runner completed a 24-job list at 251--254 runs per hour with 8 workers.

\paragraph{What remains.} Containers share the host kernel; user namespaces are not remapped; the request parser runs in the trusted runner; a job can slow others through the shared model server within its caps. Our claim is conditional: against the stated adversary and with the eight hashes matching, no tested attack succeeded.

\section{Conform: eight checks against the scorer path}\label{sec:conform}

% written by scripts/reuse_short.py from the sources of the short paper; do not edit
\texttt{conform} reads the requests a local stack sent and, where logged, the prompts the server rendered. Reference values come from the public starter notebook \citep{holbrook2026starter}, our Kaggle-kernel traces, and the Kaggle model's chat template.

\begin{table}[!tb]
\centering
\caption{The eight checks.}\label{tab:checks}
\footnotesize
\begin{tabular}{@{}
  >{\raggedright\arraybackslash}p{(\linewidth - 4\tabcolsep) * \real{0.1300}}
  >{\raggedright\arraybackslash}p{(\linewidth - 4\tabcolsep) * \real{0.5200}}
  >{\raggedright\arraybackslash}p{(\linewidth - 4\tabcolsep) * \real{0.3500}}@{}}
\toprule
\begin{minipage}[b]{\linewidth}\raggedright
Check
\end{minipage} & \begin{minipage}[b]{\linewidth}\raggedright
Compares
\end{minipage} & \begin{minipage}[b]{\linewidth}\raggedright
Mutation runs caught
\end{minipage} \\
\midrule
role & role of tool results & 53/53 \\
sampling & temperature, top-p, token and thinking budgets & 52/52 \\
tools & declared tool schemas & 20/20 \\
shape & message skeleton, system text, field encoding & 72/72 \\
template & rendered prompt against the Kaggle template & 13/21 by token count, 24/24 by token id \\
thoughts & history thoughts, under the key the server reads & 52/52 \\
newline & newline before the token closing a thought & 12/12 \\
compaction & when history is summarized (14,336 tokens \citep{holbrook2026starter}) & 19/19 \\
\bottomrule
\end{tabular}
\end{table}

\paragraph{Role.} From each google-adk release's source (only 1.36.1 was executed): 1.34.0--1.39.1 and 2.0--2.4 send \texttt{tool\_responses} when the model string contains ``gemma4''; 2.5--2.11 when it matches \texttt{gemma-?4}; 2.12--2.14 exempt three OpenAI-compatible providers \citep{adk2026toolresponses}. The harness packages require \texttt{\textgreater{}=1.34.0,\textless{}2.0}.

\paragraph{Thoughts.} vLLM 0.19.1 reads an earlier assistant thought only from the field \texttt{reasoning}. google-adk 1.36.1 writes \texttt{reasoning\_content}, which that server drops silently \citep{delta90002026vllm38488}; the competition harness writes both. Removing \texttt{reasoning} in a probe lost 55 of 55 thoughts from the rendered prompt.

\paragraph{Template.} The check re-renders each request outside vLLM with the Kaggle template and compares token counts with the server's, or token ids when the server logs them, because identical strings can tokenize differently \citep{agl2025retokenization}.

\paragraph{Evidence.} In version 2's 24-task parity run, all 2,566 tool results had role \texttt{tool}, and re-rendering reproduced the server's token count on 396 of 396 requests (with \texttt{tool\_responses}: 0 of 358 multi-turn requests). With the final runner, the checker reproduced the token ids of 243 of 243 returned calls, and a renderer written separately by the verifier matched 50 of 50 scorer-path requests. Prompts rebuilt from early calls of our Kaggle-kernel traces match the recorded counts on 181 of 189 calls with role \texttt{tool} and thoughts kept, 1 of 189 with \texttt{tool\_responses} and 0 of 189 with thoughts dropped: our only Kaggle-side evidence, and count-level.

\paragraph{Mutation suite.} 103 mutations (85 drifts, 18 benign controls) in four scripted scenarios against a mock server give 412 runs. The intended check caught 317 of 325 applicable drift runs; the 8 misses are same-length template changes seen through counts only. No check failed in 72 control runs (24 warned).

\section{Compare: a decision rule that respects repositories}\label{sec:compare}

% written by scripts/reuse_short.py from the sources of the short paper; do not edit
Tasks from one repository succeed and fail together. Clustered inference for model evaluations is prior art \citep{miller2024errorbars}; \texttt{compare} packages a small-cluster version with a frozen pass rule and its simulated behaviour.

Candidate and baseline run on the same tasks and repeats. A candidate passes only if (1) the mean gain is at least 2/58 (3.45 points, two leaderboard tasks); (2) one-sided p \ensuremath{\le} 0.025 under a null-imposed wild cluster bootstrap-t over repositories, with Rademacher weights and 9,999 draws \citep{cameron2008bootstrap,mackinnon2017wild}; (3) at least three repositories gain and every leave-one-repository-out estimate is positive; (4) agent time stays within 1.2 times baseline and infrastructure errors do not rise; and (5) no repository loses two task-equivalents or more. Two panels were frozen before any outcome: a small one (72 tasks, 18 repositories) and a large one (200 tasks, 20 repositories).

\begin{table}[!tb]
\centering
\caption{Simulated pass rates (3,000 simulations per cell; 400 for the dropping row).}\label{tab:simulated}
\footnotesize
\begin{tabular}{@{}
  >{\raggedright\arraybackslash}p{(\linewidth - 4\tabcolsep) * \real{0.5800}}
  >{\raggedright\arraybackslash}p{(\linewidth - 4\tabcolsep) * \real{0.2100}}
  >{\raggedright\arraybackslash}p{(\linewidth - 4\tabcolsep) * \real{0.2100}}@{}}
\toprule
\begin{minipage}[b]{\linewidth}\raggedright
Simulated scenario
\end{minipage} & \begin{minipage}[b]{\linewidth}\raggedright
Small panel
\end{minipage} & \begin{minipage}[b]{\linewidth}\raggedright
Large panel
\end{minipage} \\
\midrule
No effect, exchangeable tasks & 0.23--1.90\% & 0.00--0.50\% \\
No mean effect, candidate breaks 20\% of repositories & 7.50\% & 8.43\% \\
No effect, worst tasks dropped after seeing outcomes & 31.75\% & 40.75\% \\
Even gain of 3/58 (power) & \ensuremath{\approx}60\% & \ensuremath{\approx}90\% \\
Gain confined to four repositories (power) & 15.9\% & 24.9\% \\
\bottomrule
\end{tabular}
\end{table}

The 2.5\% level holds for exchangeable effects, not for skewed ones, and not if tasks are dropped after unblinding. Independent re-implementations matched the rule code on 280 of 280 clean datasets (first version) and, with rule readings aligned, 193 of 194 (amended). Applied three times, to screen leaders from one separate 48-task set (25, 22, 25 of 96 runs solved; baselines 19, 21, 20); none passed. Gains: \ensuremath{-}0.35 points on the small panel (p = 0.60); on the 128 large-panel tasks outside it, \ensuremath{-}0.39 (p = 0.60), then +4.10 (p = 0.04 \textgreater{} 0.025, uncorrected). Condition (5): first application only.

The three applications are reported in full in \cref{sec:programmes}, with the screens that selected their candidates.

\section{Experiments with the 31B model}\label{sec:big}

The work began with the 31B model, before the kit existed. The question was whether a different agent architecture
solves more tasks than the public single-agent configuration within the scorer's time limit. The answer of \RunsThirtyOne{}
runs is that no difference was detected. These experiments also produced the habits that the kit later fixed in code:
compare arms inside one block, keep task sets apart, and distrust a lead that was found by looking at many variants.

\subsection{Arms}\label{sec:big-arms}

\Cref{tab:arms} lists every arm. The 31B arms fall into four families.

\paragraph{Reference.} \arm{BF5} is the public Budget-Fit V5 configuration, run unchanged: one agent with the
tools of the harness, a limit of eight minutes and 28 tool calls per task, and the history compaction of the starter
configuration, which replaces older turns by a summary when the history grows. \arm{BF5-p}, \arm{BF5-c} and
\arm{BF5-s} change one thing each: a section of the task prompt, a checking step before the patch is submitted, and
the sampling settings.

\paragraph{Wrapped.} In \arm{W1} a root agent hands the task to a coder agent through the agent-as-tool mechanism of
the framework. The coder runs with its own history, and the harness does not summarise it (column Sum.\ of
\cref{tab:pubarms}). The first runs showed a defect: when the coder ended after a malformed tool call, the root
submitted an empty patch. \arm{W2} lets the root restart the coder a bounded number of times, and \arm{W3} leaves the
root with nothing else to do: it can pass the request on and restart the coder, and it cannot submit. \arm{W1-b} and
\arm{W1-p} change limits and the coder prompt.

\paragraph{Shell only.} In the \arm{SH} arms the coder works through a shell tool alone and edits files with shell commands.
\arm{SH1}, \arm{SH3} and \arm{SH5} are successive versions of its prompt and limits; \arm{SH6} adds an analysis step
to the prompt of \arm{SH5}.

\paragraph{A second public bundle.} \arm{ID24} is a public bundle \citep{idris2026runs} run unchanged, and
\arm{ID24-r} a variant with one tool removed.

\begin{table}[t]
\centering
\footnotesize
\caption{The 31B arms on the development tasks, one run per task, with emulated L4 latency. Top: the first 30 tasks. Middle: 40 further tasks. Bottom: all 70. Wall: mean seconds per job. Out: mean generated tokens. T/O: runs stopped by the session time limit. Empty: runs that left no patch. Sum.: runs with at least one history summary.}
\label{tab:pubarms}
\begin{tabular}{llrrrrrrrrrr}
\toprule
Block & Arm & Tasks & Solved & \% & 95\% interval & Wall & Out & Calls & T/O & Empty & Sum. \\
\midrule
\texttt{pub31-1} & \texttt{BF5} & 30 & 17 & 57 & 39--73 & 303 & 6{,}893 & 24.2 & 5 & 1 & 18 \\
\texttt{pub31-1} & \texttt{BF5-p} & 30 & 18 & 60 & 42--75 & 296 & 6{,}921 & 23.1 & 4 & 4 & 19 \\
\texttt{pub31-1} & \texttt{W1} & 30 & 20 & 67 & 49--81 & 254 & 5{,}942 & 22.4 & 3 & 4 & 0 \\
\texttt{pub31-2} & \texttt{W1-b} & 30 & 17 & 57 & 39--73 & 290 & 6{,}394 & 23.4 & 7 & 7 & 0 \\
\texttt{pub31-3} & \texttt{W1-p} & 30 & 15 & 50 & 33--67 & 304 & 6{,}412 & 25.2 & 11 & 12 & 0 \\
\texttt{pub31-4} & \texttt{W3} & 30 & 16 & 53 & 36--70 & 272 & 7{,}448 & 30.6 & 4 & 6 & 0 \\
\texttt{pub31-4} & \texttt{SH1} & 30 & 20 & 67 & 49--81 & 237 & 6{,}739 & 24.5 & 1 & 2 & 0 \\
\addlinespace
\texttt{pub31-5} & \texttt{BF5} & 40 & 20 & 50 & 35--65 & 338 & 7{,}032 & 23.3 & 8 & 2 & 29 \\
\texttt{pub31-5} & \texttt{W2} & 40 & 19 & 48 & 33--63 & 312 & 6{,}582 & 24.4 & 7 & 6 & 0 \\
\texttt{pub31-6} & \texttt{W3} & 40 & 22 & 55 & 40--69 & 260 & 7{,}399 & 25.6 & 2 & 6 & 0 \\
\texttt{pub31-6} & \texttt{SH3} & 40 & 21 & 52 & 37--67 & 223 & 6{,}109 & 24.4 & 3 & 7 & 0 \\
\addlinespace
\texttt{pub31-7} & \texttt{SH5} & 70 & 41 & 59 & 47--69 & 272 & 7{,}821 & 27.0 & 6 & 7 & 0 \\
\texttt{pub31-8} & \texttt{SH6} & 70 & 44 & 63 & 51--73 & 280 & 7{,}978 & 25.6 & 7 & 8 & 0 \\
\bottomrule
\end{tabular}
\end{table}

\subsection{The first 30 tasks}

Block \texttt{pub31-1} ran three arms in one queue on 30 development tasks (\cref{tab:pubarms}). The wrapped arm
\arm{W1} solved \PairDW{} tasks and \arm{BF5} \PairDBf: \PairDOnlyW{} tasks were solved by \arm{W1} only and
\PairDOnlyBf{} by \arm{BF5} only (sign test $p = \PairDP$). \arm{W1} was also faster, 254 against 303 seconds per job,
because none of its runs spent time on history summaries, while 18 of the 30 \arm{BF5} runs wrote at least one.
A lead of three tasks on 30 is what one hopes to see and cannot trust.

Two attempts to improve \arm{W1} by changing its limits and its prompt went the other way. \arm{W1-b}, with a shorter
time limit and more tool calls, solved 17, and \arm{W1-p}, with a rewritten coder prompt and a shorter time limit,
solved 15, with 11 runs stopped by the time limit and 12 without a patch. Both ran in blocks of their own, under a load
that differed from that of the first block, so the comparison with \arm{W1} mixes the change with the session. What can
be said is that neither shorter limits nor the rewritten prompt gave more solved tasks.

\Cref{tab:pubarms} also shows \arm{W3} and the first shell-only arm in one block. \arm{SH1} solved \PairASh{} and
\arm{W3} \PairAWthree{} (\PairAOnlySh{} tasks only \arm{SH1}, \PairAOnlyW{} only \arm{W3}; $p = \PairAP$). \arm{W3} has
the coder of \arm{W1}. In their two blocks \arm{W3} solved \PairAWthree{} and \arm{W1} \PairDW{} of the same 30 tasks:
\CoderOnlyFirst{} tasks only \arm{W1}, \CoderOnlySecond{} only \arm{W3} ($p = \CoderP$). The data cannot say whether
the changed root or the session accounts for this.

\subsection{Forty further tasks and the total of seventy}

The leads of the first 30 tasks were tested on 40 further tasks. \arm{BF5} solved \PairCBf{} and the wrapped arm
\arm{W2} \PairCW{} (\PairCOnlyBf{} and \PairCOnlyW{} tasks solved by one arm only; $p = \PairCP$). \arm{W3} solved
\PairBWthree{} and the shell-only arm \arm{SH3} \PairBSh{} (\PairBOnlyW{} and \PairBOnlySh; $p = \PairBP$). Neither
lead of the first 30 tasks appeared again, although both leading arms had been revised in between.

\begin{table}[t]
\centering
\small
\caption{Six configurations on all 70 development tasks (one run per task). The last three columns compare each configuration with \texttt{BF5} on the same tasks: tasks only it solved, tasks only \texttt{BF5} solved, and the two-sided exact sign test.}
\label{tab:pubseventy}
\begin{tabular}{lrrrrrrrr}
\toprule
Configuration & First 30 & Next 40 & All 70 & \% & 95\% interval & Only it & Only \texttt{BF5} & $p$ \\
\midrule
\texttt{BF5} & 17 & 20 & 37 & 53 & 41--64 & --- & --- & --- \\
\texttt{W1/W2} & 20 & 19 & 39 & 56 & 44--67 & 6 & 4 & 0.75 \\
\texttt{W3} & 16 & 22 & 38 & 54 & 43--65 & 6 & 5 & 1.00 \\
\texttt{SH1/SH3} & 20 & 21 & 41 & 59 & 47--69 & 8 & 4 & 0.39 \\
\texttt{SH5} & 17 & 24 & 41 & 59 & 47--69 & 7 & 3 & 0.34 \\
\texttt{SH6} & 21 & 23 & 44 & 63 & 51--73 & 10 & 3 & 0.09 \\
\bottomrule
\end{tabular}
\end{table}

\begin{figure}[t]
\centering
\includegraphics[width=\linewidth]{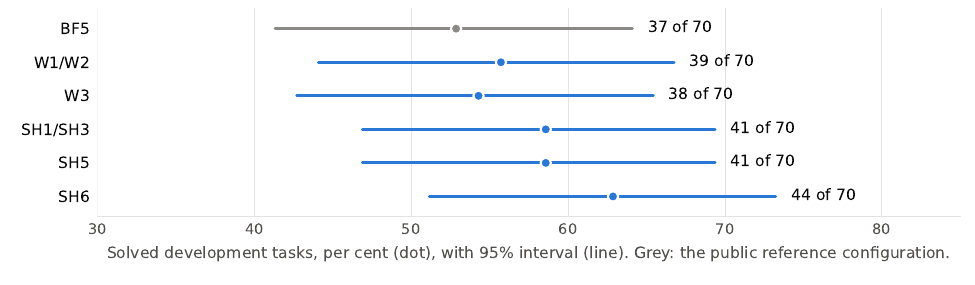}
\caption{Six configurations of the 31B agent on the same 70 development tasks, one run per task.}
\label{fig:pub-configs}
\end{figure}

\begin{figure}[t]
\centering
\includegraphics[width=\linewidth]{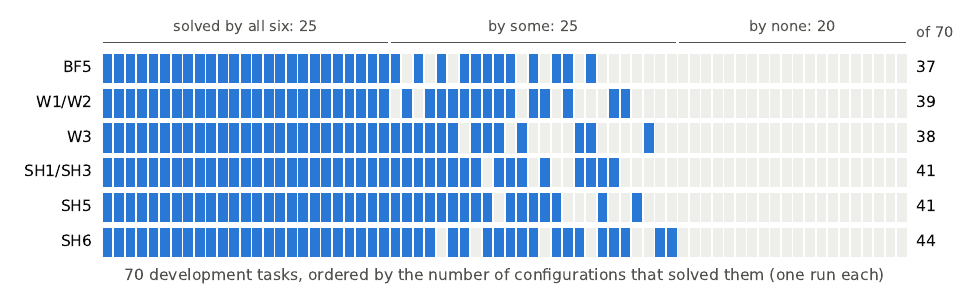}
\caption{Which configuration solved which task. One column per development task, one row per configuration; a filled
cell is a solved run. \PubAlwaysSix{} tasks were solved by all six configurations and \PubNeverSix{} by none.}
\label{fig:pub-matrix}
\end{figure}

\Cref{tab:pubseventy} joins the blocks into six configurations that each ran once on all 70 tasks. Two of them join
two versions of an arm: \arm{W1} on the first 30 tasks with \arm{W2} on the next 40, and \arm{SH1} with \arm{SH3}.
The six totals lie between \PubMinSeventy{} and \PubMaxSeventy{} of 70 (\cref{fig:pub-configs}). Against \arm{BF5},
the largest difference is that of \arm{SH6}, with 10 tasks that only it solved and 3 that only \arm{BF5} solved
($p = 0.09$, not corrected for the five comparisons). No difference was detected. \arm{SH6} is also the last of a series of
versions that were developed while looking at results on these same tasks, so its count is the one most exposed to
selection.

\Cref{fig:pub-matrix} shows where the configurations differ. \PubAlwaysSix{} tasks were solved by every configuration
and \PubNeverSix{} by none; on the remaining \PubBetweenSix{} tasks the outcome depends on the configuration or on the
run. The union of the six is \PubUnionSix{} tasks, against \PubMaxSeventy{} for the best single one. This gap is an
upper bound on what a choice between configurations could gain, and it is inflated: a task that one configuration
solves in half of its runs enters the union easily when six runs try it. With one run per task and configuration
the two sources cannot be separated. The experiments with E4B below therefore use repeats.

\subsection{Masked test runners}

The harness has an option that masks the test runners inside the sandbox, so that the agent cannot run the tests of
the repository. We did not know whether the hosted scorer sets it, so two arms were run on the first 30 tasks with the
runners masked (\cref{tab:masked}). \arm{BF5} solved \MaskBfMsk{} with masked runners against \MaskBfStd{} without,
and \arm{SH5} \MaskShMsk{} against \MaskShStd. The standard and the masked runs come from different blocks. No drop
was detected, and nothing in these counts suggests that the agents depend on running the tests.

\begin{table}[t]
\centering
\small
\caption{Runs with the test runners masked inside the sandbox, on the first 30 development tasks (solved of 30).}
\label{tab:masked}
\begin{tabular}{lrrrr}
\toprule
Arm & Standard & Masked & Only masked & Only standard \\
\midrule
\texttt{BF5} & 17 & 19 & 2 & 0 \\
\texttt{SH5} & 17 & 17 & 2 & 2 \\
\bottomrule
\end{tabular}
\end{table}

\subsection{Real hardware against emulated latency}\label{sec:big-real}

Two blocks on a Kaggle notebook with four L4 GPUs repeated an arm on tasks that it had run under emulated latency
(\cref{tab:lfour}). Mean time per job agreed within \LfourWallDiffFive{} and \LfourWallDiffSix\%, with the emulated
runs the slower ones, so the emulation did not make the time limit easier than it is on real hardware.

\begin{table}[t]
\centering
\footnotesize
\caption{The same arm and tasks on real 4$\times$L4 hardware (serial jobs) and under latency emulation (eight workers). Wall: mean seconds per job.}
\label{tab:lfour}
\begin{tabular}{lrrrrrrrrr}
\toprule
Arm & Tasks & Solved, L4 & Solved, emul. & Both & Only L4 & Only emul. & Wall, L4 & Wall, emul. & Diff. \\
\midrule
\texttt{SH5} & 10 & 6 & 3 & 3 & 3 & 0 & 280 & 292 & \ensuremath{-}4\% \\
\texttt{SH6} & 24 & 14 & 17 & 14 & 0 & 3 & 289 & 314 & \ensuremath{-}8\% \\
\bottomrule
\end{tabular}
\end{table}

The solved counts moved in both directions. \arm{SH5} solved \LfourRealFive{} of \LfourNFive{} tasks on real hardware
and \LfourEmuFive{} under emulation. These ten tasks had been picked as a hard mix, and three of them changed outcome
with identical agent files. \arm{SH6} solved \LfourRealSix{} of \LfourNSix{} on real hardware and \LfourEmuSix{}
under emulation. The \LfourNSix{} tasks of this second block were chosen with the earlier results of the arms in view,
which favours the emulated count; a fall on repetition is what such a choice predicts, and it is no evidence of a
difference between the machines. Neither block can support a claim about the solve rate. Together they show that a
single run per task moves by several tasks in ten to twenty-four when nothing changes.

\subsection{Berth tasks on real hardware}

Two later blocks ran 31B arms on Berth tasks on four L4 GPUs with the software stack of the scorer. The first is a
timing block with one job at a time (\cref{tab:timing}): the two public bundles and one variant of each on ten tasks.
Every arm solved 4 of 10. The agent phase took \TimingAgentMin{} to \TimingAgentMax{} seconds on average, and grading
about ten seconds.

\begin{table}[t]
\centering
\small
\caption{Timing block: four 31B arms on ten Berth tasks, jobs run one after another on 4$\times$L4 with the software stack of the scorer. Times are mean seconds per job.}
\label{tab:timing}
\begin{tabular}{lrrrrrr}
\toprule
Arm & Solved & Agent & Setup & Verify & Tokens out & Tool calls \\
\midrule
\texttt{BF5} & 4/10 & 191 & 2.9 & 10.5 & 6{,}454 & 18.5 \\
\texttt{BF5-s} & 4/10 & 233 & 4.8 & 10.1 & 7{,}710 & 19.8 \\
\texttt{ID24} & 4/10 & 193 & 3.8 & 10.7 & 6{,}551 & 16.7 \\
\texttt{ID24-r} & 4/10 & 179 & 4.0 & 7.8 & 5{,}792 & 17.1 \\
\bottomrule
\end{tabular}
\end{table}

The second block compared \arm{BF5} with \arm{BF5-c}, which adds a checking step before the patch is submitted, on 16
tasks with two repeats. \arm{BF5-c} solved \CheckCand{} of \CheckRuns{} runs and \arm{BF5} \CheckBase; \CheckUp{}
tasks gained a solved run and \CheckDown{} lost one, and the agent phase was \CheckTime\% longer. No difference was
detected.

\subsection{Hosted submissions}\label{sec:hosted}

\begin{table}[t]
\centering
\small
\caption{Our five hosted submissions (one per day) and the local result of the same configuration. The public score is the solved share on the public part of the hidden test set, as displayed.}
\label{tab:hosted}
\begin{tabular}{llllp{0.42\linewidth}}
\toprule
Date (UTC) & Arm & Model & Public score & Local result of the same arm \\
\midrule
2026-11-05 & \texttt{W3} & 31B & 0.12 & 38 of 70 development tasks (emulated latency) \\
2026-11-08 & \texttt{ID24} & 31B & 0.17 & 4 of 10 Berth tasks (4$\times$L4) \\
2026-11-09 & \texttt{base} & E4B & 0.06 & 20.2\% of 1{,}616 Berth runs (workstation) \\
2026-11-10 & \texttt{BF5} & 31B & 0.18 & 37 of 70 development tasks (emulated latency) \\
2026-11-11 & \texttt{v-a3} & E4B & 0.05 & 118 of 512 Berth runs (panel H) \\
\bottomrule
\end{tabular}
\end{table}

\begin{figure}[t]
\centering
\includegraphics[width=0.9\linewidth]{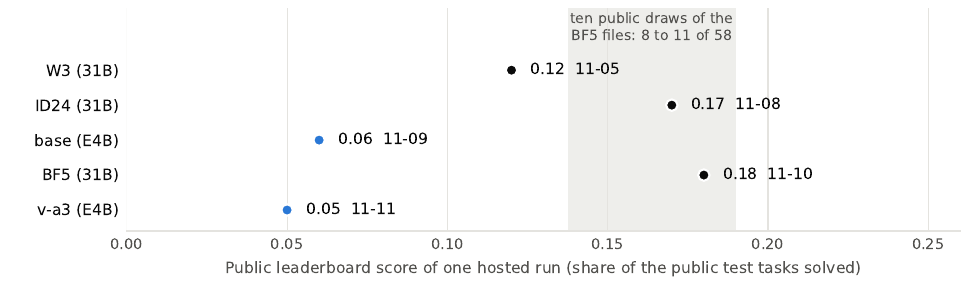}
\caption{Our hosted scores, one run each, against the range of ten public draws of the unchanged \arm{BF5} files
\citep{afshar2026census}.}
\label{fig:hosted}
\end{figure}

\Cref{tab:hosted} lists our five hosted submissions with the local result of the same arm. The scores are given as the
leaderboard displays them. Two of them, 0.06 and 0.18, are not a count of 58 tasks rounded to two digits, so we do
not convert scores into task counts.

Each score is one draw. The unchanged \arm{BF5} files scored between 8 and 11 of 58 in ten public draws
\citep{afshar2026census}, and our draw of these files, 0.18, lies in that range (\cref{fig:hosted}). \arm{W3} scored
0.12 and \arm{BF5} 0.18, while locally they solved 38 and 37 of the same 70 tasks. With a standard deviation near 1.5
tasks for one draw, the difference of two draws of the same files has a standard deviation near 0.04. The hosted
difference of 0.06 is about one and a half times that: it does not contradict the local tie, and it does not confirm
it. This is the situation that the
introduction describes: a difference of the size that development produces is not visible in a hosted score.

The two E4B files scored 0.06 (\arm{base}) and 0.05 (\arm{v-a3}, the third candidate of \cref{sec:programmes}). No
repeated draws of an E4B file exist, so the noise of an E4B score has not been measured. We read no direction from a
difference of 0.01 between two single draws. The local confirmation run of the same candidate (\cref{sec:confirm}) is
the evidence that we have, and its verdict is ``not confirmed''.

The local solve rates are far above the hosted scores for both models, on the development tasks and on Berth. Local
counts compare arms on the same tasks. They do not predict a hosted score.

\section{Experiments with E4B}\label{sec:small}

E4B generates \ServeRatioSmallBig{} times as many tokens per second as the 31B model on the scorer's hardware
(\cref{sec:serving}), and a separate prize of the competition is reserved for the small models. Most runs of this
paper, \RunsEFour{} of \RunsTotal{}, are E4B runs. This section reports what we measured before any candidate change was tested: the first
runs on hosted notebooks, the controls of the grading path, the agreement between the workstation and the hosted
notebook, how far the unchanged baseline repeats, where it succeeds and fails, and what a run costs. The three
programmes that tried to improve the baseline follow in \cref{sec:programmes}.

\subsection{First experiments on hosted notebooks}\label{sec:small-first}

\begin{table}[t]
\centering
\footnotesize
\caption{First E4B experiments on hosted 4$\times$L4 notebooks. Time: mean seconds per job (whole job in the first block, agent phase in the others).}
\label{tab:smallfirst}
\begin{tabular}{llrrrrrrr}
\toprule
Block & Arm & Tasks & Runs & Solved & \% & 95\% interval & Time & Tool calls \\
\midrule
\texttt{pubE4-1} & \texttt{e-single} & 24 & 24 & 5 & 21 & 9--40 & 124 & 8.7 \\
\texttt{pubE4-1} & \texttt{e-wrap} & 24 & 24 & 5 & 21 & 9--40 & 108 & 10.3 \\
\texttt{pubE4-1} & \texttt{e-wrap-ad} & 24 & 24 & 5 & 21 & 9--40 & 118 & 10.5 \\
\addlinespace
\texttt{berthE4-k1} & \texttt{e-BF} & 24 & 48 & 4 & 8 & 3--20 & 70 & 9.3 \\
\texttt{berthE4-k1} & \texttt{base} & 24 & 48 & 8 & 17 & 9--30 & 79 & 11.9 \\
\addlinespace
\texttt{berthE4-k2} & \texttt{base} & 8 & 16 & 7 & 44 & 23--67 & 82 & 12.4 \\
\texttt{berthE4-k2} & \texttt{base-p} & 8 & 16 & 6 & 38 & 18--61 & 80 & 12.1 \\
\texttt{berthE4-k2} & \texttt{e-BF} & 8 & 16 & 5 & 31 & 14--56 & 58 & 8.9 \\
\texttt{berthE4-k2} & \texttt{e-BF-s} & 8 & 16 & 2 & 12 & 3--36 & 60 & 9.3 \\
\bottomrule
\end{tabular}
\end{table}

\paragraph{Development tasks.}
The first block (\texttt{pubE4-1}, \cref{tab:smallfirst}) ran three E4B arms on 24 development tasks with small
reference fixes, from three of the four repositories: a single agent (\arm{e-single}), the wrapped design of
\cref{sec:big-arms} (\arm{e-wrap}), and the wrapped design with a fine-tuned adapter on the coder (\arm{e-wrap-ad}).
Each arm solved 5 of the 24 tasks. The arms largely solved the same tasks: \SmallPubAllThree{} tasks were solved by all
three and \SmallPubUnion{} by at least one. On these 24 tasks the 31B arms \arm{BF5} and \arm{SH6} had solved
\SmallPubBigBF{} and \SmallPubBigSH{}, each in one run per task on the emulated-latency machine. The block therefore shows the size of the gap
between the two models on tasks chosen to be easy, and no difference between a single agent, a wrapped agent and an
adapter. With one run per task and five solved tasks per arm, it could only have shown a very large difference. A job
took \SmallPubWall{}~s on average and used about \SmallPubTools{} tool calls: E4B stops by itself, far inside every
limit of the scorer.

\paragraph{Berth tasks.}
The second block (\texttt{berthE4-k1}) is the first on Berth tasks: 24 tasks, two repeats, and two arms in one job
queue. \arm{e-BF} is the public single-agent prompt adapted to E4B; \arm{base} is the withheld E4B agent bundle that
we later submitted. \arm{base} solved 8 of 48 runs and \arm{e-BF} 4 of 48. By task, \arm{base} was ahead on
\KOneBaseBetter{} tasks and behind on \KOneOtherBetter{} (sign test, $p = \KOneP$); no difference was detected. We
kept \arm{base} as the baseline of every later comparison. This block is not evidence that it is the better of the
two; it is the bundle we meant to submit, and it was not behind.

The third block (\texttt{berthE4-k2}) compared four arms on eight tasks, one from each of \KTwoRepos{} repositories,
with two repeats: \arm{base}, \arm{base} with one prompt change (\arm{base-p}), \arm{e-BF}, and \arm{e-BF} with
changed sampling settings (\arm{e-BF-s}). They solved 7, 6, 5 and 2 of 16 runs. The intervals of
\cref{tab:smallfirst} overlap widely, and we drew one conclusion only: 16 runs per arm on a hosted notebook cannot
rank prompt variants of this model. The eight tasks share no task with the 24 of the second block, and the record
does not say how they were picked. \arm{base} solved 44\% of its runs on them, twice its share elsewhere, so the block
says nothing about the level of any arm. Neither \arm{base-p} nor \arm{e-BF-s} was run again.

These three blocks hold 232 runs. All later E4B experiments ran on the workstation, where the same number of runs
takes about an hour.

\subsection{Task pool and grading controls}

\begin{table}[t]
\centering
\small
\caption{Control runs of the grading path: every task graded once with its reference patch (must pass) and once with no patch (must fail).}
\label{tab:controls}
\begin{tabular}{lrr}
\toprule
Tasks and machine & Reference patch passes & Empty patch passes \\
\midrule
whole pool, workstation, no container & 580/580 & 1/580 \\
panels C and H, final runner & 200/200 & 0/200 \\
screen set S, final runner & 48/48 & 0/48 \\
case-study panel and two tasks of S, final runner & 233/233 & 0/233 \\
hosted notebook, first batch & 38/38 & 0/38 \\
hosted notebook, second batch & 87/87 & 0/209 \\
\bottomrule
\end{tabular}
\end{table}

A solved share means something only if the grading path separates a correct fix from no fix. Before any agent ran on
a task, the task was graded twice: once with its reference patch, which must pass, and once with no patch, which must
fail (\cref{tab:controls}). On the workstation, without a container, the reference patch passed on all
\PoolTasks{} tasks that loaded, from \PoolRepos{} repositories, and the empty patch passed on \PoolEmptyPass{}; that
task was removed, which leaves the \PoolValid{} tasks of \cref{sec:setting}. The controls were then repeated under the
final container runner for every panel used in this paper, and on CPU notebooks of the hosting platform for
\HostedCtrlRef{} tasks with the reference patch and \HostedCtrlEmpty{} with the empty patch. Every one of these
controls gave the expected result. \CtrlPairsMulti{} pairs of task and control were graded on more than one machine or
runner, and \CtrlPairsDisagree{} of them changed its result. The file \texttt{controls.csv} holds all \ControlRows{}
rows.

The controls test the grader, not the agent's environment. They would not show a difference in what the agent sees
while it works, for example a different Python version in its shell; \cref{sec:parity} addresses that side.

\subsection{Workstation against the hosted notebook}\label{sec:parity}

\begin{table}[t]
\centering
\small
\caption{The baseline on the same 24 Berth tasks, one run per task, on the hosted notebook and on the workstation under successive runners. Agent: mean seconds of the agent phase.}
\label{tab:parity}
\begin{tabular}{lrrr}
\toprule
Run & Solved & 95\% interval, \% & Agent \\
\midrule
hosted 4$\times$L4, repeat 1 & 4/24 & 7--36 & 79 \\
hosted 4$\times$L4, repeat 2 & 4/24 & 7--36 & 80 \\
\addlinespace
workstation, no sandbox, 8 workers & 3/24 & 4--31 & 65 \\
workstation, no sandbox, 1 worker & 5/24 & 9--40 & 63 \\
workstation, container runner, version 1 & 3/24 & 4--31 & 69 \\
workstation, container runner, version 3 & 2/24 & 2--26 & 58 \\
workstation, container runner, version 4, run a & 4/24 & 7--36 & 83 \\
workstation, container runner, version 4, run b & 3/24 & 4--31 & 70 \\
workstation, container runner, version 7 (final) & 5/24 & 9--40 & 71 \\
\bottomrule
\end{tabular}
\par\vspace{3pt}{\footnotesize\raggedright A further workstation run under runner version 2 solved 4 of 24; its result rows were not kept and it is not in the released file.\par}
\end{table}

The workstation serves the model from one GPU instead of four, runs jobs in parallel, and puts each job in a
container. Any of these could change what the agent does. \Cref{tab:parity} lists every run of the baseline on the
same 24 Berth tasks, one run per task: two repeats on the hosted notebook with four L4 GPUs, and seven runs on the
workstation as the runner changed. The first two workstation runs used no sandbox and still had the role divergence
described in \cref{sec:intro}; the others used the container runner in four of its versions.

The hosted repeats solved 4 and 4 of 24. The workstation runs solved between \ParityMin{} and \ParityMax{}, in total
\ParityWsTotal{} of \ParityWsRuns{} runs (\ParityWsPct\%) against \ParityHostedTotal{} of 48 (\ParityHostedPct\%) on
the hosted notebook. Over all 216 runs only six distinct tasks were ever solved, so the totals move because the same
few tasks come out or not, not because a runner opens or closes other tasks. The mean agent phase took 79 and 80~s on
the hosted notebook and 58 to 83~s on the workstation. Figure~\ref{fig:parity}A shows the difference of each
workstation run from the hosted share; it includes an eighth run, under runner version 2, whose rows were not kept.

What this supports is limited. With 24 tasks and one run each, the 95\% interval of 4 of 24 runs from 7\% to 36\%.
The table rules out a gross difference between the machines, such as the workstation solving half or double the
hosted share. It cannot show that a specific difference in the prompts has no effect. That question needs an
experiment with power, and \cref{sec:case} reports one.

\subsection{How far the baseline repeats}

\begin{table}[t]
\centering
\footnotesize
\caption{The unchanged baseline run again on the same tasks in separate sessions on the workstation. Range: largest minus smallest session total in percentage points. Same: tasks with the same solved count in every session. Never, always: tasks solved in no run, in every run. $r$: correlation of the per-task counts of the first and last session.}
\label{tab:repeat}
\begin{tabular}{lrllrrrrr}
\toprule
Task set & Sessions & Solved runs & \% & Range & Same & Never & Always & $r$ \\
\midrule
S (48 tasks, 2 repeats) & 3 & 19, 21, 20 & 19.8, 21.9, 20.8 & 2.1 & 39 & 34 & 5 & 0.91 \\
H (128 tasks, 4 repeats) & 2 & 103, 97 & 20.1, 18.9 & 1.2 & 99 & 83 & 10 & 0.92 \\
\bottomrule
\end{tabular}
\end{table}

\begin{figure}[t]
\centering
\includegraphics[width=\linewidth]{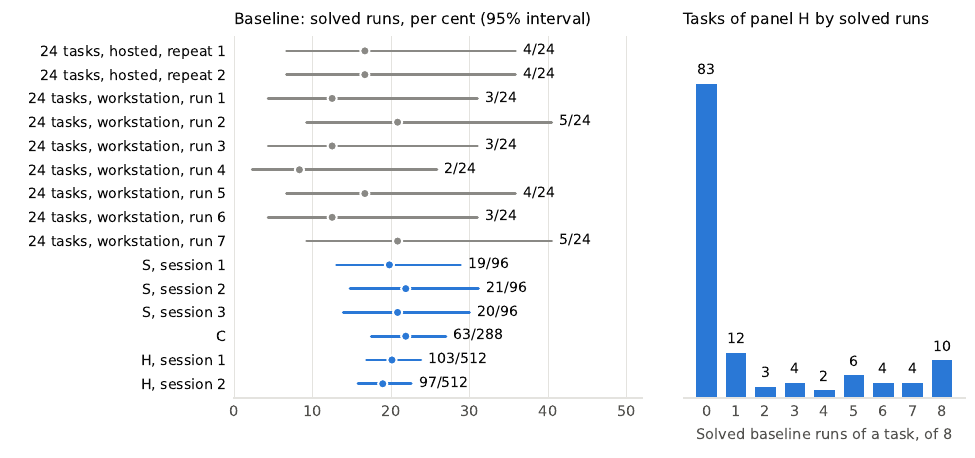}
\caption{The unchanged E4B baseline on the workstation. Left: solved runs of each session on the same tasks, with
95\% intervals. Right: the 128 tasks of panel H by the number of baseline runs that solved them, of eight runs in two
sessions.}
\label{fig:repeat}
\end{figure}

Every screen and every confirmation block ran the baseline in the same job queue as the candidates. The baseline
therefore ran three times on the 48 tasks of S and twice on the 128 tasks of H, in separate sessions
(\cref{tab:repeat,fig:repeat}). On S it solved 19, 21 and 20 of 96 runs; on H,
\RepeatHFirst{} and \RepeatHSecond{} of 512. The session totals differ by at most \RepeatSRange{} points on S and by
\RepeatHDiff{} points on H.

Repeats of one task can differ. They rarely do. On H, \HNever{} of the
128 tasks (\HNeverPct\%) were solved in none of the eight baseline runs and \HAlways{} in all eight; only \HBetween{}
tasks were solved in some runs and not in others. The per-task solved counts of the two sessions correlate at 0.92.
Because so few tasks are uncertain, the run-to-run standard deviation of the total is small. Estimated from the
variation within tasks, it is \NoiseSdRuns{} runs of 512, or \NoiseSdPts{} points, for one arm and \NoiseSdDiffPts{}
points for the difference of two arms on the same tasks. The two baseline sessions differ by
\BaseSessionsDiffRuns{} runs, which is \BaseSessionsNoiseRatio{} of that standard deviation; the one pair of repeats
that we have does not contradict the estimate. The binomial formula, which treats the 512 runs as independent with one
common probability of \HBasePct\%, gives \NoiseSdBinomPts{} points; the variation within tasks accounts for
\NoiseShare\% of that variance. \Cref{sec:analysis} uses these numbers.

\subsection{Repositories}

\begin{figure}[t]
\centering
\includegraphics[width=\linewidth]{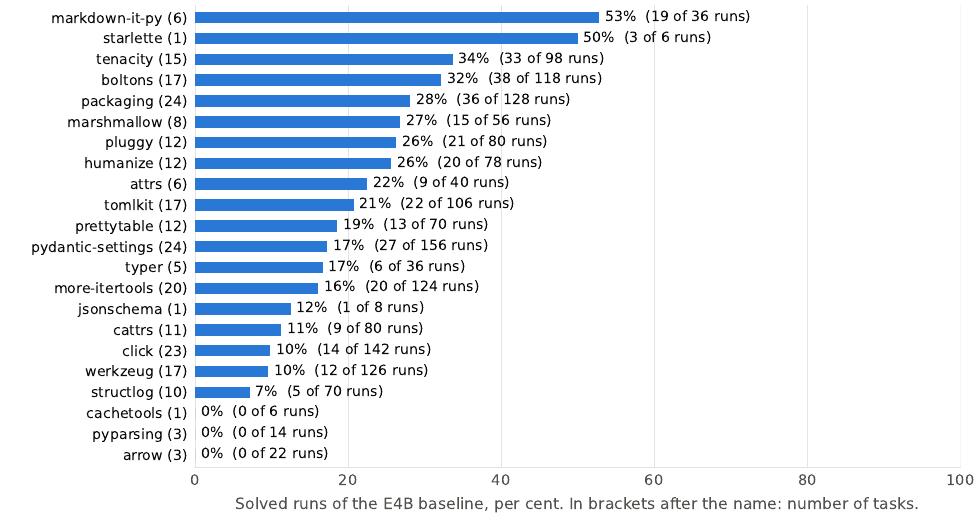}
\caption{Solved share of the E4B baseline by repository, over all its runs in the screen and confirmation blocks, with
95\% intervals that treat runs as independent. Numbers of tasks and runs are in \cref{tab:repos}.}
\label{fig:repos}
\end{figure}

\begin{table}[p]
\centering
\footnotesize
\caption{Solved share of the E4B baseline by repository, over every baseline run of the screen and confirmation blocks on the workstation. The interval treats runs as independent and is too narrow where few tasks carry many runs.}
\label{tab:repos}
\begin{tabular}{lrrrrr}
\toprule
Repository & Tasks & Runs & Solved & \% & 95\% interval \\
\midrule
\texttt{executablebooks/markdown-it-py} & 6 & 36 & 19 & 52.8 & 37--68 \\
\texttt{Kludex/starlette} & 1 & 6 & 3 & 50.0 & 19--81 \\
\texttt{jd/tenacity} & 15 & 98 & 33 & 33.7 & 25--43 \\
\texttt{mahmoud/boltons} & 17 & 118 & 38 & 32.2 & 24--41 \\
\texttt{pypa/packaging} & 24 & 128 & 36 & 28.1 & 21--36 \\
\texttt{marshmallow-code/marshmallow} & 8 & 56 & 15 & 26.8 & 17--40 \\
\texttt{pytest-dev/pluggy} & 12 & 80 & 21 & 26.2 & 18--37 \\
\texttt{python-humanize/humanize} & 12 & 78 & 20 & 25.6 & 17--36 \\
\texttt{python-attrs/attrs} & 6 & 40 & 9 & 22.5 & 12--38 \\
\texttt{python-poetry/tomlkit} & 17 & 106 & 22 & 20.8 & 14--29 \\
\texttt{jazzband/prettytable} & 12 & 70 & 13 & 18.6 & 11--29 \\
\texttt{pydantic/pydantic-settings} & 24 & 156 & 27 & 17.3 & 12--24 \\
\texttt{fastapi/typer} & 5 & 36 & 6 & 16.7 & 8--32 \\
\texttt{more-itertools/more-itertools} & 20 & 124 & 20 & 16.1 & 11--24 \\
\texttt{python-jsonschema/jsonschema} & 1 & 8 & 1 & 12.5 & 2--47 \\
\texttt{python-attrs/cattrs} & 11 & 80 & 9 & 11.2 & 6--20 \\
\texttt{pallets/click} & 23 & 142 & 14 & 9.9 & 6--16 \\
\texttt{pallets/werkzeug} & 17 & 126 & 12 & 9.5 & 6--16 \\
\texttt{hynek/structlog} & 10 & 70 & 5 & 7.1 & 3--16 \\
\texttt{arrow-py/arrow} & 3 & 22 & 0 & 0.0 & 0--15 \\
\texttt{pyparsing/pyparsing} & 3 & 14 & 0 & 0.0 & 0--22 \\
\texttt{tkem/cachetools} & 1 & 6 & 0 & 0.0 & 0--39 \\
\bottomrule
\end{tabular}
\end{table}

Over the screens and confirmations, the baseline ran \BaseWsRuns{} times on \BaseWsTasks{} tasks from \BaseWsRepos{}
repositories and solved \BaseWsSolved{} runs (\BaseWsPct\%). \Cref{tab:repos,fig:repos} split this by repository. The
share runs from \RepoTopPct\% in the best repository to zero in \RepoZero{} repositories; the three repositories with
the most solved runs hold \RepoTopThreeShare\% of all solved runs. The zero rows have one to three tasks each, so
they say little about those repositories. The wider pattern is firm, though: with 10 to 24 tasks each, the shares of
the larger repositories still range from 7\% to 34\%.

Two consequences follow. A change that helps in one repository can move the total visibly while doing nothing
elsewhere, which is why the pass rule of \cref{sec:compare} counts repositories and tests with repositories as
clusters. And a local total is a property of the repository mix as much as of the agent: the hidden tasks of the
scorer need not come from these repositories, and nothing here says how the baseline does on them.

\subsection{Time, tokens and failure kinds}\label{sec:cost}

\begin{table}[t]
\centering
\scriptsize\setlength{\tabcolsep}{3.1pt}
\caption{Time, tokens and endings of the agent runs by model and setting. Time: seconds per run (whole job for the development tasks, agent phase for Berth). In: prompt tokens summed over the requests of a run, in thousands. T/O, Budget, Empty: percentage of runs stopped by the time limit, stopped by the tool-call limit, and left without a patch.}
\label{tab:cost}
\begin{tabular}{p{0.2\linewidth}rrrrrrrrrrrr}
\toprule
Setting & Runs & Solved \% & Mean & Median & Solved & Unsolved & Out & In (k) & Tools & T/O & Budget & Empty \\
\midrule
31B, development tasks, emulated latency & 570 & 57.2 & 278 & 255 & 231 & 341 & 7{,}089 & 240 & 20.6 & 12.6 & 0.4 & 13.5 \\
31B, development tasks, 4$\times$L4 & 36 & 58.3 & 286 & 256 & 234 & 358 & 8{,}515 & 264 & 23.8 & 11.1 & 2.8 & 11.1 \\
31B, Berth, 4$\times$L4 & 104 & 38.5 & 226 & 212 & 165 & 265 & 7{,}495 & 196 & 19.2 & 8.7 & 0.0 & 4.8 \\
E4B, development tasks, 4$\times$L4 & 72 & 20.8 & 117 & 106 & 85 & 125 & 4{,}665 & 102 & 9.8 & 1.4 & 0.0 & 4.2 \\
E4B, Berth, 4$\times$L4 & 160 & 20.0 & 73 & 61 & 43 & 80 & 6{,}261 & 110 & 10.6 & 0.0 & 1.9 & 11.9 \\
E4B, Berth, workstation & 4{,}082 & 21.3 & 75 & 60 & 45 & 84 & 8{,}146 & 156 & 14.2 & 0.3 & 1.2 & 5.5 \\
\bottomrule
\end{tabular}
\end{table}

\begin{figure}[t]
\centering
\includegraphics[width=\linewidth]{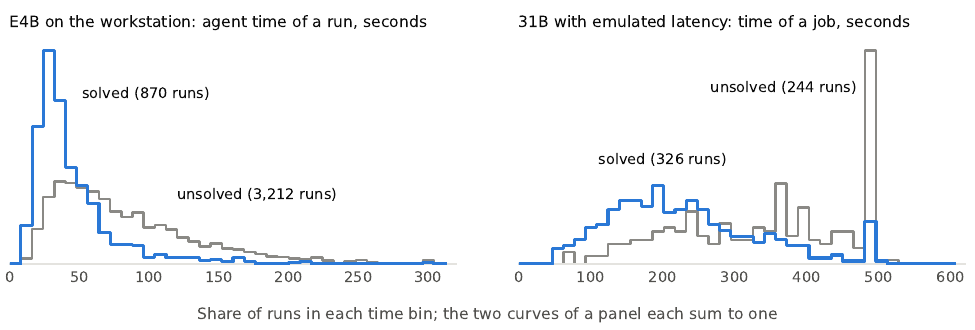}
\caption{Time of solved and unsolved runs for the two models. E4B: agent phase of the Berth runs on the workstation.
31B: whole job of the development-task runs under emulated latency.}
\label{fig:time}
\end{figure}

\Cref{tab:cost} gives time, tokens and endings for every setting in which agent runs were made, and \cref{fig:time}
the distribution of time for the two largest groups.

\paragraph{E4B is quick, and quicker when it succeeds.}
On the workstation the agent phase of an E4B run took \CostSmallMean{}~s on average: \CostSmallSolvedT{}~s when the
task was solved and \CostSmallUnsolvedT{}~s when it was not. The 31B runs on development tasks took
\CostBigMean{}~s, with \CostBigSolvedT{}~s and \CostBigUnsolvedT{}~s. The time limit ended 12.6\% of the 31B runs
and 0.3\% of the E4B runs. For the 31B model the limit is a working constraint, and the arms of \cref{sec:big} differ
mostly in how they spend it. For E4B it is not: more time would change almost nothing.

\paragraph{E4B fails by handing in a wrong fix.}
Of the \WsUnsolved{} unsolved E4B runs on the workstation, \WsUnsolvedNormalPct\% ended because the agent declared its
work finished. \WsUnsolvedWrongPct\% handed in a patch that the tests rejected, \WsUnsolvedEmptyPct\% left no patch,
and \WsUnsolvedTimeoutPct\% ran into the time limit. The typical failure is thus a wrong fix handed in after about a
minute and a half, not a run that was cut off. The three programmes of \cref{sec:programmes} accordingly changed what
the agent does within a run (its prompt, the way it edits files, an added agent step) and left the limits alone,
with one exception, an arm with a changed thinking-token limit.

\paragraph{Both models on the same tasks.}
\CommonTasks{} Berth tasks, all in S, were run with both models: with five 31B arms, the public
configurations and variants of them, on four L4 GPUs (\cref{sec:big}), and with the E4B baseline in the three screens on the workstation.
The 31B arms solved \CommonBigSolved{} of \CommonBigRuns{} runs (\CommonBigPct\%) and the E4B baseline
\CommonSmallSolved{} of \CommonSmallRuns{} (\CommonSmallPct\%). The 31B arms solved \CommonBigTasks{} of the tasks at
least once and E4B \CommonSmallTasks{}; no task was solved by E4B only. An average run took \CommonBigMean{}~s
against \CommonSmallMean{}~s, so the agent time spent per solved run is of the same order: \CommonBigPerSolved{}~s for
the 31B arms and \CommonSmallPerSolved{}~s for E4B. The comparison mixes machines and arms and covers 26 tasks; we
read it as a description, not as a test. It does say that the small model is not cheaper per solved task. It is
cheaper per attempt, and it reaches fewer tasks.

\paragraph{Totals.}
The \RunsWorkstation{} E4B runs on the workstation took \WsAgentHours{} hours of agent time in total, spread over
parallel workers, and used \WsTokensInM{} million prompt tokens and \WsTokensOutM{} million generated tokens.

\section{Case study: four prompt divergences on E4B}\label{sec:case}

\Cref{sec:conform} lists ways in which a local stack can send the model other prompts than the scorer does, and
\cref{sec:parity} shows that a comparison of 24 tasks cannot say whether such a divergence matters. This section
reports the experiment that can. It takes four divergences that we or public notebooks had actually run with, turns
each into one arm, and measures E4B's solved share against the scorer's path on 231 tasks with four repeats. The
design, the analysis and the power simulation were registered before the first job.

\Cref{sec:case-design} is the account of the short paper. The subsections after it give the full record: how the
block ran, every outcome count, the primary and secondary analyses with their sensitivity checks, how many requests
each divergence actually changed, and the exploratory measures of time and tokens. The rows are in the released
files \texttt{e8\_runs.csv}, \texttt{e8\_tasks.csv} and \texttt{e8\_repositories.csv}, and \cref{sec:app-recompute}
repeats the analysis with the released kit.

\subsection{Design and main result}\label{sec:case-design}

% written by scripts/reuse_short.py from the sources of the short paper; do not edit
Five arms share one model server, one withheld agent bundle, google-adk 1.36.1 and all sampling settings.

\begin{table}[!tb]
\centering
\caption{Arms; changes in percentage points.}\label{tab:casearms}
\footnotesize
\begin{tabular}{@{}
  >{\raggedright\arraybackslash}p{(\linewidth - 6\tabcolsep) * \real{0.0500}}
  >{\raggedright\arraybackslash}p{(\linewidth - 6\tabcolsep) * \real{0.4000}}
  >{\raggedright\arraybackslash}p{(\linewidth - 6\tabcolsep) * \real{0.1900}}
  >{\raggedright\arraybackslash}p{(\linewidth - 6\tabcolsep) * \real{0.3600}}@{}}
\toprule
\begin{minipage}[b]{\linewidth}\raggedright
Arm
\end{minipage} & \begin{minipage}[b]{\linewidth}\raggedright
Difference from the scorer path
\end{minipage} & \begin{minipage}[b]{\linewidth}\raggedright
Requests changed by the divergence
\end{minipage} & \begin{minipage}[b]{\linewidth}\raggedright
Change against F {[}95\% interval{]}; Holm p
\end{minipage} \\
\midrule
F & none: role \texttt{tool}, Kaggle template, compaction at 14,336 tokens, thoughts under both keys & --- & solve rate 23.8\% \\
R & model name contains ``gemma4'': tool results get role \texttt{tool\_responses} & 89\% & \ensuremath{-}0.3 {[}\ensuremath{-}2.8, +2.2{]}; p = 1 \\
C & compaction unset: prompts grow until rejected above 24,576 tokens (the run ends; edits are still graded) & 21\% & +0.2 {[}\ensuremath{-}2.0, +2.5{]}; p = 1 \\
H & history thoughts removed from every request, as when a client omits the key the server reads & 90\% & \ensuremath{-}2.1 {[}\ensuremath{-}4.4, +0.4{]}; p = 0.44 \\
T & the Hugging Face E4B template: a thought opener is forced after each tool result & 85\% & +0.8 {[}\ensuremath{-}1.4, +2.9{]}; p = 1 \\
\bottomrule
\end{tabular}
\end{table}

No newline arm was run: both templates render it identically on this server.

\paragraph{Panel.} Removing tasks used by other panels, reserved repositories, slow tasks and duplicates from the 579 leaves 231 tasks in 15 repositories. They run 4 repeats in 5 arms (4,620 jobs); for each task and repeat the arms run back to back in counterbalanced order.

\textbf{Analysis, frozen before any outcome} (protocol sha256 \texttt{\seqsplit{070c9655d96372d054f430727259de7b25af2ca93c3ec4c9295dfdc37aeb8cf4}}). The estimand $d_X$ is the mean over tasks of arm X's solve rate minus arm F's. The primary test is an exact two-sided task-level sign-flip test, Holm-corrected over the four arms at 0.05; only then do we say that X raises or lowers E4B's solve rate \emph{on these tasks}. Intervals (unadjusted) invert that test. A repository-cluster bootstrap interval is secondary (lower simulated power). A frozen checker re-renders every returned request from the job's dump, matches the server's token count and confirms the job's arm; a failing job is re-run. Outcomes stayed blind until a pre-registered cut-off (2026-11-11 06:00 UTC) and were unblinded once.

\paragraph{Power, simulated before data.} The family-wise false-claim rate is 1.2--4.2\%. At 231 tasks an even 5-point drop in one arm is detected with probability 82--100\% across the simulated settings; the drop or gain detected with 80\% power ranges from below 3 to about 5.2 points.

\paragraph{Deviations recorded blind.} Independent validation of the final runner finished only after launch. From about an hour in, a second workload held about 9.6 GiB of GPU memory, so timings are under mixed load. The block stopped three times: after 967 jobs at the runner's open-file limit (cause unknown), after 378 more by an external termination signal on the shared machine, and after 3,074 more when the kernel's out-of-memory killer ended the model server (cause undiagnosed). Each resume ran only never-started jobs; the 24 interrupted jobs were re-run once.

\begin{figure}[!tb]
\centering
\includegraphics[width=\linewidth]{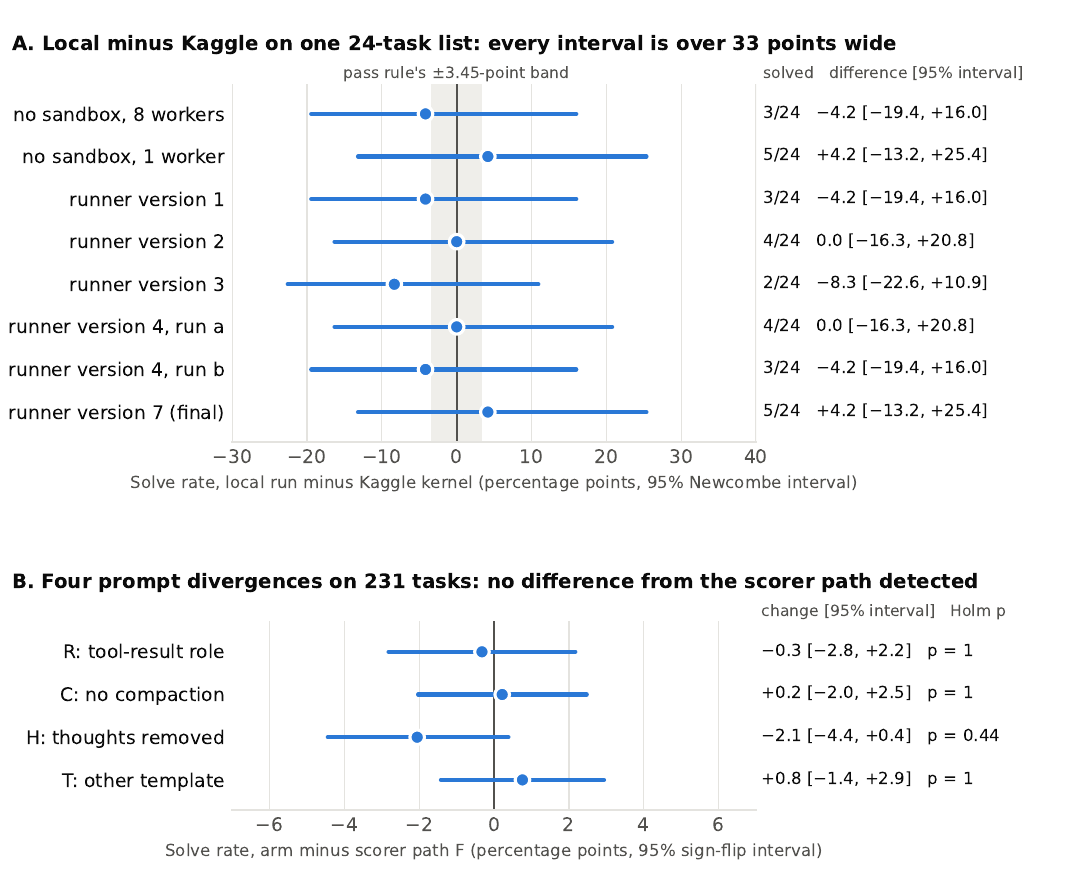}
\caption{(A) Solve-rate difference, local minus Kaggle (pooled 8/48), for eight local runs of one 24-task list (95\% Newcombe intervals, runs treated as independent); shaded: the pass rule's \ensuremath{\pm}3.45 points. (B) Change in solve rate against F on 231 tasks: estimates, 95\% intervals and Holm-adjusted p.}\label{fig:parity}
\end{figure}

\paragraph{Results.} All jobs finished before the cut-off. Every arm has a valid row in 924 of 924 slots, and the checker passed all 4,620 analysed jobs. No pre-emption was recorded; 6.0\% of runs took over 240 s, above the pre-registered 5\% threshold. No difference from F was detected for any arm (Table~\ref{tab:casearms}, Figure~\ref{fig:parity}B; smallest Holm-adjusted p 0.44); the secondary repository-cluster intervals all include zero. The largest estimate is H's \ensuremath{-}2.1 points (unadjusted p 0.11; \ensuremath{-}1.3 with time-limit rows as missing). In exploratory measures, H runs averaged 137 s against 78 s and ended unsolved at the bundle's five-minute limit in 11.5\% of runs against 1.8\%; in C, 12.3\% of runs ended unsolved on a rejected over-long prompt. These are null results, not evidence of no effect: the simultaneous 98.75\% intervals allow a drop of 5.1 points (H) and a gain of 3.5 (T).

% written by scripts/reuse_short.py from the sources of the short paper; do not edit
\subsection{Execution record}

The first job started on \SFirstStart{} UTC and the last one ended on \SLastEnd{} UTC, before the cut-off
(2026-11-11 06:00 UTC). The block ran as four batches and one batch of re-runs (Table~\ref{tab:s-batches}). Three
batches stopped early, for the reasons in the last column, and each following batch started only jobs that had never
started. In all, \SJobsStarted{} jobs were started: the \SAgentRows{} analysed runs and \SInfraRowsAll{} interrupted
jobs. Of the interrupted jobs, \SInfraRowsWritten{} left an infrastructure row; the other \SInfraNoRow{}, all in
batch~1, left none and are counted from the list of started jobs. Each interrupted slot was re-run once, so every slot
has exactly one analysed run. The released file holds every written row.

\begin{table}[!htb]
\centering
\caption{Execution record: the block ran as four batches and one batch of re-runs. An interrupted job is an infrastructure row, with or without a written result row; its slot was re-run once.}\label{tab:s-batches}
\footnotesize
\begin{tabular}{rrrrrl}
\toprule
Batch & \parbox[b]{1.2cm}{\raggedleft Jobs started} & \parbox[b]{1.4cm}{\raggedleft Analysed runs} & \parbox[b]{1.9cm}{\raggedleft Interrupted, row written} & \parbox[b]{1.9cm}{\raggedleft Interrupted, no row} & How the batch ended \\
\midrule
1 & 967 & 959 & 0 & 8 & runner reached its open-file limit \\
2 & 378 & 370 & 8 & 0 & external termination signal \\
3 & 3{,}074 & 3{,}066 & 8 & 0 & model server ended (out of memory) \\
4 & 201 & 201 & 0 & 0 & job list completed \\
5 & 24 & 24 & 0 & 0 & re-runs of the interrupted jobs completed \\
\bottomrule
\end{tabular}
\end{table}

\SLoadAbove{} of the \SLoadRows{} runs (\SLoadShare\%) took more than \SLoadThreshold{}~s of agent time. This is above
the \SLoadFlag\% share that the protocol set as the sign of a loaded machine: from about an hour into the block a
second workload held part of the memory of the GPU (\cref{sec:case-design}), so the times are under mixed load. No pre-emption
was recorded.

\subsection{Outcomes and solve rates}

Table~\ref{tab:s-outcomes} splits the runs of each arm by how they ended and Table~\ref{tab:s-rates} gives the solve
rates, which lie between \SMinRate\% and \SMaxRate\%. Two arms end differently from F. In C, \SCRejected{} runs
(\SCRejectedShare\%) ended unsolved because a prompt grew past the server's limit and was rejected; no F run ended
this way, and \SFCompactedShare\% of the F runs made at least one history summary. In H, \SHUnsolvedAtLimit{} runs
(\SHUnsolvedAtLimitShare\%) were unsolved at the bundle's five-minute limit, against \SFUnsolvedAtLimit{}
(\SFUnsolvedAtLimitShare\%) in F.

\begin{table}[!htb]
\centering
\caption{Outcome of every analysed run, by arm (one row per task, arm and repeat).}\label{tab:s-outcomes}
\footnotesize
\begin{tabular}{lrrrrrrr}
\toprule
Arm & \parbox[b]{1.75cm}{\raggedleft Solved} & \parbox[b]{1.75cm}{\raggedleft Unsolved, run completed} & \parbox[b]{1.75cm}{\raggedleft Five-minute limit} & \parbox[b]{1.75cm}{\raggedleft Over-long prompt rejected} & \parbox[b]{1.75cm}{\raggedleft Tool-call budget (28)} & \parbox[b]{1.75cm}{\raggedleft No submission} & Runs \\
\midrule
F & 220 & 676 & 17 & 0 & 6 & 5 & 924 \\
R & 217 & 677 & 20 & 0 & 9 & 1 & 924 \\
C & 222 & 568 & 17 & 114 & 1 & 2 & 924 \\
H & 201 & 597 & 106 & 1 & 18 & 1 & 924 \\
T & 227 & 668 & 19 & 0 & 4 & 6 & 924 \\
\bottomrule
\end{tabular}

\par\smallskip\parbox{0.96\linewidth}{\footnotesize Only the runs of the first column are solved; the other columns split the unsolved runs by how the run ended. A run that ended at a limit was still graded on the edits it had made: 5 solved runs lasted the full five minutes (all in arm H), and 86 of the 179 runs unsolved at that limit had left edits that were tested.}
\end{table}

\begin{table}[!htb]
\centering
\caption{Solve rate per arm: task-weighted mean with a 95\% interval from a repository-clustered (CR1) standard error; solved runs per repeat.}\label{tab:s-rates}
\footnotesize
\begin{tabular}{lrrrrrrrr}
\toprule
Arm & Solve rate (\%) & 95\% interval & Solved runs & Repeat 1 & 2 & 3 & 4 & Runs \\
\midrule
F & 23.81 & [16.7, 30.9] & 220 & 50 & 59 & 57 & 54 & 924 \\
R & 23.48 & [17.7, 29.3] & 217 & 53 & 53 & 54 & 57 & 924 \\
C & 24.03 & [16.6, 31.5] & 222 & 58 & 56 & 55 & 53 & 924 \\
H & 21.75 & [15.9, 27.6] & 201 & 50 & 49 & 54 & 48 & 924 \\
T & 24.57 & [18.6, 30.5] & 227 & 59 & 56 & 51 & 61 & 924 \\
\bottomrule
\end{tabular}
\end{table}

Table~\ref{tab:s-taskhist} shows why a difference between arms is hard to detect on this panel: \STasksNever{} of the
\STasks{} tasks were solved in no run of any arm and \STasksAlways{} in every run, so the other \STasksMixed{} tasks
carry every difference.

\begin{table}[!htb]
\centering
\caption{Tasks by the number of solved repeats (of four), per arm.}\label{tab:s-taskhist}
\footnotesize
\begin{tabular}{lrrrrrr}
\toprule
Arm & 0 solved & 1 solved & 2 solved & 3 solved & 4 solved & Tasks \\
\midrule
F & 149 & 16 & 21 & 18 & 27 & 231 \\
R & 152 & 14 & 21 & 15 & 29 & 231 \\
C & 151 & 15 & 18 & 17 & 30 & 231 \\
H & 155 & 21 & 13 & 14 & 28 & 231 \\
T & 151 & 14 & 13 & 25 & 28 & 231 \\
\bottomrule
\end{tabular}

\par\smallskip\parbox{0.96\linewidth}{\footnotesize 135 of the 231 tasks were solved in no run of any arm (0 of 20 runs) and 13 in every run; the other 83 tasks carry every difference between arms.}
\end{table}

\subsection{Primary analysis: tasks}

For a task, let $K$ be the number of solved repeats in the arm minus the number in F, from $-4$ to $+4$. The estimate
$d$ is the mean of $K/4$ over the \STasks{} tasks. Under the null hypothesis the arm and F are exchangeable within a
task, so each $K$ is as likely to carry one sign as the other; the test compares $|\sum K|$ with its distribution
over all sign assignments, computed exactly. Holm's method adjusts the four $p$-values. The intervals invert the same
test under a constant shift, with \SReproDraws{} frozen Monte Carlo sign draws.

\begin{figure}[!htb]
\centering
\includegraphics[width=\linewidth]{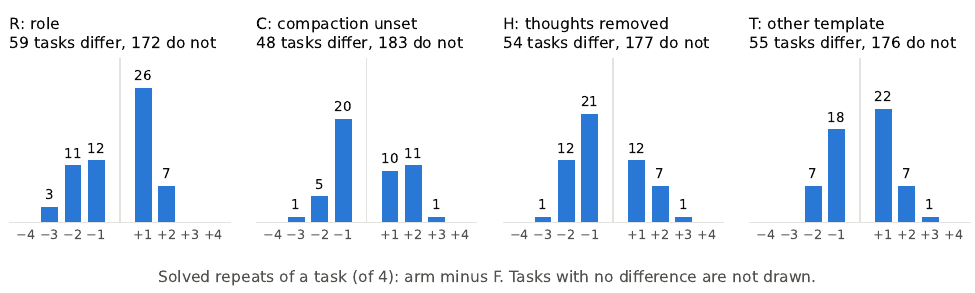}
\caption{Tasks whose number of solved repeats differs between an arm and F. Bars to the left of the centre line are
tasks with fewer solved repeats in the arm.}
\label{fig:s-taskdiff}
\end{figure}

No difference from F was detected for any arm (Table~\ref{tab:s-primary}, Figure~\ref{fig:s-taskdiff}). In every arm
between \SNoDifferMin{} and \SNoDifferMax{} of the \STasks{} tasks have the same number of solved repeats as in F. The
largest estimate is that of H: \SHTasksDown{} tasks have fewer solved repeats than in F and \SHTasksUp{} have more,
which gives $-\SHd$ points with exact $p = \SHp$ and Holm-adjusted $p = \SHHolm$. These are null results, not
evidence of no effect. The simultaneous 98.75\% intervals show what the experiment cannot exclude: a drop of up to
\SHSimLo{} points (H) and a gain of up to \STSimHi{} points (T).

\begin{table}[!htb]
\centering
\caption{Primary analysis: change in solve rate against F in percentage points, tasks with fewer and with more solved repeats than in F, exact two-sided task-level sign-flip test, Holm adjustment over the four arms, and intervals from inverting the test.}\label{tab:s-primary}
\footnotesize\setlength{\tabcolsep}{4pt}
\begin{tabular}{lrrrrrrrr}
\toprule
Arm & $d$ & Exact $d$ & \parbox[b]{1.5cm}{\raggedleft Tasks lower; higher} & Exact $p$ & Holm $p$ & \parbox[b]{1.7cm}{\raggedleft Monte Carlo $p$} & 95\% interval & 98.75\% simultaneous \\
\midrule
R & ${-}0.32$ & ${-}1/308$ & 26; 33 & 0.8652 & 1.000 & 0.8650 & [${-}2.82$, ${+}2.17$] & [${-}3.51$, ${+}2.85$] \\
C & ${+}0.22$ & $1/462$ & 26; 22 & 0.9253 & 1.000 & 0.9243 & [${-}2.03$, ${+}2.48$] & [${-}2.65$, ${+}3.09$] \\
H & ${-}2.06$ & ${-}19/924$ & 34; 20 & 0.1102 & 0.441 & 0.1106 & [${-}4.45$, ${+}0.38$] & [${-}5.09$, ${+}0.95$] \\
T & ${+}0.76$ & $1/132$ & 25; 30 & 0.5604 & 1.000 & 0.5599 & [${-}1.42$, ${+}2.94$] & [${-}2.02$, ${+}3.54$] \\
\bottomrule
\end{tabular}
\end{table}

\subsection{Secondary analysis: repositories}

\begin{figure}[!htb]
\centering
\includegraphics[width=\linewidth]{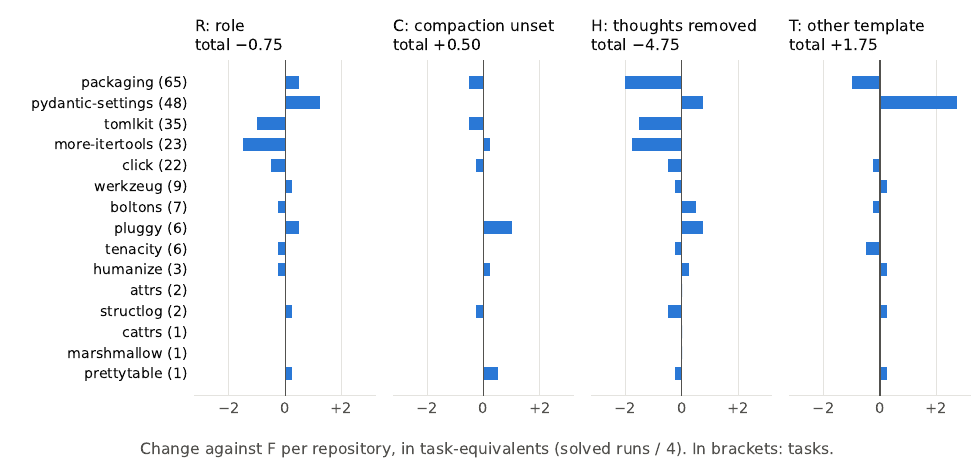}
\caption{Change against F in each repository, per arm. Repositories are ordered by their number of tasks.}
\label{fig:s-repos}
\end{figure}

Tasks of one repository tend to succeed and fail together, so the registered secondary analysis takes the \SRepos{}
repositories as its units (Table~\ref{tab:s-repolevel}). Table~\ref{tab:s-repos} and Figure~\ref{fig:s-repos} show
each repository. All four repository-level intervals include zero, and the smallest two-sided $p$ is \SMinRepoP{} (H).
The leave-one-repository-out estimates of H stay negative ($-\SHLooLo$ to $-\SHLooHi$ points), so no single
repository produces the estimate of H; those of R, C and T change sign.

\begin{table}[!htb]
\centering
\caption{Secondary analysis at the repository level (15 repositories): exact wild cluster bootstrap-$t$ with Rademacher weights, its inverted 95\% interval, the six-point (Webb) variant, a CR1 interval, the repository-equal estimate and the range of leave-one-repository-out estimates. Estimates in percentage points.}\label{tab:s-repolevel}
\footnotesize\setlength{\tabcolsep}{4pt}
\begin{tabular}{lrrrrrrr}
\toprule
Arm & \parbox[b]{1.9cm}{\raggedleft Repositories that differ} & Exact $p$ & Webb $p$ & 95\% interval & CR1 interval & \parbox[b]{1.9cm}{\raggedleft Repository-equal} & Leave-one-out range \\
\midrule
R & 12 & 0.8188 & 0.7780 & [${-}3.81$, ${+}2.38$] & [${-}2.70$, ${+}2.05$] & ${+}1.62$ & [${-}1.09$, ${+}0.36$] \\
C & 8 & 0.8438 & 0.7788 & [${-}0.88$, ${+}2.81$] & [${-}1.22$, ${+}1.65$] & ${+}4.02$ & [${-}0.22$, ${+}0.60$] \\
H & 12 & 0.1831 & 0.1736 & [${-}4.94$, ${+}1.56$] & [${-}4.65$, ${+}0.53$] & ${-}2.98$ & [${-}3.01$, ${-}1.44$] \\
T & 9 & 0.8047 & 0.7374 & [${-}1.44$, ${+}4.62$] & [${-}2.09$, ${+}3.61$] & ${+}2.65$ & [${-}0.55$, ${+}1.66$] \\
\bottomrule
\end{tabular}
\end{table}

\begin{table}[!htb]
\centering
\caption{The 15 repositories of the panel: tasks, runs solved in arm F (of four per task) and, per arm, the summed change against F in task-equivalents (solved runs divided by four).}\label{tab:s-repos}
\footnotesize
\begin{tabular}{lrrrrrr}
\toprule
Repository & Tasks & F solved runs & R & C & H & T \\
\midrule
\texttt{pypa/packaging} & 65 & 65 of 260 & ${+}0.50$ & ${-}0.50$ & ${-}2.00$ & ${-}1.00$ \\
\texttt{pydantic/pydantic-settings} & 48 & 27 of 192 & ${+}1.25$ & 0 & ${+}0.75$ & ${+}2.75$ \\
\texttt{python-poetry/tomlkit} & 35 & 31 of 140 & ${-}1.00$ & ${-}0.50$ & ${-}1.50$ & 0 \\
\texttt{more-itertools/more-itertools} & 23 & 39 of 92 & ${-}1.50$ & ${+}0.25$ & ${-}1.75$ & 0 \\
\texttt{pallets/click} & 22 & 13 of 88 & ${-}0.50$ & ${-}0.25$ & ${-}0.50$ & ${-}0.25$ \\
\texttt{pallets/werkzeug} & 9 & 3 of 36 & ${+}0.25$ & 0 & ${-}0.25$ & ${+}0.25$ \\
\texttt{mahmoud/boltons} & 7 & 12 of 28 & ${-}0.25$ & 0 & ${+}0.50$ & ${-}0.25$ \\
\texttt{jd/tenacity} & 6 & 9 of 24 & ${-}0.25$ & 0 & ${-}0.25$ & ${-}0.50$ \\
\texttt{pytest-dev/pluggy} & 6 & 3 of 24 & ${+}0.50$ & ${+}1.00$ & ${+}0.75$ & 0 \\
\texttt{python-humanize/humanize} & 3 & 11 of 12 & ${-}0.25$ & ${+}0.25$ & ${+}0.25$ & ${+}0.25$ \\
\texttt{hynek/structlog} & 2 & 2 of 8 & ${+}0.25$ & ${-}0.25$ & ${-}0.50$ & ${+}0.25$ \\
\texttt{python-attrs/attrs} & 2 & 0 of 8 & 0 & 0 & 0 & 0 \\
\texttt{jazzband/prettytable} & 1 & 1 of 4 & ${+}0.25$ & ${+}0.50$ & ${-}0.25$ & ${+}0.25$ \\
\texttt{marshmallow-code/marshmallow} & 1 & 0 of 4 & 0 & 0 & 0 & 0 \\
\texttt{python-attrs/cattrs} & 1 & 4 of 4 & 0 & 0 & 0 & 0 \\
\midrule
All & 231 & 220 of 924 & ${-}0.75$ & ${+}0.50$ & ${-}4.75$ & ${+}1.75$ \\
\bottomrule
\end{tabular}
\end{table}

\subsection{Sensitivity analyses}

Three sensitivity analyses were registered with the primary one (Table~\ref{tab:s-sens}). The first treats the runs
that were unsolved at the five-minute limit as missing, in both arms; the estimate of H is then $\SHTimeCapMissing$
points ($p = \SHTimeCapMissingP$) instead of $-\SHd$. The other two score as unsolved every slot whose first job was
interrupted, in the arm or in F. No sensitivity analysis gives a $p$-value below 0.05.

\begin{table}[!htb]
\centering
\caption{Pre-registered sensitivity analyses of the primary estimate (percentage points; exact two-sided sign-flip $p$, unadjusted).}\label{tab:s-sens}
\footnotesize
\begin{tabular}{llrrrr}
\toprule
Arm & Analysis & Tasks & Runs affected & $d$ & Exact $p$ \\
\midrule
R & five-minute-limit runs of both arms treated as missing & 229 & 37 & ${-}0.29$ & 0.8652 \\
 & slots of the arm with an infrastructure row scored unsolved & 231 & 5 & ${-}0.43$ & 0.8011 \\
 & slots of F with an infrastructure row scored unsolved & 231 & 3 & ${-}0.22$ & 0.9321 \\
C & five-minute-limit runs of both arms treated as missing & 229 & 34 & ${+}0.22$ & 0.9253 \\
 & slots of the arm with an infrastructure row scored unsolved & 231 & 5 & ${+}0.11$ & 1.0000 \\
 & slots of F with an infrastructure row scored unsolved & 231 & 3 & ${+}0.32$ & 0.8506 \\
H & five-minute-limit runs of both arms treated as missing & 226 & 123 & ${-}1.25$ & 0.3414 \\
 & slots of the arm with an infrastructure row scored unsolved & 231 & 8 & ${-}2.16$ & 0.0980 \\
 & slots of F with an infrastructure row scored unsolved & 231 & 3 & ${-}1.95$ & 0.1270 \\
T & five-minute-limit runs of both arms treated as missing & 229 & 36 & ${+}0.76$ & 0.5604 \\
 & slots of the arm with an infrastructure row scored unsolved & 231 & 3 & ${+}0.76$ & 0.5604 \\
 & slots of F with an infrastructure row scored unsolved & 231 & 3 & ${+}0.87$ & 0.4947 \\
\bottomrule
\end{tabular}
\end{table}

\subsection{Exposure and the prompt check}

A divergence can only matter in the requests it changes. The frozen checker re-rendered all \SRequests{} returned
requests from the request dumps of the jobs. For each of the \SChecksPassed{} analysed runs it confirmed that the
prompts carry the divergence of the run's arm and no other difference; \SChecksFailed{} runs failed the check. The
divergence changed \SExposureMin{} to \SExposureMax\% of the requests in R, H and T, and \SCChanged\% in C
(Table~\ref{tab:s-exposure}).

\begin{table}[!htb]
\centering
\caption{Exposure: requests re-rendered by the frozen checker, and those whose rendered prompt the arm's divergence changed.}\label{tab:s-exposure}
\footnotesize
\begin{tabular}{lrrrrr}
\toprule
Arm & Runs & Requests checked & Changed by the divergence & Share (\%) & Failed calls recorded \\
\midrule
F & 924 & 15{,}043 & 0 & 0.0 & 17 \\
R & 924 & 15{,}290 & 13{,}616 & 89.1 & 20 \\
C & 924 & 13{,}036 & 2{,}709 & 20.8 & 131 \\
H & 924 & 17{,}450 & 15{,}684 & 89.9 & 112 \\
T & 924 & 14{,}787 & 12{,}629 & 85.4 & 19 \\
\bottomrule
\end{tabular}

\par\smallskip\parbox{0.96\linewidth}{\footnotesize No run has more than one failed call. Of the 299 failed calls, 178 are in runs unsolved at the five-minute limit, 115 in runs that ended on a rejected over-long prompt and 6 in solved runs.}
\end{table}

\subsection{Exploratory run measures}

\begin{figure}[!htb]
\centering
\includegraphics[width=\linewidth]{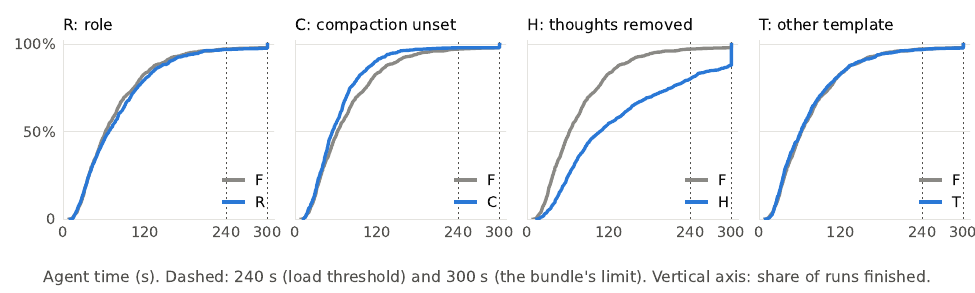}
\caption{Agent time per run: each arm (blue) against F (grey).}
\label{fig:s-time}
\end{figure}

Table~\ref{tab:s-measures} compares thirteen measures of a run between each arm and F. These comparisons were not
part of the registered tests and are not adjusted for multiplicity. H runs are longer: \SHTimeMean{}~s of agent time
on average against \SFTimeMean{}~s in F (medians \SHTimeMedian{} and \SFTimeMedian{}~s; Figure~\ref{fig:s-time}), with
\SHTokOut{} completion tokens per run against \SFTokOut.

\begin{table}[!htb]
\centering
\caption{Exploratory run measures: mean in arm F, and the paired difference of each arm against F; below it, its 95\% interval (CR1, repository clusters). No multiplicity adjustment.}\label{tab:s-measures}
\footnotesize\setlength{\tabcolsep}{3.5pt}
\begin{tabular}{lrrrrr}
\toprule
Measure & F mean & R $-$ F & C $-$ F & H $-$ F & T $-$ F \\
\midrule
history summaries per run & 0.69 & ${+}0.13$ & ${-}0.69$ & ${-}0.30$ & ${-}0.02$ \\
 &  & {\scriptsize [${+}0.04$, ${+}0.22$]} & {\scriptsize [${-}0.86$, ${-}0.53$]} & {\scriptsize [${-}0.47$, ${-}0.12$]} & {\scriptsize [${-}0.10$, ${+}0.06$]} \\[-2pt]
\addlinespace[2pt]% \\
runs with a summary (share) & 0.482 & ${+}0.038$ & ${-}0.482$ & ${-}0.161$ & ${-}0.014$ \\
 &  & {\scriptsize [${-}0.024$, ${+}0.099$]} & {\scriptsize [${-}0.572$, ${-}0.391$]} & {\scriptsize [${-}0.249$, ${-}0.073$]} & {\scriptsize [${-}0.059$, ${+}0.030$]} \\[-2pt]
\addlinespace[2pt]% \\
largest prompt (tokens) & 12{,}721 & ${+}220$ & ${+}2{,}160$ & ${-}1{,}499$ & ${-}95$ \\
 &  & {\scriptsize [${+}111$, ${+}329$]} & {\scriptsize [${+}1{,}666$, ${+}2{,}655$]} & {\scriptsize [${-}1{,}903$, ${-}1{,}095$]} & {\scriptsize [${-}284$, ${+}95$]} \\[-2pt]
\addlinespace[2pt]% \\
over-long prompt rejected (share) & 0.000 & ${+}0.000$ & ${+}0.123$ & ${+}0.001$ & ${+}0.000$ \\
 &  & {\scriptsize [${+}0.000$, ${+}0.000$]} & {\scriptsize [${+}0.075$, ${+}0.172$]} & {\scriptsize [${-}0.001$, ${+}0.003$]} & {\scriptsize [${+}0.000$, ${+}0.000$]} \\[-2pt]
\addlinespace[2pt]% \\
five-minute limit (share) & 0.018 & ${+}0.003$ & ${+}0.000$ & ${+}0.096$ & ${+}0.002$ \\
 &  & {\scriptsize [${-}0.005$, ${+}0.012$]} & {\scriptsize [${-}0.005$, ${+}0.005$]} & {\scriptsize [${+}0.073$, ${+}0.119$]} & {\scriptsize [${-}0.002$, ${+}0.006$]} \\[-2pt]
\addlinespace[2pt]% \\
tool-call budget reached (share) & 0.006 & ${+}0.003$ & ${-}0.005$ & ${+}0.013$ & ${-}0.002$ \\
 &  & {\scriptsize [${-}0.004$, ${+}0.011$]} & {\scriptsize [${-}0.012$, ${+}0.001$]} & {\scriptsize [${-}0.002$, ${+}0.028$]} & {\scriptsize [${-}0.006$, ${+}0.002$]} \\[-2pt]
\addlinespace[2pt]% \\
no submission (share) & 0.005 & ${-}0.004$ & ${-}0.003$ & ${-}0.004$ & ${+}0.001$ \\
 &  & {\scriptsize [${-}0.015$, ${+}0.006$]} & {\scriptsize [${-}0.013$, ${+}0.007$]} & {\scriptsize [${-}0.015$, ${+}0.006$]} & {\scriptsize [${-}0.010$, ${+}0.012$]} \\[-2pt]
\addlinespace[2pt]% \\
agent time (s) & 77.8 & ${+}3.9$ & ${-}11.2$ & ${+}58.9$ & ${-}1.2$ \\
 &  & {\scriptsize [${-}0.1$, ${+}7.9$]} & {\scriptsize [${-}14.7$, ${-}7.8$]} & {\scriptsize [${+}49.4$, ${+}68.3$]} & {\scriptsize [${-}6.5$, ${+}4.0$]} \\[-2pt]
\addlinespace[2pt]% \\
model calls & 16.28 & ${+}0.27$ & ${-}2.17$ & ${+}2.60$ & ${-}0.28$ \\
 &  & {\scriptsize [${-}0.25$, ${+}0.79$]} & {\scriptsize [${-}2.51$, ${-}1.83$]} & {\scriptsize [${+}1.31$, ${+}3.90$]} & {\scriptsize [${-}1.05$, ${+}0.50$]} \\[-2pt]
\addlinespace[2pt]% \\
failed model calls & 0.018 & ${+}0.003$ & ${+}0.123$ & ${+}0.103$ & ${+}0.002$ \\
 &  & {\scriptsize [${-}0.005$, ${+}0.012$]} & {\scriptsize [${+}0.077$, ${+}0.169$]} & {\scriptsize [${+}0.079$, ${+}0.127$]} & {\scriptsize [${-}0.002$, ${+}0.006$]} \\[-2pt]
\addlinespace[2pt]% \\
tool calls & 12.87 & ${+}0.19$ & ${-}1.23$ & ${+}2.52$ & ${-}0.23$ \\
 &  & {\scriptsize [${-}0.31$, ${+}0.70$]} & {\scriptsize [${-}1.42$, ${-}1.05$]} & {\scriptsize [${+}1.64$, ${+}3.41$]} & {\scriptsize [${-}0.92$, ${+}0.45$]} \\[-2pt]
\addlinespace[2pt]% \\
prompt tokens, summed & 136{,}123 & ${+}3{,}585$ & ${+}9{,}332$ & ${+}12{,}785$ & ${-}3{,}332$ \\
 &  & {\scriptsize [${-}1{,}493$, ${+}8{,}662$]} & {\scriptsize [${+}4{,}824$, ${+}13{,}841$]} & {\scriptsize [${+}443$, ${+}25{,}127$]} & {\scriptsize [${-}11{,}389$, ${+}4{,}725$]} \\[-2pt]
\addlinespace[2pt]% \\
completion tokens & 7{,}690 & ${+}440$ & ${-}1{,}253$ & ${+}6{,}707$ & ${-}245$ \\
 &  & {\scriptsize [${-}18$, ${+}899$]} & {\scriptsize [${-}1{,}635$, ${-}871$]} & {\scriptsize [${+}5{,}645$, ${+}7{,}768$]} & {\scriptsize [${-}796$, ${+}305$]} \\
\bottomrule
\end{tabular}
\end{table}

\subsection{What the case study does and does not show}

The four divergences are not small edits. Three of them change between 85\% and 90\% of all requests of a run, and
one removes the model's earlier reasoning from every request. Yet the simultaneous intervals exclude a loss of more
than \SHSimLo{} points and a gain of more than \STSimHi{} points for any of them. For E4B on these tasks, the
solved share is insensitive to prompt details that we had expected to matter, and that public discussion treats as
traps.

Three limits apply. First, the result holds for this model, this agent bundle and these 15 repositories; we did not
run the experiment with the 31B model. Second, the arm without history thoughts shows the largest estimate, takes
almost twice as long per run and ends at the time limit more often. Its estimate therefore depends on the time
limit: with the time-limit rows treated as missing it is $-1.3$ instead of $-2.1$ points. Third, an insensitive solved
share does not make conformance checks useless. A check is cheap, and it removes a question that otherwise needs
4,620 jobs to answer.

The case study also calibrates the rest of this paper. Its reference arm solved 23.8\% of its runs, and with 924 runs per arm the 95\% interval of each difference
reaches 2.2 to 2.5 points to either side of the estimate. A difference between two E4B arms that
is to be seen needs this order of size, and \cref{sec:programmes} shows what happens with fewer runs.

\section{Three attempts to improve the E4B baseline}\label{sec:programmes}

With a runner that could be trusted and a baseline that repeats, we tried three times to raise E4B's solved share.
The three programmes changed different things. The first changed the prompt; one of its variants changed a
thinking-token limit instead. The second changed the way the agent edits files. The third added a sub-agent step. What
exactly was changed is withheld with the bundle until the end of the competition (\cref{sec:setting}); the arms are
named in \cref{tab:arms}, and every run is in \texttt{runs.csv}.

Each programme followed the same four steps.
\begin{enumerate}[leftmargin=*,itemsep=1pt,topsep=2pt]
\item \emph{Build} several variants of the baseline bundle.
\item \emph{Pilot} (programmes 2 and 3 only): run first builds on 15 or 16 tasks, one run each, to find defects.
\item \emph{Screen}: run the variants and the unchanged baseline in one job queue on the 48 tasks of S, two repeats
each, and take the variant with the most solved runs as the candidate.
\item \emph{Confirm}: register the candidate, the tasks, the rule and the analysis script, then run candidate and
baseline once on a frozen panel that shares no task with S, and apply the pass rule of \cref{sec:compare}.
\end{enumerate}
The whole sequence for one programme took 1,152 to 1,402 runs and three to five hours of workstation time.

\subsection{Pilots and screens}\label{sec:screens}

\begin{table}[p]
\centering
\footnotesize
\caption{Pilot, development and screen blocks of the E4B candidates on the workstation. Lead: difference to the baseline of the same block in percentage points. Agent: mean seconds of the agent phase.}
\label{tab:screens}
\begin{tabular}{llrrrrrr}
\toprule
Block & Arm & Tasks & Runs & Solved & \% & Lead & Agent \\
\midrule
\texttt{pilot-2a} & \texttt{v-t1} & 15 & 15 & 4 & 26.7 & --- & 98 \\
\texttt{pilot-2a} & \texttt{v-t2} & 15 & 15 & 5 & 33.3 & --- & 71 \\
\addlinespace
\texttt{pilot-2b} & \texttt{v-t3} & 15 & 15 & 3 & 20.0 & --- & 96 \\
\texttt{pilot-2b} & \texttt{v-t4} & 15 & 15 & 7 & 46.7 & --- & 89 \\
\texttt{pilot-2b} & \texttt{v-t5} & 15 & 15 & 6 & 40.0 & --- & 77 \\
\texttt{pilot-2b} & \texttt{v-t6} & 15 & 15 & 6 & 40.0 & --- & 66 \\
\addlinespace
\texttt{dev-3} & \texttt{base} & 16 & 16 & 4 & 25.0 & --- & 58 \\
\texttt{dev-3} & \texttt{v-a1} & 16 & 16 & 2 & 12.5 & \ensuremath{-}12.5 & 48 \\
\texttt{dev-3} & \texttt{v-a2} & 16 & 16 & 3 & 18.8 & \ensuremath{-}6.2 & 54 \\
\addlinespace
\texttt{screen-1} & \texttt{base} & 48 & 96 & 19 & 19.8 & --- & 64 \\
\texttt{screen-1} & \texttt{v-b1} & 48 & 96 & 22 & 22.9 & +3.1 & 72 \\
\texttt{screen-1} & \texttt{v-p1} & 48 & 96 & 24 & 25.0 & +5.2 & 64 \\
\texttt{screen-1} & \texttt{v-p2} & 48 & 96 & 25 & 26.0 & +6.2 & 75 \\
\texttt{screen-1} & \texttt{v-p3} & 48 & 96 & 22 & 22.9 & +3.1 & 63 \\
\texttt{screen-1} & \texttt{v-p4} & 48 & 96 & 24 & 25.0 & +5.2 & 77 \\
\addlinespace
\texttt{screen-2} & \texttt{base} & 48 & 96 & 21 & 21.9 & --- & 72 \\
\texttt{screen-2} & \texttt{v-t4} & 48 & 96 & 18 & 18.8 & \ensuremath{-}3.1 & 88 \\
\texttt{screen-2} & \texttt{v-t6} & 48 & 96 & 22 & 22.9 & +1.0 & 89 \\
\addlinespace
\texttt{screen-3} & \texttt{base} & 48 & 96 & 20 & 20.8 & --- & 66 \\
\texttt{screen-3} & \texttt{v-a3} & 48 & 96 & 25 & 26.0 & +5.2 & 77 \\
\texttt{screen-3} & \texttt{v-a4} & 48 & 96 & 19 & 19.8 & \ensuremath{-}1.0 & 77 \\
\bottomrule
\end{tabular}
\end{table}

\Cref{tab:screens} lists every pilot and screen block.

\paragraph{Pilots.}
Programme 2 began with two first builds on 15 tasks of the case-study panel (\texttt{pilot-2a}), which solved 4 and
5 of 15, and continued with four rebuilds on the same tasks (\texttt{pilot-2b}), which solved 3, 7, 6 and 6. No
baseline ran in these two blocks. Of the four rebuilds, \arm{v-t4} and \arm{v-t6} went on to the screen. Programme 3
began with two first builds on 16 tasks of the small panel C (\texttt{dev-3}); they solved 2 and 3 of 16 against 4
for the baseline in the same queue, and both were rebuilt before the screen. The 16 tasks of \texttt{dev-3} belong to
the panel of the first confirmation, which had ended by then, and to no later one.

The pilots were meant to find builds that did not work. They ranked nothing. \arm{v-t4} had the best pilot
result, 7 of 15, and then solved fewer runs than the baseline in its screen.

\paragraph{Screens.}
In the first screen all five variants solved more runs than the baseline, 22 to 25 of 96 against 19, and the leader,
\arm{v-p2}, led by $\LeadOne$ points. In the second screen, \arm{v-t6} led by $\LeadTwo$ points and \arm{v-t4}
trailed. In the third, \arm{v-a3} led by $\LeadThree$ points and \arm{v-a4} trailed. All three leaders used more
agent time than the baseline of their screen: 75 against 64~s, 89 against 72~s and 77 against 66~s.

\subsection{Confirmation runs}\label{sec:confirm}

\begin{figure}[t]
\centering
\includegraphics[width=\linewidth]{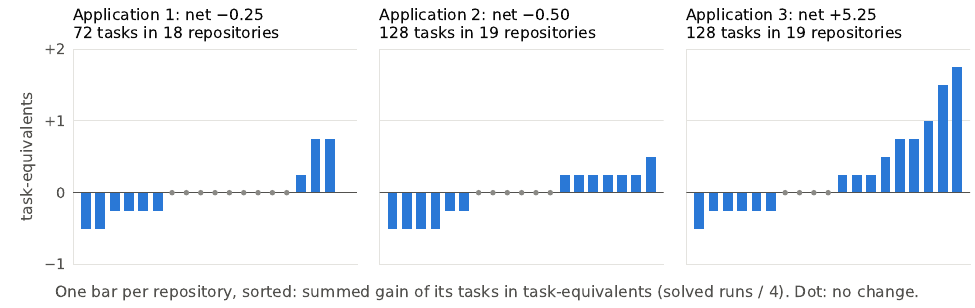}
\caption{Gain of each repository in the three applications of the pass rule (candidate minus baseline). Repositories
are sorted by their total within each application.}
\label{fig:s-gates}
\end{figure}

\begin{table}[!htb]
\centering
\caption{Every application of the pass rule of \cref{sec:compare} to a real candidate, from the frozen gate reports. Baseline and candidate are two versions of the withheld agent bundle.}\label{tab:s-gates}
\footnotesize
\begin{tabular}{lrrr}
\toprule
 & Application 1 & Application 2 & Application 3 \\
\midrule
Task set & small panel & large panel minus small & as application 2 \\
Tasks; repositories; repeats & 72; 18; 4 & 128; 19; 4 & 128; 19; 4 \\
Runs per arm (all with a result row) & 288 & 512 & 512 \\
Solved runs: baseline, candidate & 63, 62 & 103, 101 & 97, 118 \\
Mean gain (points; exact fraction) & ${-}0.35$ (${-}1/288$) & ${-}0.39$ (${-}1/256$) & ${+}4.10$ ($21/512$) \\
\quad (1) gain at least $2/58$ & not met & not met & met \\
Bootstrap-$t$; one-sided $p$ (9{,}999 draws) & ${-}0.175$; 0.6034 & ${-}0.366$; 0.6034 & ${+}1.954$; 0.0405 \\
One-sided $p$ by exact enumeration & 0.6113 & 0.6067 & 0.0360 \\
\quad (2) $p \le 0.025$ & not met & not met & not met \\
Repositories gaining; losing & 3; 6 & 7; 6 & 9; 6 \\
\quad (3a) at least three repositories gain & met & met & met \\
Lowest leave-one-repository-out gain (points) & ${-}1.47$ & ${-}0.85$ & ${+}2.94$ \\
\quad (3b) every leave-one-out gain positive & not met & not met & met \\
Agent time, candidate / baseline (s) & 68.4 / 66.6 & 89.3 / 75.6 & 79.4 / 72.0 \\
\quad (4a) ratio at most 1.2 & 1.026: met & 1.182: met & 1.103: met \\
Infrastructure rows: baseline, candidate & 0, 0 & 0, 0 & 0, 0 \\
\quad (4b) infrastructure errors do not rise & met & met & met \\
Worst repository total (task-equivalents) & ${-}0.50$ & ${-}0.50$ & ${-}0.50$ \\
\quad (5) no repository loses two or more & met & not registered & not registered \\
\midrule
Decision of the frozen rule & \textbf{FAIL} & \textbf{FAIL} & \textbf{FAIL} \\
\bottomrule
\end{tabular}
\end{table}

% written by scripts/reuse_short.py from the sources of the short paper; do not edit
The pass rule of \papersec{5} was applied \SGateCount{} to a real candidate: a changed version of the withheld
agent bundle, tested against the unchanged baseline on the same tasks and repeats. Each application ran once. Its rule,
tasks and analysis script were frozen before its first job, and the outcomes stayed unread until the run had ended.
Table~\ref{tab:s-gates} gives every quantity that the rule uses, and Figure~\ref{fig:s-gates} the gain of every
repository.

\paragraph{How the candidates were chosen.}
Each candidate was the leader of a screen on one separate 48-task set, with two repeats per task: 576 runs over the
baseline and five variants for the first candidate, and 288 runs over the baseline and two variants for each of the
other two. Each leader solved more runs than the baseline in its screen (25, 22 and 25 of 96, against 19, 21 and 20).
A screen selects a candidate and tests nothing, because the leader of several variants is expected to look better in
its screen than it is. No task of the screens is among the tasks of an application. Two small pilots came before the
screens of the later candidates. The variants of the second candidate were first tried on 15 tasks of the case-study
panel, which shares no task with an application. Two predecessors of the third candidate were tried on 16 tasks of the
small panel, that is, on tasks of the first application and on none of the second and third.

\paragraph{Tasks and conditions.}
The first application used the small panel (\SGateOneTasks{} tasks) and all five conditions of \papersec{5}. The
second used the \SGateTwoTasks{} tasks of the large panel that are outside the small one. The third used the same
\SGateThreeTasks{} tasks as the second, and both arms ran again. Its registration stated three things before the run:
that this was the second use of that task set, that the candidate had run on none of these tasks, and that the
candidate was built after the result of the second application was known. It also named two mechanical reads of
earlier runs that went into a repair of the candidate and that may include runs on these tasks: a scan of the
agent's commands in the \SGateTwoSessions{} runs of the second application, and the output of one command in six
earlier runs. Neither read used whether a task was solved. The candidate had been submitted to the competition
once before the run; its hosted score had not arrived when the outcomes were opened, and it is not an input of the
rule. The second and third applications registered conditions (1) to (4) only.

\paragraph{Results.}
The rule asks for a mean gain of at least \SGateMinGain{} points with a one-sided $p$ of at most \SGateAlpha, and for
further conditions on repositories, time and infrastructure errors. No candidate passed. The first two estimates are
below zero: \SGateOneGain{} points (\SGateOneCand{} against \SGateOneBase{} solved runs of \SGateOneRuns; $p =
\SGateOneP$) and \SGateTwoGain{} points (\SGateTwoCand{} against \SGateTwoBase{} of \SGateTwoRuns; $p = \SGateTwoP$).
The third candidate solved \SGateThreeMore{} runs more than the baseline (\SGateThreeCand{} against \SGateThreeBase{}
of \SGateThreeRuns), a gain of \SGateThreeGain{} points. This gain is above the required size, and conditions (3) and
(4) are met: \SGateThreeReposUp{} of the \SGateThreeRepos{} repositories gain, \SGateThreeReposDown{} lose, and the
lowest leave-one-repository-out gain is \SGateThreeLoo{} points. But the one-sided $p$ is \SGateThreeP, above
\SGateAlpha, so condition (2) is not met and the decision is FAIL.

\paragraph{How to read the decisions.}
The verdict of a failed application is ``not confirmed''. It is not evidence that the candidate has no effect: the
registration of the third application estimated, before the run, that a run of this size detects a true gain of the
required size about half of the time. Nor do we call the third candidate an improvement. Its $p$ is not corrected for
the three applications, and it is also above \SGateAlphaSplit, the level divided by three. Its gain is uneven: the
three repositories with the largest totals hold \SGateThreeTopThreeTasks{} of the \SGateThreeTasks{} tasks and gain
\SGateThreeTopThree{} task-equivalents, while the net gain over all repositories is \SGateThreeNet{}
(Figure~\ref{fig:s-gates}). A gain concentrated in a few repositories is the case in which the rule has little
power (\papertab{3}), and it is also what a few repositories can produce by chance. The first two leaders did not keep
their screen lead on new tasks; this is consistent with the selection effect that the rule exists to catch.

\subsection{What a screen lead is worth}\label{sec:leadgain}

\begin{figure}[t]
\centering
\includegraphics[width=0.72\linewidth]{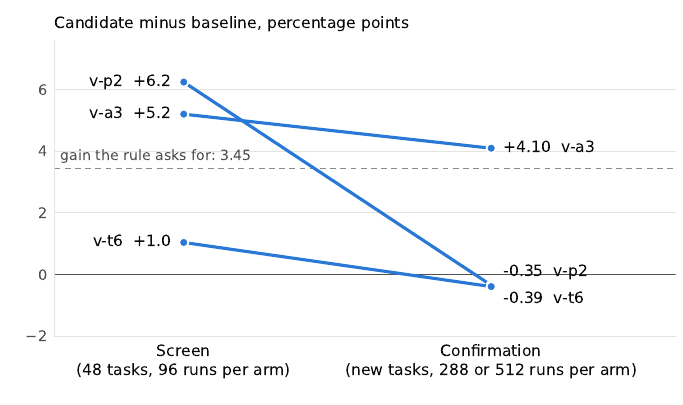}
\caption{Lead of each candidate in its screen on S against its gain in the confirmation run on other tasks, both in
percentage points against the baseline of the same block. The horizontal line is the gain that the pass rule asks
for.}
\label{fig:leadgain}
\end{figure}

The three candidates led their screens by $\LeadOne$, $\LeadTwo$ and $\LeadThree$ points. On new tasks their gains
were $\GainOne$, $\GainTwo$ and $\GainThree$ points (\cref{fig:leadgain}). Two leads vanished; the third candidate
kept most of its lead and still did not pass. We ask here, after the fact, what a lead on S is worth.

\paragraph{The noise of a screen.}
The baseline ran in all three screens, so each task of S has six baseline runs. The variation within tasks gives a
standard deviation of \ScreenSdArmRuns{} runs, or \ScreenSdArmPts{} points, for the total of one arm of 96 runs, and
\ScreenSdDiffPts{} points for the difference of two arms. S is a hard set for this model: \SNeverBase{} of its 48
tasks were solved in none of the six baseline runs, so the whole signal of a screen sits on 14 tasks.

\paragraph{The leader of equal arms.}
Suppose that no variant differs from the baseline, and that every arm total varies independently with the standard
deviation above. Then the best of five variants leads the baseline by \ScreenNullLeadFive{} points on average, and the
best of two by \ScreenNullLeadTwo{} points. A lead as large as the one observed, or larger, has probability
\ScreenNullPOne{} in the first screen (\ScreenLeadRunsOne{} runs, five variants), \ScreenNullPTwo{} in the second
(\ScreenLeadRunsTwo{} run, two variants) and \ScreenNullPThree{} in the third (\ScreenLeadRunsThree{} runs, two
variants). None of the three leads was unusual for a screen in which nothing differs.

That all five variants of the first screen were ahead looks stronger than it is. The five leads share one baseline
total, and that total, 19, is the lowest of the three baseline sessions on S (19, 21, 20). A low draw of the shared
baseline lifts every lead at once.

This calculation is a normal approximation with a noise estimate from six runs per task, and it assumes that a
variant has the same run-to-run noise as the baseline. We give it as a guide for planning, not as a test.

\paragraph{Consequences for the procedure.}
A screen of this size selects a candidate; it does not measure one. The leader's lead has to be discounted by the
selection effect before any plan is built on it, and the first two confirmations show what can remain of it. To rank
variants, the standard deviation of a difference would have to fall from \ScreenSdDiffPts{} points to about one
point, which needs roughly nine times the runs: about 830 runs per arm on tasks like those of S. That is larger than
a confirmation block. We conclude that the screen should be kept small and cheap, as a filter for broken builds and
a tie-break among variants, and that the number of confirmations, not the number of screens, limits what can be
learned.

\paragraph{The third candidate.}
In its confirmation the third candidate solved \ConfCandThree{} runs against \ConfBaseThree{} for the baseline. By
task, it was ahead on \CandThreeTasksUp{} tasks, behind on \CandThreeTasksDown{} and level on
\CandThreeTasksSame{}. The difference of \CandThreeDiffRuns{} runs is \CandThreeNoiseRatio{} standard deviations of
the run-to-run noise of a difference on H (\cref{sec:small}). Against the baseline session of the second
confirmation, which solved \RepeatHFirst{} runs on the same tasks, the difference is \CandThreeDiffEarlier{} runs, or
\CandThreeNoiseRatioEarlier{} standard deviations.

These ratios are post hoc, and they answer a narrower question than the rule does. Run-to-run noise asks whether the
same tasks, run again, would give the same difference. The rule asks whether the gain would hold on other tasks and
repositories, and its test, with repositories as clusters, gives $p = \GatePThree$, above the registered level. The
registered verdict stands: not confirmed. We do not call the third candidate an improvement. H has now been used for
two confirmations, and the third candidate was built after the result of the first of them was known. A further test
of this candidate needs tasks that no experiment has touched.

\subsection{Recomputation from the released rows}

The three confirmation blocks are in \texttt{runs.csv} with the rest. \Cref{sec:app-recheck} recomputes every
quantity of the three decisions from these rows with the released kit and compares it with the frozen reports;
all \RecheckQuantities{} quantities agree. The $p$-values of the first two applications are equal to four decimals,
\GatePOne{} and \GatePTwo{}. The recomputation confirms that this is a coincidence of two different data sets, with
63 against 62 solved runs of 288 in one and 103 against 101 of 512 in the other, and not a copying error.

\section{Cross-experiment analysis}\label{sec:analysis}

The experiments of \cref{sec:big,sec:small,sec:case,sec:programmes} share one task pool, one runner and, for E4B, one
baseline. This section reads across them. It asks how large the noise of each kind of comparison is, which tasks can
move at all, what a hosted score adds to a local result, and what the record implies for the test of a further
candidate. Every analysis in this section was chosen after the results were known. None of them changes a registered
verdict.

\subsection{Three comparisons and their noise}\label{sec:noise}

\begin{table}[t]
\centering
\caption{Standard deviation of the difference of two arms that run on the same tasks, from run-to-run variation
alone. The first two rows are estimated from the repeated baseline sessions (\cref{sec:small}); the third row uses
the repeated public draws of unchanged 31B files \citep{afshar2026census}.}
\label{tab:noise}
\footnotesize
\begin{tabular}{lrrl}
\toprule
Comparison of two arms & Runs per arm & SD, points & Time as run \\
\midrule
Screen on S (48 tasks, 2 repeats) & 96 & \ScreenSdDiffPts{} & 1--2 hours (3--6 arms) \\
Confirmation on H (128 tasks, 4 repeats) & 512 & \NoiseSdDiffPts{} & about 3 hours (2 arms) \\
Hosted, one draw per file (58 public tasks) & 58 & \HostedSdDiffPts{} & 2 days \\
\bottomrule
\end{tabular}
\end{table}

\Cref{tab:noise} puts the three comparisons that we used side by side. All three hold the tasks fixed, so the numbers
answer one question only: would the difference appear again if the same two builds ran again on the same tasks? For
that question a confirmation block on the workstation is the most exact instrument that we have, and a pair of hosted
scores is the least exact.

The pass rule asks a harder question: would the gain hold on tasks from repositories that the block did not contain?
Its test resamples repositories, and its answer can differ widely from the answer for fixed tasks. The third
application shows the size of that difference. The candidate solved \CandThreeDiffRuns{} more runs than the baseline,
a gain of $\GainThree$ points. Against run-to-run noise alone this is \CandThreeNoiseRatio{} standard deviations.
Under the repository-clustered test the one-sided $p$ is \GatePThree{}, above the limit of 0.025
(\cref{sec:leadgain}). Both statements are correct. The first says that the two builds very probably differ on the
128 tasks of H. The second says that the 19 repositories of H do not yet show that the gain carries over to other
repositories.

A third source of error is selection. The candidate that enters a confirmation is the leader of a screen, and the
leader of several equal arms is ahead by construction: \ScreenNullLeadFive{} points on average when five variants are
screened against the baseline, and \ScreenNullLeadTwo{} points when two are (\cref{sec:leadgain}). A reader who compares a screen
lead with \cref{tab:noise} must subtract this first.

\subsection{Tasks that the small model does not solve}\label{sec:never}

On most tasks the outcome of an E4B run does not depend on the run, and it does not depend on the arm. The baseline
solved \HNever{} of the 128 tasks of H in none of its eight runs. Each of these tasks was also run eight times by the
two candidates that were tested on H. They solved \HNeverRescued{} of them at least once, in \HNeverRescuedRuns{} of \HNeverCandRuns{}
runs; \HNeverAny{} tasks stayed unsolved in all 16 runs. On the screen set S the baseline never solved \SNeverBase{}
of 48 tasks in six runs, and \SNeverAny{} tasks stayed unsolved in the 24 runs of all arms of the three screens. Over
all E4B runs on the workstation, \WsTasksAny{} tasks were attempted, each in \WsRunsPerTaskMin{} to
\WsRunsPerTaskMax{} runs. \WsTasksSolvedAny{} of them were solved at least once. The other
\WsTasksNeverPct\% were never solved.

``Never'' is a statement about our runs. A task that is solved in one run of ten stays unsolved in eight runs with a
probability of 43\%. The few candidate runs that solved a task of the never-solved group are therefore what such tasks
would give without any change to the agent, and we do not read them as tasks that a candidate opened.

Two consequences follow. First, the changes that we tested act on a minority of tasks. Even the largest gain,
that of the third candidate, changed the solved count of \CandThreeTasksUp{} tasks upwards and of \CandThreeTasksDown{}
tasks downwards, and left \CandThreeTasksSame{} of 128 as they were. Second, the task set fixes most of a solved share before any arm
runs. The share of tasks that some arm solved in some run, \WsTasksSolvedAny{} of \WsTasksAny{}, is about twice the
baseline's solved share of runs (\BaseWsPct\%). It bounds what any choice among our variants could reach on these
tasks. \Cref{sec:cost} showed how the failures look: in \WsUnsolvedWrongPct\% of the unsolved workstation runs the
agent handed in a patch that the tests reject, and in \WsUnsolvedTimeoutPct\% it was stopped by the time limit.

\subsection{What a hosted score adds}\label{sec:hostednoise}

\paragraph{Precision.}
The census of repeated public draws gives a standard deviation of \HostedSdTasks{} tasks of 58 for one draw of
unchanged 31B files (our recount of its 29 draws of six file sets). The difference of two submissions then has a
standard deviation of \HostedSdDiffPts{} points. To detect a gain of two tasks of 58 with a power of 80\% at a
one-sided level of 0.025, each of the two files would need about \HostedDrawsNeeded{} draws. At one submission per
day this is twenty days for one comparison. The difference of two arms in one confirmation block has about
\LocalBlockRatio{} times less run-to-run variance than the difference of two single hosted draws, and the block takes
three hours. For E4B files no repeated draws exist, so the first number is not known for the small model.

\paragraph{Relevance.}
The hosted score is measured on the tasks that decide the competition. Local runs are not. For the four arms with at
least 70 local runs, the hosted score is between \HostedRatioMin{} and \HostedRatioMax{} of the local solved share
(\cref{tab:hosted}): 0.12 and 0.18 against 54\% and 53\% of the development tasks for the two 31B arms, and 0.06
and 0.05 against 20\% and 23\% of Berth runs for the two E4B arms. The hidden tasks need not resemble either local
set. The development tasks are public, and the Berth tasks are those that load and pass validation. We did not test
which of these reasons accounts for the gap.

\paragraph{Order.}
We have two pairs of arms with both a local and a hosted result. \arm{W3} and \arm{BF5} tied locally (38 and 37 of
70) and scored 0.12 and 0.18. The third E4B candidate was ahead of the baseline locally by $\GainThree$ points and
scored 0.05 against 0.06. In both pairs the hosted order is the reverse of the local order. The hosted difference of
the 31B pair is about one and a half standard deviations of a difference of single draws (\cref{sec:hosted}); that
of the E4B pair is 0.01. Neither can be told from noise. Two pairs cannot show whether a local gain carries over to
the hidden tasks. They show that one hosted draw per arm cannot answer the question.

\subsection{Time limits and the two models}\label{sec:limits-models}

The time limit of a session shapes the results of the large model and hardly touches the small one. On the development
tasks under emulated latency, \TimeoutBigDevPct\% of the 31B runs were stopped by the limit; on Berth tasks on
4$\times$L4, \TimeoutBigBerthPct\%. Of the E4B runs on the workstation, \TimeoutSmallWsPct\% were stopped by the time
limit and \BudgetSmallWsPct\% by the limit on tool calls (\cref{tab:cost}). For 31B the limit binds in about one
run of eight, so the way an arm uses its time is part of its result. The first wrapped arm, for example, finished its
jobs sooner than the reference because it wrote no history summaries (\cref{sec:big-arms}). For E4B, time is not the
binding limit. The three candidates used \RecheckTimeOne{}, \RecheckTimeTwo{} and \RecheckTimeThree{} times the
agent time of the baseline, inside the factor of 1.2 that the pass rule allows. With so few runs stopped by a limit,
a faster E4B agent could gain at most that fraction of runs. A change that helps must alter the patch that the agent
submits.

\subsection{What the record implies for a further candidate}\label{sec:next}

\begin{enumerate}[leftmargin=*,itemsep=2pt,topsep=3pt]
\item \textbf{Confirmations are the scarce resource.} A screen costs an hour and ranks nothing
(\cref{sec:leadgain}). A confirmation costs about a thousand runs and three hours, and it spends its panel: H has
been used twice, and the third candidate was built after the first result on H was known.
\item \textbf{The validated pool is nearly used up.} Of the \PoolValid{} validated tasks, \PoolUsed{} have been
used by some experiment of this paper. The remaining \PoolUntouched{} tasks lie in \PoolUntouchedRepos{}
repositories; one repository holds \PoolUntouchedTopRepo{} of them and \PoolUntouchedSmallRepos{} repositories hold
one task each. The pass rule asks for three repositories with a gain, so these tasks alone cannot carry a
confirmation. A further \CaseOnlyTasks{} tasks were used only by the case study of \cref{sec:case}, which compared
prompt renderings and no candidate. They can serve as a fresh panel only for a candidate whose design did not draw
on the per-task results of that study.
\item \textbf{The size of the panel sets the power.} For an even gain of three tasks of 58 the simulated pass rate
of the rule is about 60\% on the small panel and about 90\% on the large one (\cref{tab:simulated}). A gain that is
concentrated in few repositories passes far less often.
\item \textbf{A hosted draw is a check of the file.} It shows that a bundle loads and runs under the scorer. It does
not show which of two bundles is better (\cref{sec:hostednoise}).
\end{enumerate}

\section{Related work}\label{sec:related}

\paragraph{Benchmarks, agents and task pools.}
Repository-level coding benchmarks grade an agent by running tests in an environment built for each task. SWE-bench
holds 2,294 tasks from 12 Python repositories \citep{jimenez2024swebench}, and a human-validated subset of 500 tasks is
in wide use \citep{openai2024swebenchverified}. Agent frameworks and training environments bring their own sandboxes
and task pools \citep{yang2024sweagent,wang2025openhands,pan2025swegym,yang2025swesmith}, and SWE-rebench keeps
collecting new tasks so that an evaluation can use tasks that a model has not seen
\citep{badertdinov2025swerebench}. For this competition, Berth \citep{patel2026berth} and FreshSWE-Py
\citep{moshegolderson2026freshswe} supply tasks in the format of the harness, and a third data set reports on the
soundness of tasks together with local results \citep{gluzdov2026soundness}. We built no task set. All tasks of our
experiments come from the competition's development set and from Berth, and we add grading controls for every Berth
task that we used (\cref{tab:controls}).

\paragraph{Evaluation practice.}
Audits of agent benchmarks show that validity has to be checked and cannot be assumed \citep{zhu2025abc}, that scores
move with the evaluation setup and with the format of the prompt \citep{biderman2024trenches,sclar2024formatting},
and that agent evaluations are often hard to reproduce \citep{kapoor2024agentsthatmatter}. We address one narrow case
of this: whether a local stack reproduces one hosted scorer. We compare prompts as token ids and not as message text,
because equal strings need not give equal tokens \citep{agl2025retokenization} and because the control tokens of a
chat template matter \citep{hf2026chattemplates}. The checks run on the serving engine of the scorer
\citep{kwon2023vllm}. The prompt-construction divergences that we measure in \cref{sec:case} are documented upstream
\citep{adk2026toolresponses,google2026gemma4prompt,dimondev2026hf44,gerganov2026hf135}. We did not discover them. We
add executable checks for them and a measurement of their effect on E4B.

\paragraph{Isolation.}
MicroVMs \citep{agache2020firecracker} and user-space kernels \citep{young2019gvisor} are alternatives to plain
containers that are built for security, and the overhead of the latter has been measured \citep{young2019gvisor}. Our
runner uses plain containers; we tried neither alternative. Containers can consume resources outside their own
accounting \citep{gao2019houdini}, which is the kind of route through shared host resources that our verifiers used
(\cref{sec:contain}).
% written by scripts/reuse_short.py from the sources of the short paper; do not edit
SandboxEval \citep{rabin2025sandboxeval} and SandboxEscapeBench \citep{marchand2026sandboxescapebench} test sandboxes and models' ability to escape them.

Several paper-track resources overlap this one (described from their public text only): LAB \citep{marcofpereira2026lab}, a Docker sandbox with 11 tests and a bundle gate with 16 detectors validated by fault injection; Szeto et al.'s cloud rig with 141 run folders \citep{szeto2026rig}; gemma4-swe-kit \citep{damsolanke2026swekit}, a proxy reproducing the scorer's template, call parsing and compaction on other servers; a dataset card reporting count parity on 107 captured vLLM requests \citep{tyfeng2026template}; Silent Zeros \citep{houchiunyo2026silentzeros}, a bundle checker tested on 22 broken bundles; and ThoughtKeep \citep{hemantkumarbk2026thoughtkeep}, 1,470 Gemma 4 31B attempts on thought retention. We did not find in those texts, and offer here: a runner attacked by independent verifiers with its canary misses published, token-id comparison backed by a mutation suite, and a trap measurement on E4B with a per-job prompt check.

\paragraph{Statistics of comparisons.}
We follow the advice to treat an evaluation as an experiment, with clustered and paired errors and a power analysis
\citep{miller2024errorbars}, and to size decisions to measured noise, as in signal-to-noise analyses of benchmarks
\citep{heineman2025signalnoise}. The test of the pass rule is a wild cluster bootstrap-t over repositories
\citep{cameron2008bootstrap} with Rademacher weights \citep{davidson2008wild}. Few or unbalanced clusters can break
the nominal size of such tests \citep{mackinnon2017wild,canay2021wild}, and with few clusters only a small set of
$p$-values can be attained \citep{webb2023reworking}; for that reason the rule's size and power were simulated on the
frozen panels (\cref{tab:simulated}). The estimates of leaderboard noise for this competition come from third parties
\citep{afshar2026census,mobassir2026toosmall}. What this version adds to them is the local counterpart: the run-to-run
noise of one fixed agent on fixed tasks, measured by repeated sessions (\cref{sec:noise}), and three worked cases of
a screen lead that was tested on new tasks (\cref{sec:leadgain}).

\section{Limitations}\label{sec:limits}

The limitations of the kit and of the case study are those of the short paper.

% written by scripts/reuse_short.py from the sources of the short paper; do not edit
\begin{itemize}
\tightlist
\item
  \textbf{Solve-level parity is not refuted, not established.} The evidence (Figure~\ref{fig:parity}A) is one 24-task list, one repeat per run; ten runs solved only six distinct tasks.
\item
  \textbf{Kaggle-side facts are partly inferred}: the hosted google-adk release, tool-schema equality, and the identity of the hosted template and weights with the public Kaggle model.
\item
  \textbf{Environment differences remain.} One RTX 5090 instead of four L4s; 2 CPUs and 10 GiB per job; a bare \texttt{pytest} command, used in 30 of 216 commands in our Kaggle traces, is absent locally.
\item
  \textbf{The case study} covers one model, one bundle, one google-adk release and one configuration per divergence. Arm T swaps a whole template; in arm H thoughts still reach compaction summaries.
\end{itemize}

The record of experiments adds the following.

\begin{itemize}[leftmargin=*,itemsep=2pt,topsep=3pt]
\item \textbf{Rows that are missing.} One 31B block, an arm with a ten-minute limit that solved 18 of 30 development
tasks, has no result rows that we could retrieve. It is in no table and in no released file. The rows of one
workstation parity run under runner version 2 (4 of 24 solved) were not kept either; that run appears only in
\cref{fig:parity}.
\item \textbf{One run per task for 31B.} On the development tasks every 31B configuration ran each task once.
Differences between configurations and differences between runs cannot be separated there (\cref{sec:big}).
\item \textbf{Arms and sessions are confounded in places.} Some 31B arms ran in blocks of their own, under a
different load. The E4B candidates always ran in one job queue with the baseline.
\item \textbf{Task sets were not always chosen by a recorded rule.} The tasks of the two 31B blocks on real hardware
were picked with earlier results in view. For the eight tasks of block \texttt{berthE4-k2} the method of selection
was not recorded. The screen set S is not fresh: 32 of its 48 tasks had been run in earlier blocks. The development
block of the third programme used 16 tasks of the small panel after that panel's confirmation.
\item \textbf{The hold-out H was used twice.} The third candidate was built after the result of the first
confirmation on H was known. Its confirmation was registered with this disclosure, and its verdict is ``not
confirmed''.
\item \textbf{Post hoc analyses.} The noise estimates, the selection effect of \cref{sec:leadgain} and all of
\cref{sec:analysis} were chosen after the results were known. They describe the record. They are not tests.
\item \textbf{One baseline.} The share of tasks with an uncertain outcome, and with it the run-to-run noise, was
measured for one E4B agent on Berth tasks. Another agent, another model or another task source can have a different
share.
\item \textbf{Hosted scores are single draws.} We have one hosted draw per file and no repeated draw of an E4B file.
\item \textbf{Machines differ between experiments.} The 31B runs on development tasks used emulated latency on a
single GPU, the first E4B blocks used hosted notebooks with four L4 GPUs, and most E4B runs used the workstation.
Times are comparable within a block and not across machines.
\item \textbf{Arms cannot be rebuilt from this paper.} Prompts, the E4B bundle and the adapter are withheld
(\cref{sec:setting}), and each arm is described by the kind of its change. The rows of the development tasks carry
opaque task codes, because the competition data may not be redistributed; they support counts and pairings and
cannot be joined with the task texts.
\end{itemize}

\section{Availability and released data}\label{sec:avail}

% written by scripts/reuse_short.py from the sources of the short paper; do not edit
\texttt{drydock-eval} 0.4.0 (Apache-2.0) has three modules: \texttt{conform} (checks, mock server, mutation suite), \texttt{contain} (a reference runner tree, its escape tests, a gate helper) and \texttt{compare} (pass rule, sign-flip tests, simulations). Its 435 CPU tests pass. The released runner has not been run in a container as released: the 200 of 200 result is for the deployed tree, of which the release is an identifier-substituted copy with two canaries rewritten. Code: \url{https://github.com/MinhThanghus/drydock-eval} (tag \texttt{v0.4.0}; branch \texttt{paper}: supplement, \cref{sec:case} outcome rows). The package also carries copies of the \cref{sec:case} protocol, analysis code and job-list generator, redacted where needed, each with its original's sha256; the agent bundle and request dumps are planned for 2026-12-03.

\paragraph{Data of this version.}
The arXiv submission of this extended version holds its sources and its figures; its ancillary files hold the
scripts named below and the data behind every table and figure (\cref{sec:app-data}): \texttt{runs.csv} with one row for each of the \RunsTotal{}
agent runs of \cref{sec:big,sec:small,sec:programmes}, \texttt{controls.csv} with the \ControlRows{} grading
controls, \texttt{serving\_bench.csv} with the \ServeRows{} serving measurements, the tables of blocks, arms and
hosted submissions, the three applications of the pass rule, and the \CaseRows{} outcome rows of the case study. A row
of \texttt{runs.csv} gives the block, the arm, the task and its repository, the outcome, the way the session ended,
times, token counts and the number of tool calls. It holds no prompt, no model output, no patch and no error text.

\paragraph{Scripts.}
\texttt{scripts/analyze\_long.py} computes every number and every table of
\cref{sec:serving,sec:big,sec:small,sec:programmes,sec:analysis} from these files and stops if a check against the
rows fails. \texttt{scripts/recheck\_gates.py} applies the pass rule of the released kit to the rows of the three
confirmation blocks and compares the result with the registered decisions (\cref{sec:app-recheck}).
\texttt{figsrc/figs\_long.py} draws the figures. The passages taken from the short paper are copied by
\texttt{scripts/reuse\_short.py} and are not edited by hand.

\paragraph{Withheld.}
Not released are the prompts, the agent bundles and their changed versions, the adapter and its training data, the
request dumps, and the table that maps the task codes of the development set to the tasks of the competition.

\paragraph{Licences.} Code: Apache-2.0. Text, figures and data: CC BY 4.0.

\section{Conclusion}\label{sec:conclusion}

\kit{} gives a team with one GPU three things: a runner that has been attacked and whose misses are published, checks
that compare a local serving stack with the scorer down to token ids, and a pass rule with known error rates. This
extended version adds what the kit was used for: \RunsTotal{} agent runs in \BlocksTotal{} blocks and \ArmsTotal{}
arms, and the \CaseRows{} rows of the case study.

Most of what the record shows is the absence of a difference. Six configurations of a 31B agent lay between
\PubMinSeventy{} and \PubMaxSeventy{} of 70 development tasks, and no pairwise difference was detected. Four
prompt divergences changed the E4B solve rate by no detectable amount. Three changed E4B agents led their screens and
were not confirmed on new tasks, with gains of $\GainOne$, $\GainTwo$ and $\GainThree$ points. The third of them is
the closest case: its gain is large against run-to-run noise and does not pass the test across repositories. We do
not call it an improvement.

The record also explains why this is the expected outcome at this scale. On a fixed set of tasks the small model is
nearly deterministic at the level of a task: about two thirds of the tasks were solved in no run, and the baseline
repeated its total within \RepeatHDiff{} points. A screen of 96 runs per arm cannot rank variants, and its leader is
ahead by selection. A hosted score is one draw with a standard deviation near \HostedSdTasks{} tasks of 58. The
instrument that can decide is a confirmation on tasks from repositories that no earlier experiment has used, and
each one spends its tasks. The validated pool that we used is almost spent.

We release every run-level result so that others can recompute these numbers, plan the size of their own
comparisons from the measured noise, and avoid paying again for the same screens.

\bibliographystyle{plainnat}
\bibliography{refs}

\appendix
\crefalias{section}{appendix}
\appendix

\section{Arms and blocks}\label{sec:app-arms}

\Cref{tab:arms} lists every arm with the number of its runs and what was changed against the arm it was built from.
\Cref{tab:blocks} lists every block. The same two tables are in the released files \texttt{arms.csv} and
\texttt{experiments.csv}.

\begin{table}[p]
\centering
\footnotesize
\caption{Arms of the experiments. Labels are neutral; the category names what was changed against the arm it was built from. Prompt text and bundle files are not released.}
\label{tab:arms}
\begin{tabular}{llrp{0.58\linewidth}}
\toprule
Arm & Model & Runs & What it is \\
\midrule
\texttt{BF5} & 31B & 142 & Single agent: the public Budget-Fit V5 configuration, run unchanged (\emph{reference}) \\
\texttt{BF5-p} & 31B & 30 & BF5 with one changed section of the task prompt (\emph{prompt}) \\
\texttt{BF5-c} & 31B & 32 & BF5 with a prompt-only checking step before the patch is submitted (\emph{prompt}) \\
\texttt{BF5-s} & 31B & 10 & BF5 with changed sampling settings (\emph{sampling}) \\
\texttt{W1} & 31B & 30 & Wrapped: a root agent calls a coder sub-agent as a tool; the coder has its own context (\emph{architecture}) \\
\texttt{W2} & 31B & 40 & W1 with a bounded restart of the coder after a malformed tool call (\emph{architecture}) \\
\texttt{W3} & 31B & 70 & W2 with a root that can only pass the request on and restart the coder (\emph{architecture}) \\
\texttt{W1-b} & 31B & 30 & W1 with a shorter time limit, more tool calls and half the output limit per call (\emph{budget}) \\
\texttt{W1-p} & 31B & 30 & W1 with a rewritten coder prompt, a shorter time limit and half the output limit per call (\emph{prompt}) \\
\texttt{SH1} & 31B & 30 & Wrapped, coder with a shell tool only; first version (\emph{architecture}) \\
\texttt{SH3} & 31B & 40 & SH1, third version of prompt and limits (\emph{architecture}) \\
\texttt{SH5} & 31B & 112 & SH1, fifth version of prompt and limits (\emph{architecture}) \\
\texttt{SH6} & 31B & 94 & SH5 with an added analysis step in the prompt (\emph{prompt}) \\
\texttt{ID24} & 31B & 10 & Public bundle of Team Muhammed Idris, version 24, run unchanged (\emph{reference}) \\
\texttt{ID24-r} & 31B & 10 & ID24 with one file-writing tool removed, a shorter command timeout, added prompt sections (\emph{tools}) \\
\texttt{e-single} & E4B & 24 & Single agent (\emph{architecture}) \\
\texttt{e-wrap} & E4B & 24 & Wrapped, as W1 (\emph{architecture}) \\
\texttt{e-wrap-ad} & E4B & 24 & e-wrap with a fine-tuned adapter on the coder (withheld) (\emph{adapter}) \\
\texttt{e-BF} & E4B & 64 & The public single-agent prompt adapted to E4B (six tools, 28 calls) (\emph{prompt}) \\
\texttt{e-BF-s} & E4B & 16 & e-BF with changed sampling settings (\emph{sampling}) \\
\texttt{base} & E4B & 1{,}848 & The withheld E4B agent bundle; the baseline of every later comparison (\emph{baseline}) \\
\texttt{base-p} & E4B & 16 & base with one prompt change (\emph{prompt}) \\
\texttt{v-p1} & E4B & 96 & base with one prompt-only change (\emph{prompt}) \\
\texttt{v-p2} & E4B & 384 & base with one prompt-only change (candidate 1) (\emph{prompt}) \\
\texttt{v-p3} & E4B & 96 & base with one prompt-only change (\emph{prompt}) \\
\texttt{v-p4} & E4B & 96 & base with one prompt-only change (\emph{prompt}) \\
\texttt{v-b1} & E4B & 96 & base with a changed thinking-token limit (\emph{budget}) \\
\texttt{v-t1} & E4B & 15 & base with a changed file-editing path, first build (\emph{tools}) \\
\texttt{v-t2} & E4B & 15 & as v-t1, second design, first build (\emph{tools}) \\
\texttt{v-t3} & E4B & 15 & rebuild of v-t1, option a (\emph{tools}) \\
\texttt{v-t4} & E4B & 111 & rebuild of v-t1, option b (\emph{tools}) \\
\texttt{v-t5} & E4B & 15 & rebuild of v-t2, option a (\emph{tools}) \\
\texttt{v-t6} & E4B & 623 & rebuild of v-t2, option b (candidate 2) (\emph{tools}) \\
\texttt{v-a1} & E4B & 16 & base with an added sub-agent step, first design, first build (\emph{architecture}) \\
\texttt{v-a2} & E4B & 16 & as v-a1, second design, first build (\emph{architecture}) \\
\texttt{v-a3} & E4B & 608 & rebuild of v-a1 (candidate 3) (\emph{architecture}) \\
\texttt{v-a4} & E4B & 96 & rebuild of v-a2 (\emph{architecture}) \\
\bottomrule
\end{tabular}
\end{table}

\begin{table}[p]
\centering
\footnotesize\setlength{\tabcolsep}{4.5pt}
\caption{Every block of agent runs in the released file \texttt{runs.csv}. A block is one session on one machine; its arms ran in one job queue. Machine: one GPU with emulated L4 latency (emulated), a hosted notebook with four L4 GPUs (4$\times$L4), or the single-GPU workstation. Source: the development tasks of the competition (dev.) or Berth. Rep.: repeats per task and arm.}
\label{tab:blocks}
\begin{tabular}{llllrrrrp{0.27\linewidth}}
\toprule
Block & Model & Machine & Source & Tasks & Arms & Rep. & Runs & Note \\
\midrule
\texttt{pub31-1} & 31B & emulated & dev. & 30 & 3 & 1 & 90 & first 30 development tasks \\
\texttt{pub31-2} & 31B & emulated & dev. & 30 & 1 & 1 & 30 & same 30 tasks \\
\texttt{pub31-3} & 31B & emulated & dev. & 30 & 1 & 1 & 30 & same 30 tasks \\
\texttt{pub31-4} & 31B & emulated & dev. & 30 & 2 & 1 & 60 & same 30 tasks, both arms in one queue \\
\texttt{pub31-5} & 31B & emulated & dev. & 40 & 2 & 1 & 80 & 40 further development tasks \\
\texttt{pub31-6} & 31B & emulated & dev. & 40 & 2 & 1 & 80 & same 40 tasks, both arms in one queue \\
\texttt{pub31-7} & 31B & emulated & dev. & 70 & 1 & 1 & 70 & all 70 tasks \\
\texttt{pub31-8} & 31B & emulated & dev. & 70 & 3 & 1 & 130 & all 70 tasks; 30 tasks with masked test runners \\
\texttt{l4-1} & 31B & 4$\times$L4 & dev. & 10 & 1 & 1 & 12 & 10 tasks, plus 2 masked \\
\texttt{l4-2} & 31B & 4$\times$L4 & dev. & 24 & 1 & 1 & 24 & 24 tasks \\
\texttt{pubE4-1} & E4B & 4$\times$L4 & dev. & 24 & 3 & 1 & 72 & 24 development tasks with small fixes \\
\texttt{berth31-time} & 31B & 4$\times$L4 & Berth & 10 & 4 & 1 & 40 & 10 tasks, serial \\
\texttt{berth31-check} & 31B & 4$\times$L4 & Berth & 16 & 2 & 2 & 64 & 16 tasks \\
\texttt{berthE4-k1} & E4B & 4$\times$L4 & Berth & 24 & 2 & 2 & 96 & 24 tasks \\
\texttt{berthE4-k2} & E4B & 4$\times$L4 & Berth & 8 & 4 & 2 & 64 & 8 tasks \\
\texttt{parity-1} & E4B & workstation & Berth & 24 & 1 & 1 & 24 & no sandbox, 8 workers \\
\texttt{parity-2} & E4B & workstation & Berth & 24 & 1 & 1 & 24 & no sandbox, 1 worker \\
\texttt{parity-3} & E4B & workstation & Berth & 24 & 1 & 1 & 24 & container runner, version 1 \\
\texttt{parity-4} & E4B & workstation & Berth & 24 & 1 & 1 & 24 & container runner, version 3 \\
\texttt{parity-5} & E4B & workstation & Berth & 24 & 1 & 1 & 24 & container runner, version 4, run a \\
\texttt{parity-6} & E4B & workstation & Berth & 24 & 1 & 1 & 24 & container runner, version 4, run b \\
\texttt{parity-7} & E4B & workstation & Berth & 24 & 1 & 1 & 24 & container runner, version 7 (final) \\
\texttt{screen-1} & E4B & workstation & Berth & 48 & 6 & 2 & 576 & screen set S \\
\texttt{confirm-1} & E4B & workstation & Berth & 72 & 2 & 4 & 576 & small panel C \\
\texttt{pilot-2a} & E4B & workstation & Berth & 15 & 2 & 1 & 30 & 15 tasks of the case-study panel \\
\texttt{pilot-2b} & E4B & workstation & Berth & 15 & 4 & 1 & 60 & same 15 tasks \\
\texttt{screen-2} & E4B & workstation & Berth & 48 & 3 & 2 & 288 & screen set S \\
\texttt{confirm-2} & E4B & workstation & Berth & 128 & 2 & 4 & 1{,}024 & panel H \\
\texttt{dev-3} & E4B & workstation & Berth & 16 & 3 & 1 & 48 & 16 tasks of panel C \\
\texttt{screen-3} & E4B & workstation & Berth & 48 & 3 & 2 & 288 & screen set S \\
\texttt{confirm-3} & E4B & workstation & Berth & 128 & 2 & 4 & 1{,}024 & panel H, second use \\
\bottomrule
\end{tabular}
\end{table}

\section{Recomputation of the case study with the released kit}\label{sec:app-recompute}

% written by scripts/reuse_short.py from the sources of the short paper; do not edit
The released table \texttt{e8\_compare\_table.csv} is in the input format of the \texttt{compare} module of the kit.
From a clone of the repository, two commands recompute the analysis:

\begin{verbatim}
pip install "./drydock-eval[compare]"
drydock-eval compare signflip e8_compare_table.csv --reference F --repeats 4
\end{verbatim}

We ran this command from the released source tree with the script \texttt{supp/reproduce.py}, which stops on any
mismatch. For all four arms the command returned the registered estimate, the exact $p$ as the same fraction, the
Holm-adjusted $p$ and every repository-level quantity of Table~\ref{tab:s-repolevel}.

Two quantities depend on Monte Carlo draws: the inverted intervals, and the Monte Carlo $p$ that is reported next to
the exact one. The draws are frozen (seed \SReproSeed), but which draw meets which task depends on the order of the
tasks. The registered analysis used the order of the panel file; the command sorts the task identifiers. With the
tasks in panel order, the routine of the kit returns the registered values to the last digit
(Table~\ref{tab:s-repro}). With the order of the command, the ends of the 95\% intervals move by at most
\SReproMoveNinetyFive{} points and those of the simultaneous intervals by at most \SReproMoveSim{} points. The largest
move is the upper end of the interval of H, $+\SReproHCmdHi$ instead of $+\SReproHRegHi$ points. The $p$-value of the
test is a step function of the shift, and across that stretch it stays between \SReproStepLo{} and \SReproStepHi{},
next to the 0.05 it is compared with. Both intervals include zero. The paper reports the registered intervals.

\begin{table}[!htb]
\centering
\caption{Reproduction with the released kit: the two quantities that depend on Monte Carlo draws. 95\% intervals of the change against F in percentage points, and the Monte Carlo $p$ that accompanies the exact $p$.}\label{tab:s-repro}
\footnotesize
\begin{tabular}{lrrrrrr}
\toprule
 & \multicolumn{3}{c}{95\% interval} & \multicolumn{3}{c}{Monte Carlo $p$} \\
\cmidrule(lr){2-4}\cmidrule(lr){5-7}
Arm & Registered & Kit, panel order & Kit command & Registered & Kit, panel order & Kit command \\
\midrule
R & [${-}2.82$, ${+}2.17$] & [${-}2.82$, ${+}2.17$] & [${-}2.80$, ${+}2.16$] & 0.8650 & 0.8650 & 0.8642 \\
C & [${-}2.03$, ${+}2.48$] & [${-}2.03$, ${+}2.48$] & [${-}2.04$, ${+}2.48$] & 0.9243 & 0.9243 & 0.9268 \\
H & [${-}4.45$, ${+}0.38$] & [${-}4.45$, ${+}0.38$] & [${-}4.45$, ${+}0.25$] & 0.1106 & 0.1106 & 0.1095 \\
T & [${-}1.42$, ${+}2.94$] & [${-}1.42$, ${+}2.94$] & [${-}1.42$, ${+}2.94$] & 0.5599 & 0.5599 & 0.5624 \\
\bottomrule
\end{tabular}
\end{table}

\section{The three decisions recomputed from the released rows}\label{sec:app-recheck}

\begin{table}[t]
\centering
\footnotesize\setlength{\tabcolsep}{3.6pt}
\caption{The three applications of the pass rule (four repeats per task and arm). Gain: candidate minus baseline in percentage points. $p$: one-sided, repository-clustered. $+/-$: repositories with a positive or negative summed gain. LOO: lowest gain when one repository is left out. Time: ratio of mean agent time, candidate to baseline.}
\label{tab:confirm}
\begin{tabular}{rlrrrrrrrrrrl}
\toprule
No. & Candidate & Tasks & Repos & Runs/arm & Base & Cand. & Gain & $p$ & $+/-$ & LOO & Time & Verdict \\
\midrule
1 & \texttt{v-p2} & 72 & 18 & 288 & 63 & 62 & \ensuremath{-}0.35 & 0.6034 & 3/6 & \ensuremath{-}1.47 & 1.03 & not confirmed \\
2 & \texttt{v-t6} & 128 & 19 & 512 & 103 & 101 & \ensuremath{-}0.39 & 0.6034 & 7/6 & \ensuremath{-}0.85 & 1.18 & not confirmed \\
3 & \texttt{v-a3} & 128 & 19 & 512 & 97 & 118 & +4.10 & 0.0405 & 9/6 & +2.94 & 1.10 & not confirmed \\
\bottomrule
\end{tabular}
\end{table}

The frozen reports of the three applications of the pass rule were written by the registered analysis scripts on the
machine that ran the jobs, from result files that are not released. The released file \texttt{runs.csv} holds the
same runs with a reduced set of columns. The script \texttt{scripts/recheck\_gates.py} builds, for each application,
the input table of the \texttt{compare} module from these rows, calls the routines of the kit at tag
\texttt{v0.4.0} with the registered thresholds and seed, and compares the result with the released copy of the
frozen report, \texttt{pass\_rule\_applications.json}. It stops on the first mismatch.

All \RecheckQuantities{} compared quantities agree: for each application the solved counts of both arms, the gain,
the bootstrap $p$ with its 9,999 draws, the gain and the leave-one-out estimate of every repository, the counts of
repositories that gain and lose, and the decision. \Cref{tab:confirm} gives the values. The time ratio is the one
quantity that the recomputation can only approximate, because the reports computed it from timings with more
digits than the released rows keep; recomputed from the released rows it is \RecheckTimeOne{}, \RecheckTimeTwo{} and
\RecheckTimeThree{}, which round to the reported values.

\section{Released data}\label{sec:app-data}

\begin{table}[t]
\centering
\footnotesize
\caption{Data files released with this paper, among the ancillary files of its arXiv submission, in the directory
\texttt{data/}.}
\label{tab:datafiles}
\begin{tabular}{@{}lrp{0.56\linewidth}@{}}
\toprule
File & Records & Content \\
\midrule
\texttt{runs.csv} & \RunsTotal & one row per agent run of \cref{sec:big,sec:small,sec:programmes} \\
\texttt{controls.csv} & \ControlRows & one row per control grading (reference patch or empty patch) \\
\texttt{serving\_bench.csv} & \ServeRows & one row per cell of the serving benchmark of \cref{sec:serving} \\
\texttt{serving\_servers.csv} & 7 & start-up time and cache size of each model server of the benchmark \\
\texttt{experiments.csv} & \BlocksTotal & one row per block (\cref{tab:blocks}) \\
\texttt{arms.csv} & \ArmsTotal & one row per arm (\cref{tab:arms}) \\
\texttt{submissions.csv} & 5 & our hosted submissions with their displayed scores \\
\texttt{pass\_rule\_applications.json} & 3 & the frozen report of each application of the pass rule \\
\texttt{e8\_runs.csv} & \CaseRows & one row per job of the case study, with the rows of interrupted jobs \\
\texttt{e8\_tasks.csv} & 231 & the tasks of the case study with their solved counts per arm \\
\texttt{e8\_repositories.csv} & 15 & the same per repository \\
\texttt{e8\_compare\_table.csv} & 231 & the case study in the input format of the \texttt{compare} module \\
\texttt{e8\_summary.json} & 1 & the estimates, intervals and $p$-values of the case study \\
\bottomrule
\end{tabular}
\end{table}

\Cref{tab:datafiles} lists the files. \texttt{data/README.md} describes every column, and \texttt{data/SHA256SUMS}
holds the checksums. The file \texttt{runs.csv} has 29 columns:
\begin{description}[leftmargin=1.2em,itemsep=0pt,topsep=2pt,font=\normalfont\ttfamily]
\item[experiment, block] the experiment family and the block (\cref{tab:blocks}).
\item[model, machine, task\_source] 31B or E4B; the machine of \cref{sec:setting}; development tasks or Berth.
\item[arm, condition] the arm (\cref{tab:arms}); \texttt{standard}, or \texttt{masked\_test\_runners} for the runs
of \cref{tab:masked}.
\item[task, repository, repeat] the task (an opaque code for development tasks, the Berth identifier otherwise), its
repository, and the number of the repeat within the block.
\item[resolved] 1 if the tests of the task passed on the patch that the run left behind, else 0.
\item[end\_reason, outcome\_id] how the run ended (normally, at the time limit, at the tool-call limit, without a
submission), and the outcome class of the runner.
\item[wall\_s, t\_agent\_s, t\_setup\_s, t\_verify\_s] seconds of the whole job and of its three phases, where the
harness recorded them.
\item[emulation\_pad\_s] seconds of padding added to model calls on the emulated-latency machine.
\item[model\_calls, model\_calls\_failed, tool\_calls] counts per run.
\item[tokens\_in, tokens\_out, max\_prompt\_tokens] prompt tokens summed over the requests of the run, generated
tokens, and the longest prompt.
\item[summaries] number of history summaries that the harness made in the run.
\item[patch\_chars, test\_exit] length of the final patch, and the exit status of the test command.
\item[start\_utc, end\_utc] times of the job, where recorded.
\end{description}
No column holds prompt text, model output, error text, a path or a machine name. The columns of
\texttt{e8\_runs.csv} are listed in \cref{tab:s-columns}.

\begin{table}[!htb]
\centering
\caption{Columns of \texttt{e8\_runs.csv}. The rows of interrupted jobs hold the first ten columns and the two times of day only.}\label{tab:s-columns}
\footnotesize
\begin{tabular}{lp{11.4cm}}
\toprule
Column & Content \\
\midrule
\texttt{batch} & 1 to 5: the batch of the block in which the job ran; batch 5 holds the re-runs \\
\texttt{stage} & 1, 2 or 3: the nested stage of the job list that holds the task (118, 37 and 76 tasks) \\
\texttt{arm} & F, R, C, H or T \\
\texttt{task} & task identifier in the public task collection the panel was drawn from \\
\texttt{repository} & repository of the task (owner/name); the unit of clustering \\
\texttt{repeat} & 1 to 4 \\
\texttt{attempt} & 1; 2 for the re-run of an interrupted job \\
\texttt{row\_class} & agent: an analysed run. infrastructure: an interrupted job, not analysed \\
\texttt{outcome\_id} & A00 solved; A01 unsolved, run completed; A02 tool-call budget reached; A05 unsolved at the five-minute limit; A06 no submission; A18 unsolved, over-long prompt rejected; I09 interrupted job \\
\texttt{resolved} & 1 if the run is solved (the tests of the task pass on the edits the run left), else 0 \\
\texttt{t\_agent\_s} & seconds of the agent phase of the job \\
\texttt{t\_setup\_s} & seconds of the setup phase, before the agent starts \\
\texttt{t\_verify\_s} & seconds of the grading phase; empty if no test was run \\
\texttt{model\_calls} & chat requests of the run that returned a response \\
\texttt{model\_calls\_failed} & chat requests of the run that raised an error instead \\
\texttt{tool\_calls} & tool calls made by the agent \\
\texttt{tokens\_in} & prompt tokens, summed over the requests of the run \\
\texttt{tokens\_out} & completion tokens, summed over the requests of the run \\
\texttt{max\_prompt\_tokens} & largest prompt of the run, in tokens \\
\texttt{summaries} & history summaries (compactions) made during the run \\
\texttt{patch\_chars} & characters of the diff the run left; 0 if none \\
\texttt{test\_exit} & exit status of the test command; 0 = tests passed; $-1$ = no test was run \\
\texttt{start\_utc} & start of the job (UTC) \\
\texttt{end\_utc} & end of the job (UTC) \\
\texttt{requests\_checked} & returned requests of the run that the frozen checker re-rendered \\
\texttt{requests\_changed} & of them, requests whose prompt the reference rule of arm F would render differently (the exposure count; 0 in F) \\
\texttt{check\_passed} & 1 if the run passed the prompt check \\
\bottomrule
\end{tabular}
\end{table}

\end{document}